\documentclass[letterpaper]{article} %
\pdfoutput=1
\usepackage{aaai22}  %
\usepackage{times}  %
\usepackage{helvet} %
\usepackage{courier}  %
\usepackage[hyphens]{url}  %
\usepackage{graphicx} %
\usepackage{natbib}  %
\usepackage{caption} %
\usepackage{color}
\usepackage{appendix}
\usepackage{amsmath}
\usepackage{bm}
\usepackage{amssymb}

\usepackage{adjustbox}
\usepackage[acronym]{glossaries}
\usepackage{chngcntr}
\usepackage{etoolbox}

\providetoggle{arxiv}
\settoggle{arxiv}{true}

\makeglossaries

\newacronym{destination_predictor}{PDPC}{Physical Destination Predictor for Commands}

\def\s#1{{\mathbb{#1}}}
\def\m#1{{\boldsymbol{#1}}}
\def\v#1{{\boldsymbol{#1}}}

\usepackage{subfig}
\usepackage{makecell}
\usepackage[switch]{lineno}

\title{
Predicting Physical World Destinations for Commands Given to Self-Driving Cars
}
\author{
     Dusan Grujicic\footnotemark[1]\textsuperscript{\rm 2},
    Thierry Deruyttere\thanks{equal contribution}\textsuperscript{\rm 1},
    Marie-Francine Moens\textsuperscript{\rm 1},
    Matthew Blaschko\textsuperscript{\rm 2}
\\
}
\affiliations {

    \textsuperscript{\rm 1} Department of Computer Science, KU Leuven, Belgium \\
    \textsuperscript{\rm 2} Department of Electrical Engineering, KU Leuven, Belgium \\ 
    \{dusan.grujicic, thierry.deruyttere, sien.moens, matthew.blaschko\}@kuleuven.be
}

\begin{document}

\maketitle

\begin{abstract}
\begin{quote}
In recent years, we have seen significant steps taken in the development of self-driving cars. Multiple companies are starting to roll out impressive systems that work in a variety of settings. 
These systems can sometimes give the impression that full self-driving is just around the corner and that we would soon build cars without even a steering wheel. 
The increase in the level of autonomy and control given to an AI provides an opportunity for new modes of human-vehicle interaction. 
However, surveys have shown that giving more control to an AI in self-driving cars is accompanied by a degree of uneasiness by passengers.
In an attempt to alleviate this issue, recent works have taken a natural language-oriented approach by allowing the passenger to give commands that refer to specific objects in the visual scene.
Nevertheless, this is only half the task as the car should also understand the physical destination of the command, which is what we focus on in this paper.
We propose an extension in which we annotate the 3D destination that the car needs to reach after executing the given command and evaluate multiple different baselines on predicting this destination location. Additionally, we introduce a model that outperforms the prior works adapted for this particular setting. 

\end{quote}
\end{abstract}

\section{Introduction}

\begin{figure*}%
    \centering
    \subfloat[\centering Frontal View]{{\includegraphics[width=\columnwidth, height=4.4cm]{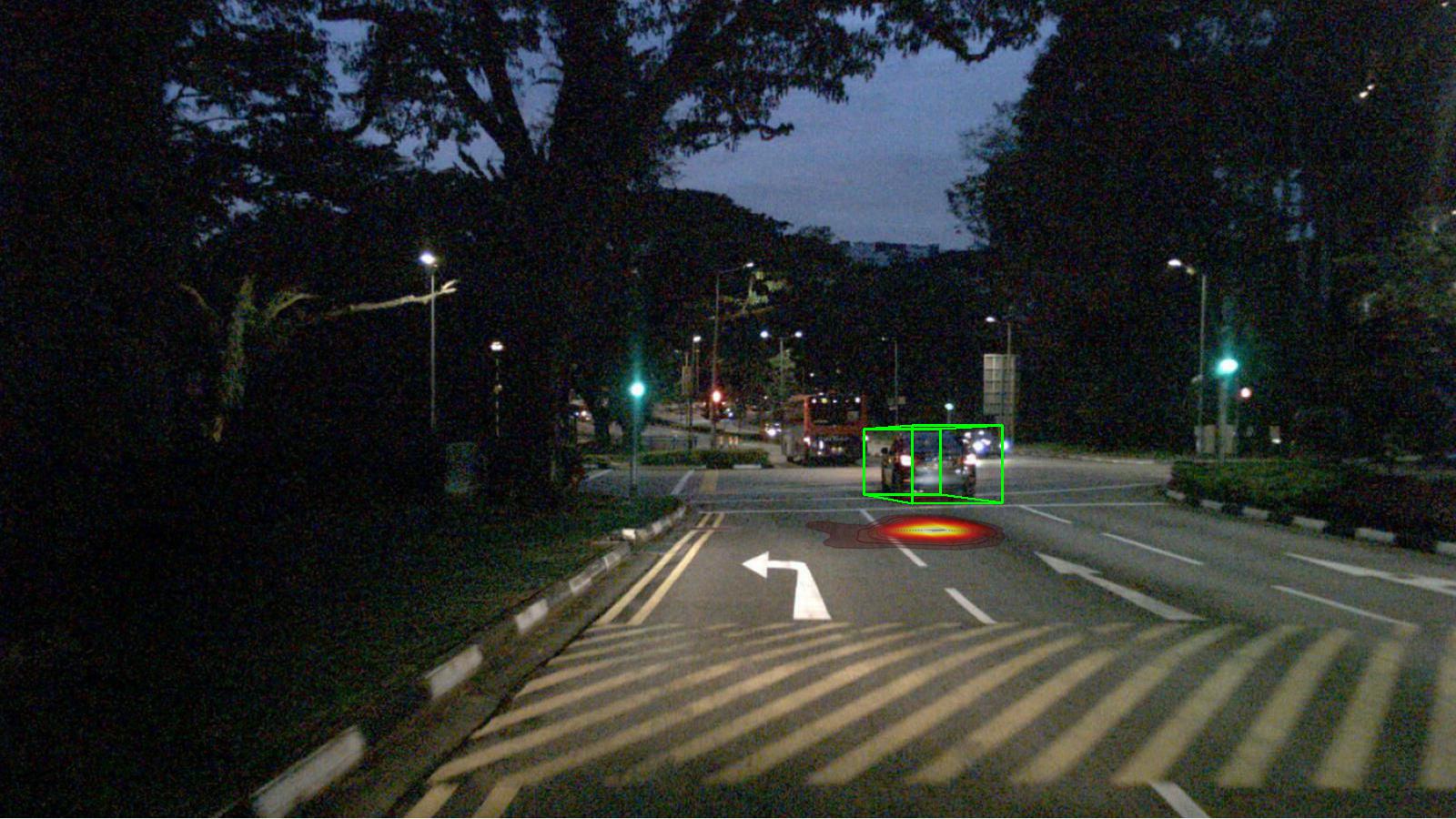} }}%
    \qquad
    \subfloat[\centering Top-down %
    View]{{\includegraphics[width=\columnwidth, height=4.4cm]{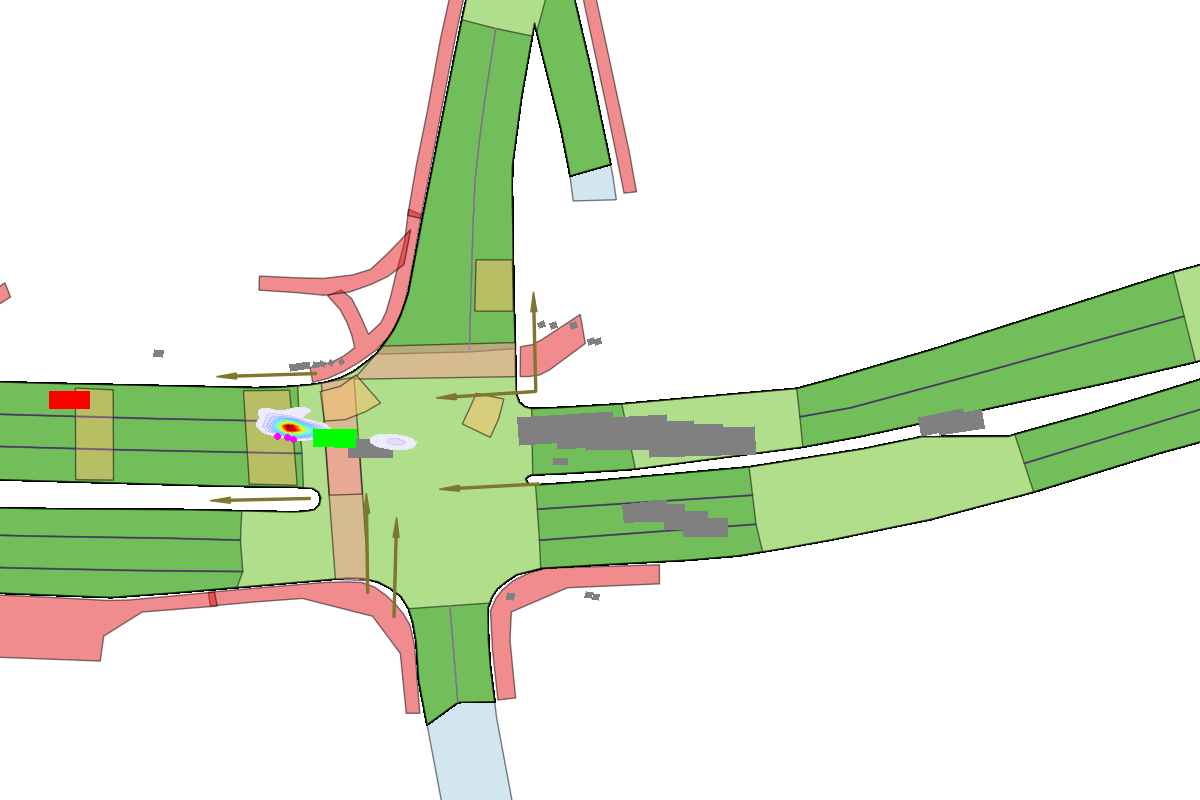} }}%
    \caption{Visualization of the predictions made for the task at hand. The issued command here is ``Oh no I'm on the wrong lane, change lanes so I drive behind the car in front''. On the left, we have visualized the predicted destination of the command as a heatmap in the physical world and the referred object with a green bounding box. On the right, we have the prediction on a top-down view of the scene (under the purple dots).
    The red rectangle is our location, the green rectangle is the predicted object referred to by the command, and the gray rectangles are the other detected objects. The purple dots are the ground truth destinations.
    Prediction made with our proposed model from Sect \ref{sect:destination_predictor}. 
    }%
    \label{fig:show_prediction}%
\end{figure*}

Many companies are in the race to be the first to develop fully self-driving cars. 
It is expected that once this technology matures, manufacturers might entirely remove the steering wheel. 
While some enthusiasts are eagerly waiting for this day, surveys \cite{othman2021public,deruyttere2021giving,schoettle2014survey}
have indicated that an average person is wary of relinquishing physical control of the car. However, it was found that the notion of being able to give spoken commands that can change the behavior of the vehicle tends to make people much more at ease \cite{deruyttere2021giving}.

In recent years, researchers have investigated systems where passengers can give commands to self-driving cars. For instance, \cite{vasudevan2021talk2nav,chen2019touchdown} consider navigational commands such as ``Take the first left and at the red building turn right. Afterwards, drive to the white building''. 
On the other hand, Talk2Car \cite{deruyttere2019talk2car} considers commands that abruptly change the driving dynamics in the imminent future, such as ``Let me out near my friend with the red shirt on the left''.
The two former datasets feature graph-based navigation at a city level, where each node represents an individual street scene. The latter dataset takes a lower level approach and focuses on predicting objects referred to by the command within an individual street scene, without addressing the navigational aspects of executing the command.

Taking the same low-level approach, we focus on the environment within the observable vicinity of the current location and extend the Talk2Car dataset with the 3D physical destination for each given command. 
This extension is interesting as it requires models to interpret spatial language in a 3D visual context.
The works of \cite{dendorfer2020goal,mangalam2020not} suggest that it is also beneficial to first predict the end goal of a path before starting to navigate.
Therefore, we focus on predicting the end destination given the passenger command (an example is given in Figure \ref{fig:show_prediction}), after which, in a practical setting, one of the many already existing systems could be tasked with safely navigating to it within the dynamic environment \cite{messaoud2020trajectory}.

Additionally, indicating the destination as understood from the command offers additional visual insight to the passenger in their interaction with the car. It may also alleviate a degree of uneasiness that comes with the absence of direct physical control.
To the best of our knowledge, our dataset is the first of its kind where a model needs to predict the physical, 3D destination of the self-driving car of the passenger that has given a command. We will refer to the car in which the passenger is as ego car from now on.
In this paper, we investigate if existing models are adequate enough to perform well on this task.
In addition to this, we propose a new model called \textbf{\gls{destination_predictor}} for this destination task which outperforms our evaluated baselines. 
Our proposed model uses a feature pyramid network to predict distributions at different feature levels and finally aggregates them into a distribution mixture.
\newpage
\noindent
The contributions in this paper are the following:
\begin{itemize}
    \item We introduce an expansion, \textbf{Talk2Car-Destination}, for the Talk2Car dataset \cite{deruyttere2019talk2car} where we annotate the \textbf{possible destinations in the physical world} for a given command, i.e., where the ego car should be after it has executed the command.
    \item We evaluate multiple baselines on our expansion and show that the dataset is challenging.
    \item We propose a novel method called \textbf{\acrfull{destination_predictor}} that achieves state-of-the-art results compared to our evaluated baselines and outperforms them by as much as 32\% in certain measures.
\end{itemize}
\noindent
The Talk2Car-Destination dataset is available at \url{https://github.com/ThierryDeruyttere/Talk2Car-Destination}

\section{Related Work}
In this section, we describe previous works on future prediction relevant for the destination prediction task featured in the Talk2Car-Destination dataset. Moreover, our task requires to detect the object referred to by the command and the 3D locations of objects in the physical world, and hence, we also list the relevant works on query understanding and 3D object detection.

\subsection{Future Prediction}
Future prediction features a broad range of different tasks, such as future image segmentation prediction \cite{luc2017predicting}, future action prediction \cite{rodriguez2018action}, to future frame prediction \cite{liu2018dyan}. 
The task of predicting the future location of a moving object in the physical world featured in the works of \cite{lee2017desire,yagi2018future,makansi2019overcoming} is conceptually related to our task of predicting the car's end location after executing the command. 
As the future is often uncertain, it is beneficial that the model predicts multimodal distributions of the destination location. \citet{makansi2019overcoming} tackle this task with an approach based on Mixture Density Networks (MDN).
However, they observe that the Mixture Density Networks are prone to experiencing mode collapse.
To overcome this, they propose an adaptation to the Winner-Takes-All loss \cite{guzman2012multiple} where they only update the top-$k$ hypotheses while decreasing the value of $k$ over time. Subsequently, they fit the mixture distribution to the estimated hypotheses through soft assignments. A problem with the aforementioned solution is the need for defining the number of mixture components a priori. To overcome this limitation, RegFlow \cite{zikeba2020regflow} avoids making any assumptions on the underlying distribution by using Continuous Normal Flows (CNF) \cite{chen2018neural}, thus decreasing the number of distribution parameters. 

Another relevant future prediction task is trajectory prediction, where the models are tasked with jointly predicting the future location of the object and the path to it~\cite{narayanan2021divide,liang2020garden,mohamed2020social}. 
\citet{dendorfer2020goal} propose GoalGAN, which first encodes the trajectory of pedestrians with an LSTM before passing the encoded trajectory to an encoder-decoder architecture that takes the top-down view of the scene as input.
The output of the encoder-decoder is a probability map of the pedestrian's future locations.%
This information is passed to the next module that predicts the possible path towards this goal.
Their model uses a GAN to predict if the generated trajectories come from the same distribution as the ground truth trajectories.
Another example is PECNet
\cite{mangalam2020not} where the future destination is modeled using a CVAE \cite{lee2017desire}, by first encoding the past trajectory of the object and passing it to the CVAE, without the use of visual features of the image.
Then, along with the predicted destination, the model predicts the trajectory by applying social pooling \cite{alahi2016social}.
This paper utilizes the components from GoalGAN, PECNet, and RegFlow as the baselines on our new dataset. We opt for RegFlow over \cite{makansi2019overcoming} as the training code of their model was not available.

\subsection{Query Understanding}
Our task requires the model to find the referred object of the command (Visual Grounding) and predict the car's end location after executing the command (Spatial Language Interpretation).
In Visual Grounding \cite{hu2016natural,deng2018visual,akbari2019multi}, the model receives a query, an image, and potentially a set of objects. The model is asked to predict the object in the image referred to by the query.
A popular approach for joint reasoning over the query and vision is the use of Transformers \cite{stacked_vlbert,kamath2021mdetr,du2021visual}.
Multi-step reasoning models \cite{Hudson2018,deng2018visual} and modular models \cite{yu2018mattnet} that reason over image and/or objects have also been popular. In this paper we use a similar model to \cite{rufus2020cosine}.

With regards to grounding text to physical locations, \citet{lourentzou2017text} focus on predicting physical geographic origins of Twitter posts, while \citet{grujicic2020learning} localize medical text referring to anatomical concepts to their corresponding physical locations in the human body. Finally, \cite{ collell2021probing} have also worked on understanding the implicit spatial relationships of queries and objects.

Focusing on the domain of autonomous-driving, the work of \citet{sriram2019talk} presents a dataset of natural language instructions and trajectories in a synthetic environment. Our dataset, on the other hand, features command and destination annotation in the natural, non-simulated environments. In a work concurrent to ours, \citet{rufus2021grounding} introduce a dataset built on top of the Talk2Car which features destination annotations in the frontal camera view. In contrast, our work focuses on destinations in the car's 3D environment, in which the physical layout and distances can be considered.  

\subsection{3D Object Detection}
The locations of other objects in the scene give important cues for predicting the final destination in the physical world. Therefore, aside from predicting the referred object, an essential step in our task is the accurate prediction of 3D bounding boxes of the objects visible in the frontal view. A popular approach to this task is to use LIDAR/RADAR point clouds \cite{zhou2018voxelnet,lang2019pointpillars,engelcke2017vote3deep,zheng2021se}. 
For instance, \citet{yin2021center} use LIDAR to predict 3D objects by first predicting the center of each object in a top-down point cloud view. Afterwards, the model regresses the centers to the object's 3D size, orientation, and velocity. In the second stage, it refines the predicted attributes. 
\citet{lang2019pointpillars} perform 3D object detection by aggregating voxels into vertical columns, also called pillars. Afterwards, 2D convolutions are applied on said pillars to predict 3D objects.

In addition to using point clouds to predict 3D bounding boxes, there are approaches that rely on image data alone. This is known as monocular 3D object detection \cite{chen20153d,roddick2018orthographic}.
An example of such a method is FCOS3D \cite{wang2021fcos3d}, which is very similar to the CenterNet \cite{zhou2019objects} architecture. 
It predicts 3D bounding boxes by creating a feature pyramid network of an image, and then, at each level and cell of the feature map, it predicts the 3D attributes of objects. In this work, we use FCOS3D to predict 3D object locations.

\section{Dataset}

\begin{figure*}%
    \centering
    \subfloat[\centering Frontal View]{{\includegraphics[width=\columnwidth,height=4.4cm]{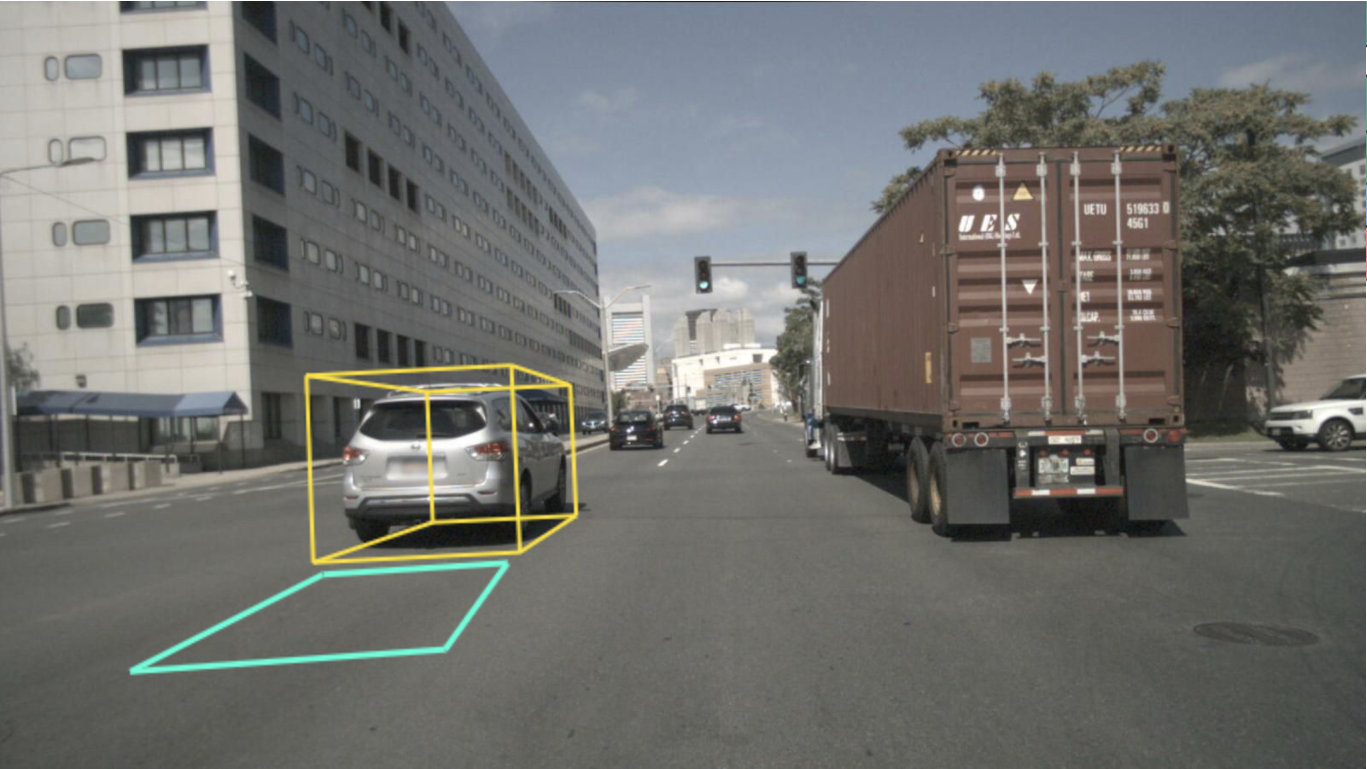} }}%
    \qquad
    \subfloat[\centering Interactive Top-down %
    View]{{\includegraphics[width=\columnwidth, height=4.4cm]{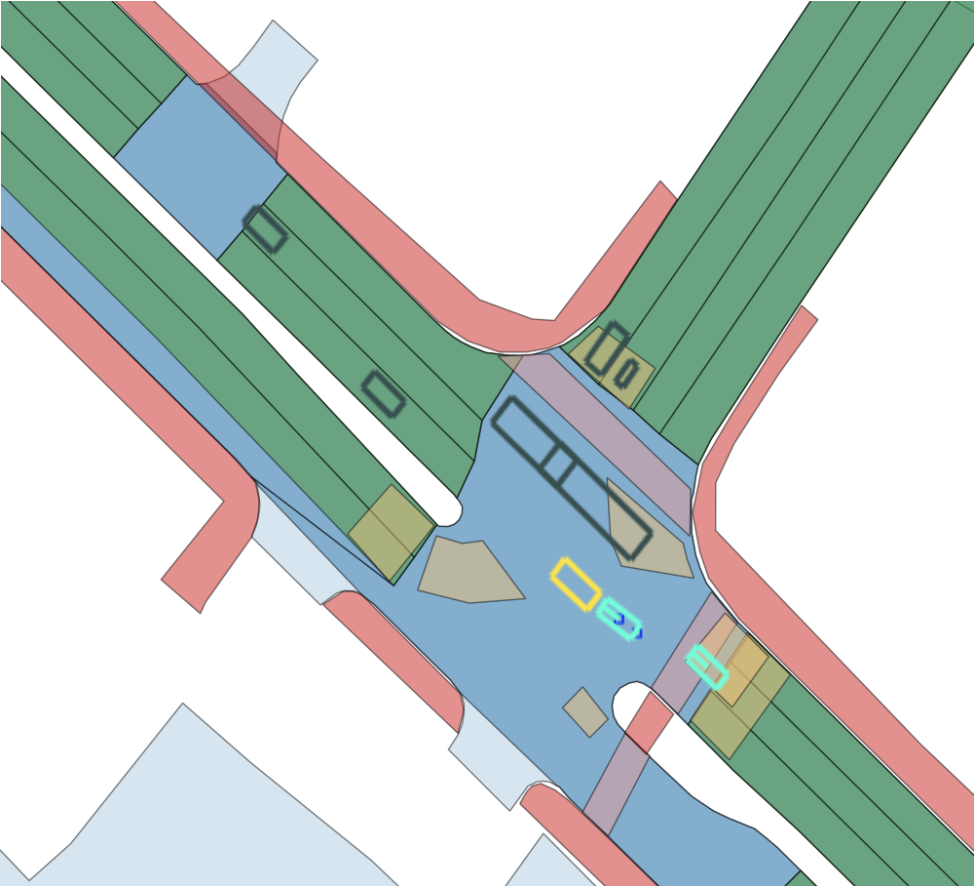} }}%
    \caption{An example from our annotation tool that was made with EasyTurk \cite{krishna2019easyturk}, given the command \textit{``get behind this silver SUV"} and the Talk2Car object that it refers to (indicated with a yellow bounding box), the annotators were asked to indicate the destination in the interactive %
    top-down view. The annotated destination is the cyan rectangle visible in the frontal view and the cyan rectangle behind the yellow rectangle in the top-down %
    view.}%
    \label{fig:annotation}%
\end{figure*}

In this paper, we extend the Talk2Car dataset \cite{deruyttere2019talk2car} which is built on top of the nuScenes dataset \cite{caesar2020nuscenes} and features driving scenes from Boston (right-hand traffic) and Singapore (left-hand traffic) in different lighting (night vs day) or weather conditions (sun vs rain). In addition to the 360 degrees camera view, it also contains LIDAR/RADAR point clouds.
Talk2Car adds unrestricted natural language commands that refer to a specific object in a street setting, and consists of 8349, 1163, and 2447 samples in the training, validation, and test sets.
We extend the Talk2Car dataset by having three annotators per each command indicate the physical end position of the car after executing the command in the top-down view (Figure \ref{fig:annotation})
together with the true intent of the command i.e., the intent of the command ``park next to the tree'' is ``park''.
For the true intent of the command, the annotators received a set of predefined options to choose from and the final annotation was determined in a majority vote among three annotators.
In case of no majority, additional annotators were asked to indicate the intent until a majority was reached. 
The annotation process was performed using a custom annotation interface created with EasyTurk \cite{krishna2019easyturk}, hosted on Amazon Mechanical Turk.

In addition to the annotation, we also normalized the data for easier training. We first rotated all top-down views (as seen in Figure \ref{fig:annotation}(b)) in such a way that the ego car is always facing directly right. The resolution of the frontal camera view images is $1600\times900$. We normalize the sizes of the top-down views such that they correspond to physical map regions of $120 \times 80$ meters in size, with the ego car located seven meters from the left boundary and halfway along the height of the top-down view. The resulting resolution of the top-down views is $1200 \times 800$, i.e., distance of ten pixels in the top-down view corresponds to the distance of one meter. 

\begin{figure}
    \centering
    \includegraphics[width=0.65\linewidth]{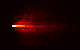}
    \caption{Destination distribution of the Talk2Car-Destination dataset. This heatmap represents the destinations 50 meters in front of the ego car and 20 meters to the right and left of the ego car.}
    \label{fig:destination_heatmap}
\end{figure}

We removed the commands that referred to objects outside such a map patch, which was the case in 0.51\% of the samples. This process resulted in 8301, 1159, and 2439 samples in the train, validation, and test set.
The average distance from the center of the ego car to the destination is 26.54 meter.
In Figure \ref{fig:destination_heatmap} we display a heatmap of the destinations. Each sample consists of a command, frontal and top-down view, ground-truth 3D object annotations from the NuScenes dataset, three different destination annotations, as well as the annotated intent of the command. \iftoggle{arxiv}{
A breakdown of the intents over the splits can be seen in Table \ref{tab:intent_distribution} in Appendix \ref{app:dataset}.
}
{
}

We dub this extension \textbf{Talk2Car-Destination}. The statistics from the original Talk2Car dataset \cite{deruyttere2019talk2car} regarding the driving locations, weather conditions, times of day etc. apply to Talk2Car-Destination as well. \iftoggle{arxiv}{
Interested readers can find more details in Appendix~\ref{app:dataset}.
}
{
}

\section{Methods}
\begin{figure*}
    \centering
    \includegraphics[width=\linewidth,height=12cm,keepaspectratio]{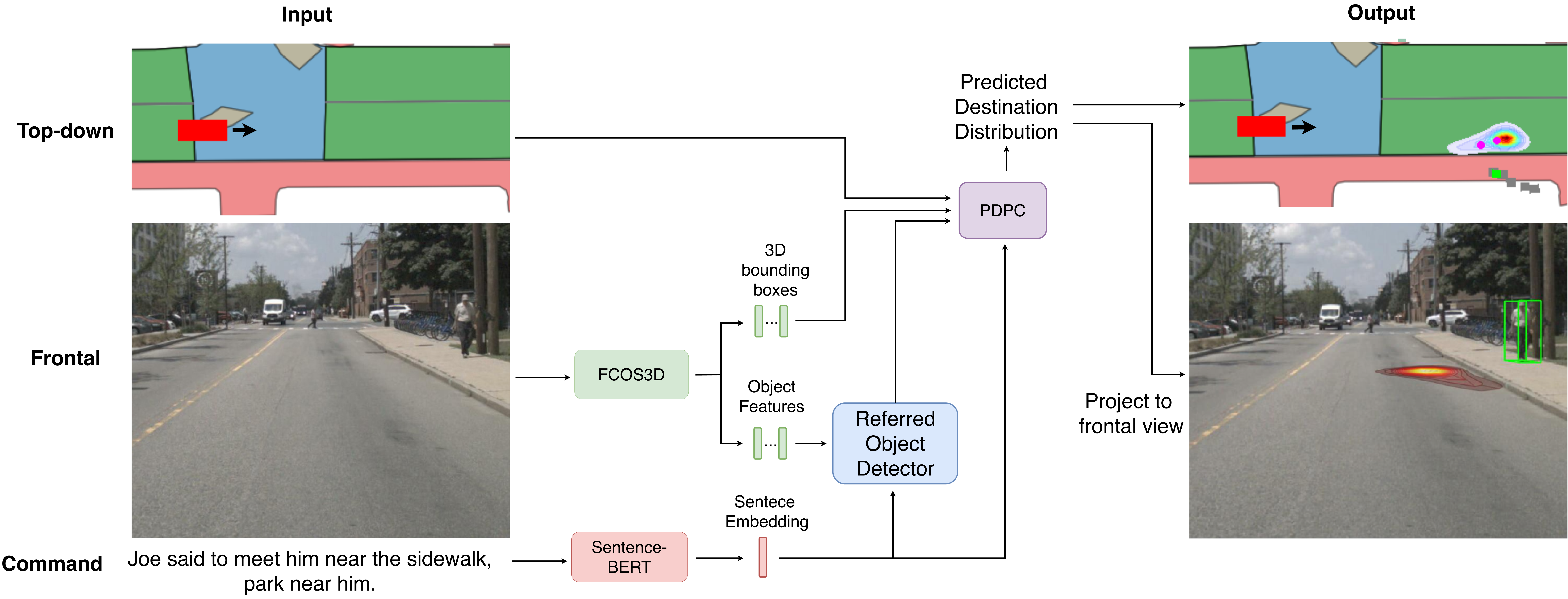}
    \caption{The above image depicts the flow of the architecture used in this paper. As input, we require a top-down image of the road, a car's frontal image, and a (written) command. The frontal image is passed through a monocular 3D object detector (i.e., FCOS3D) to extract 3D bounding boxes and object features. The command is passed through a sentence encoder (i.e., Sentence-BERT). The embedded sentence and object features are then passed through a Referred Object Predictor (Sect. \ref{sect:referred_obj_detector}) to predict the object referred by the command. Afterward, the predicted referred object, together with the sentence embedding, the 3D bounding boxes, and the top-down image are passed through our proposed destination predictor (Sect. \ref{sect:destination_predictor}) to finally predict a destination distribution on the top-down image. This distribution can also be projected to the frontal view as seen in the image. The red car on the top-down image represents the ego car. The black arrow indicates the driving direction. The green rectangle is the predicted referred object. Gray objects are other detected objects.}
    \label{fig:flowchart}
\end{figure*}

We formally define the destination prediction task as follows: given an image of the frontal view of the car $\textbf{I}$, a command $q$ and the representation of the top-down view containing the road layout $\textbf{L}$, the model is asked to predict a distribution $\textbf{P}$ of the destination the car needs to reach in order to execute the command. We first extract the locations of the objects around the ego car using a 3D object detector. As we later demonstrate in Sect. \ref{subsubsect:exp_ref_vs_no_ref}, the location of the object indicated by the command is highly indicative of the the final destination. Therefore, we also formulate the task of referred object detection, where given the command and the set of object proposals obtained from a 3D object detector, we predict the object referred by the command. Figure \ref{fig:flowchart} presents the flowchart of our approach. 
Focusing on the destination prediction, we decouple it from the already established task of referred object detection on Talk2Car \cite{deruyttere2019talk2car}. 
The predicted referred object and the 3D object proposals are subsequently provided as input to the destination prediction model during the training and evaluation phases. We do not utilize ground-truth annotation from either NuScenes or Talk2Car during training, as they would be unavailable in a deployed setting.
In the following subsections, we describe our main building blocks. We use the following notation: vectors are represented as $\v{a}$ and their size $\s{R}^{d1}$ where $d1$ is the size. 
Matrices are denoted as $\m{M}$ and their size $\s{R}^{d1 \times d2}$ where $d1$ and $d2$ are the size of the first and second dimensions. The layers of multi-layer perceptrons (MLP) are indicated between square brackets, e.g. $[\m{W}^{d1 \times d2}, \m{W}^{d2 \times d3}]$ denotes a two layer MLP. 

\subsection{3D Object Detector}

The availability of LIDAR in the Talk2Car-Destination points towards the use of LIDAR-based 3D object detectors, given the effectiveness such an approach has demonstrated in previous works.
However, in our case, we require accurate bounding boxes for the visible objects in the frontal view of the car, which are the focus of the given commands.
From preliminary experiments, we found the use of LIDAR-based methods to be inadequate for our use case. The reason is twofold.
First, we observed that when projecting the predictions of the LIDAR-based method to the frontal image, there was often a substantial shift between the projected bounding boxes and the objects. This is likely due to the different speeds at which the camera and LIDAR sensors capture data, as well as the positioning offset between the camera and the LIDAR sensor, causing both temporal and spatial parallax.
Second, the objects referred to by the commands tend to be at greater distances from the ego car. We find that after 70 meters, the LIDAR-based object detector abruptly becomes unable to detect objects accurately\footnote{https://www.nuscenes.org/nuscenes\#data-collection}.
In contrast, the performance drop of monocular-based detectors at larger distances tends to be less dramatic.

We performed a comparison between FCOS3D \cite{wang2021fcos3d}, and CenterPoint \cite{yin2021center}, where we evaluated the average distance between the ground-truth location of the referred object and the closest 3D object proposal.
We found that FCOS3D achieves an average discrepancy of 1.50 meters, which is a significant improvement over the 2.64 meters of CenterPoint as the latter does not output predictions at distances greater 70 meters. 
\iftoggle{arxiv}{A detailed comparison of the models' performance at different distance ranges of the referred object can be found in the Appendix~\ref{app:3d_obj}. }

We therefore opt for a monocular 3D object detector, which we found to be more suitable for our task in the light of aforementioned limitations of LIDAR-based approaches. More specifically, we use FCOS3D as, at the time of writing, it is one of the highest performing vision-only 3D object detectors on nuScenes with publicly available code\footnote{https://github.com/open-mmlab/mmdetection3d}. 

\subsection{Referred Object Detector}
\label{sect:referred_obj_detector}
From our data and preliminary experiments, we observe that the object referred to by the command is often in relation to the destination of the car, and that the information on the referred object location is central for predicting the destination of the command. For instance, to execute the command ``Park next to the man on the bench'', the system ought to know where the ``man on the bench'' is before predicting the parking location. We train a referred object detector to indicate which object the command relates to and feed the predicted referred object and other detected objects as the input to the destination prediction model.

We notice that the use of the features from the object detector was left unexplored by the previous works on Talk2Car, all of which relied on using pre-trained image feature extractors and then optionally fine-tuning them.
In the vein of \cite{anderson2018bottom}, we evaluate the benefits of directly using the object features $\s{R}^{1536}$ of FCOS3D as local object representations in the referred object detection task by feeding these features as input to the model used in the work of \cite{rufus2020cosine}.
The model uses a Sentence-BERT \cite{reimers2019sentence} to create a $\s{R}^{768}$ command embedding, which is then passed through the following MLP: [$\m{W}^{768\times1024}$, ReLU, $\m{W}^{1024\times1024}$]. The extracted visual features from FCOS3D of size $\s{R}^{1538}$, which contain the concatenated outputs of all prediction heads, extracted at the activation map location corresponding to the predicted bounding box, are passed through the following MLP: [$\m{W}^{1538\times1024}$, ReLU, $\m{W}^{1024\times1024}$].
Finally, we compute the dot product between command and the visual features of each object and construct a probability distribution over object proposals by performing the softmax operation on the products. 

\subsection{Layout Encoding}
\label{sect:layout_encoding}
We construct a top-down layout representation ($L^*$), leveraging the road information, the ego car position, the top-down view projections of all FCOS3D detected objects in 3D, as well as the prediction of the object referred to by the command. The layout representation takes a form of a 3D tensor, with 3 channels reserved for the road information (an RGB image representing the lanes and the road layout), 1 channel reserved for the ego car (a binary grayscale image with a white polygon representing the car at the position of the ego car), 1 channel reserved for the referred object bounding box, and one channel per each object class, as detected by the 3D object detector, where each channel contains a binary grayscale image with white polygons at the positions of the top-down view projections of the detected 3D bounding boxes of objects of it corresponding class.

\subsection{Destination Predictor}
\label{sect:destination_predictor}
In this subsection, we describe our proposed destination predictor which we dub \textbf{\acrfull{destination_predictor}}.
We opt for a fully convolutional and spatial-based approach, where the destination distribution is modeled as a mixture distribution, where a distribution mixture component is predicted at every location in the extracted spatial feature maps.

Our method receives a command and a top-down layout augmented with the detected object locations and the referred object (Sect. \ref{sect:layout_encoding}) in the current visual scene as input.
We pass the layout through a Feature Pyramid Network (FPN) to extract feature maps $\s{R}^{C \times W_i \times H_i}$ at four different scales with $C$, $W_i$, and $H_i$ being the number of channels (256 in our case), width and height of a feature map at scale $i$, respectively. 
The different scales share a series of fully convolutional blocks that process the feature maps into a set of spatial features at each grid location. 
We use stride of 1, kernel size of 3 and padding of 1 in order to preserve the size of the original feature map. 
Then, a multi-layer perceptron $[\m{W}^{768\times 512}, ReLU, \m{W}^{512 \times C}]$ is used to project the command embedding to the same size as the channel dimension of the convolutional block. 
Afterwards, a dot product between the projected command and the channel dimension of the feature maps is performed to produce a $\s{R}^{1 \times W_s \times H_s}$ tensor.
A softmax operation is then used to get a probability distribution over the spatial cells to produce a weights tensor that indicates which cells align more with the command.
Finally, we multiply the feature map $\s{R}^{C \times W_i \times H_i}$ with this weights tensor.
Another shared series of fully convolutional blocks is applied on the attended feature maps at each scale, after which the task specific convolutional heads, also shared across different scales, are used to predict the offset between the grid location and the distribution mean, as well as the standard deviation and mixture weight for each Gaussian mixture component. 
We aggregate the mixture components from each location at each scale level into one large density mixture with $\sum_{i=1}^{S} W_i \times H_i$ components, which we fit to the destination samples. The distribution of the destination $\v{y}$, given the layout encoding $\m{L^*}$ (Sect \ref{sect:layout_encoding}) and command $\v{c}$ is therefore represented with a Gaussian mixture: 

\begin{equation}
p(\v{y}|\m{L^*}, \v{c})=\sum_{i=1}^{S} \sum_{w=1,h=1}^{W_i, H_i} \pi_{iwh} \phi\left(\v{y} | \v{\mu}_{iwh}, s_i \v{\sigma}_{iwh}\right)
\end{equation}
where $S$ is the number of resolution scales, while $H_i$ and $W_i$ represent the height and the width of the feature map at the $i$-th scale. The Gaussian mixture component $\phi$ at the location $h,w$ in the $i$-th scale is parametrized by its mean $\v{\mu}_{iwh} \in \s{R}^2$ and standard deviation $\v{\sigma}_{iwh} \in \s{R}^2$, while $\pi_{ihw}$ represents its mixture weight. We use softmax to normalize the mixture weights such that $\sum_{i=1}^{S} \sum_{w=1,h=1}^{W_i, H_i} \pi_{iwh} = 1$. 
For each location in low resolution scale grids $[w, h]$, we keep track of the corresponding spatial location in the input layout, $\v{l}_{ihw} = [w', h']$, where $w'$ and $h'$ are computed as $w' = wk_i + \left \lfloor{\frac{k_i}{2}} \right \rfloor$ and $h' = hk_i + \left \lfloor{\frac{k_i}{2}} \right \rfloor$ and $k_i$ represents the down-sampling rate at scale $i$. At each location, the network directly predicts the mixture weights and the standard deviation, as well as the offset $\v{o}_{ihw}$, after which the mean is computed as $\v{\mu}_{ihw} = \v{o}_{ihw} + \v{l}_{ihw}$.
Predicting the offset from the corresponding physical location, as opposed to directly predicting the mean, allows for a regularized model to predict small offsets within the immediate neighborhood of each grid cell.
At each resolution scale, the predicted standard deviation is multiplied by a learnable parameter $s_i$, allowing the network to adjust the variance of the Gaussian components for each resolution scale.

Our model uses Negative Log Likelihood (NLL) to minimize the log probability of the ground truth destinations under the predicted Gaussian Mixture, or in other words we minimize: 

\begin{equation}
\mathcal{L}_j=-\log \left(p(\v{y}_j|\m{L}, \v{c})\right)
\end{equation} ,

where for $N$ different destinations provided by different annotators we minimize $ \frac{1}{N} \sum_{j=1}^{N} \mathcal{L}_j$. 
The benefit of our model over \cite{makansi2019overcoming} is that we do not require to a priori define how many components we want. Our model predicts a component in each cell in the spatial grid and can learn to give certain components very low weights, and effectively ignore them, if needed.

\section{Experiments}
In this section we explain the performed experiments, the evaluation measures and the baselines. 

\subsection{3D Object Detector}
We train FCOS3D \cite{wang2021fcos3d} only on the Talk2Car training scenes to predict bounding boxes as they do not overlap with the scenes from the Talk2Car validation or test sets.
We use the default parameters provided by the authors.

\subsection{Referred Object Detector}
Our referred object detector uses a Sentence-BERT \cite{reimers2019sentence} to compute the command embedding, whose parameters remain fixed during training. \iftoggle{arxiv}{Information about the training and hyperparameters of this model are in the Appendix. }

We evaluate this model by measuring the Intersection over Union (IoU) between between the boxes of the predicted and ground truth objects. 
If IoU $ > 0.5$, we consider the predicted object to be correct.
We refer to this as $IoU_{0.5}$. The IoU is defined as:

\begin{equation}
\label{eq:iou}
IoU = \frac{\text{Area of Overlap between two boxes}}{\text{Area of Union of the two boxes}}.
\end{equation}

\subsection{Destination Prediction Measures}

To evaluate the performance of the destination prediction models on our dataset, we use the following three measures: 
\begin{itemize}
    \item Average Displacement Error (\textbf{ADE}): Sample average of the average distance between the samples of the predicted destination distribution from the closest ground truth destination. Measured in meters [m].
    \item Median Displacement Error (\textbf{MDE}): Sample median of the average distance between the samples of the predicted destination distribution from the closest ground truth destination. Measured in meters [m].
    \item Prediction Accuracy with threshold $k$ meters (\textbf{PA}$_k$): measures the rate of predicting the destination within the radius of $k$ meters around the ground truth destination. 
\end{itemize}

 \subsection{Command Encoding}
 The command is processed by using the pre-trained Sentence-BERT \cite{reimers2019sentence}, where the input command is tokenized and fed through the model, and the command representation is obtained by computing the mean of the output token representations. 
 The Sentence-BERT is trained on the combination of SNLI \cite{bowman2015large} and MultiNLI datasets \cite{williams2017broad} to perform the Natural Language Inference (NLI) task of predicting semantic labels for annotated sentence pairs, and has been shown to perform well on a variety of downstream tasks that require high quality sentence representations in a wide range of semantic domains.

\subsection{Baselines}
\begin{table}[]
    \centering
    \begin{adjustbox}{width=0.85\linewidth}

    \begin{tabular}{c||c|c}%
       Model & IoU$_{0.5}$ & \# Params (M) \\
       \hline
       \hline
       \cite{stacked_vlbert} & \textbf{0.710} & 683.80 \\ %
       \cite{luo2020c4av} & 0.691 & 194.97 \\
       \cite{rufus2020cosine} & 0.686 & 366.50 \\
       \hline
    Sentence-BERT+FCOS3D & 0.701 & \textbf{166.29} \\
    \end{tabular}
    \end{adjustbox}
    \caption{The IoU$_{0.5}$ of different models on the Talk2Car test set. The first three models are the previous state-of-the-art models for reference. The ``Sentence-BERT+FCOS3D'' model is the referred object predictor described in Sect. \ref{sect:referred_obj_detector} and is based on \cite{rufus2020cosine} but uses the extracted object features from FCOS3D.}
    \label{tab:referred_object}
\end{table}

\begin{table*}[]
    \centering
    \begin{adjustbox}{width=0.68\linewidth}
    \begin{tabular}{c||c|c|c|c}
      Method & ADE [m] & MDE [m] & PA$_2$[\%] & PA$_4$ [\%]  \\
      \hline 
      \hline
        Random Point & $43.98 \pm 0.12$ & 44.24 & $0.28 \pm 0.00$ & $0.97 \pm 0.00$ %
        \\
        Random Road Point & $37.38 \pm 0.17$ & 37.35 & $0.81 \pm 0.01$ & $2.60 \pm 0.03$ %
        \\
        Pick Ego Car & $25.62 \pm 0.32$ & 21.61 & $0.00 \pm 0.00$ & $0.00 \pm 0.00$  \\
        Random Object & $28.02 \pm 0.23$ & 26.77 & $1.11 \pm 0.06$ & $4.62 \pm 0.14$ \\
        Pick Referred Object & $9.04 \pm 0.18$ & 6.07 & $6.60 \pm 0.50$ & $27.96 \pm 0.91$ \\
        \hline
        SinglePoint & $8.15 \pm 0.34$ & 5.52 & $13.41 \pm 1.38$ & $35.88 \pm 1.94$ \\
        NonParametric & $9.65 \pm 0.36$ & 6.72 & $12.81 \pm 0.64$ & $31.22 \pm 1.20$ \\
        UnimodalNormal & $8.16 \pm 0.32$ & 5.61  & $17.61 \pm 0.92$ & $38.86 \pm 1.44$ \\
        MDN & $8.08 \pm 0.32$ & 5.30 & $16.82 \pm 1.34$ & $38.42 \pm 1.48$ \\
        \hline
        Adapted GoalGAN & $8.65 \pm 0.40$  & 5.31 & $22.60 \pm 1.20$ & $43.09 \pm 1.56$ \\
        Endpoint VAE & $\mathbf{7.84 \pm 0.34}$ & 5.22 & $16.38 \pm 1.34$ & $38.83 \pm 1.86$ \\
        Adapted RegFlow & $17.79 \pm 0.36$ & $15.89$ & $13.01 \pm 0.70$ & $32.23 \pm 0.70$  \\
        \hline
        \gls{destination_predictor} - Base (Ours) & $8.21 \pm 0.36$ & $5.05$ & $28.45 \pm 1.28$ & $47.56 \pm 1.52$ \\
        
        \gls{destination_predictor} - Top-64 Components (Ours) & $7.96 \pm 0.36$ & $4.61$ & $28.97 \pm 1.30$ & $48.42 \pm 1.56$ \\
        
        \gls{destination_predictor} - Top-32 Components (Ours) & $\mathbf{7.82 \pm 0.38}$ & $\mathbf{4.39 }$ & $\mathbf{29.88 \pm 1.36}$ & $\mathbf{49.51 \pm 1.64}$ \\
    \end{tabular}
    \end{adjustbox}
    \caption{ADE, MDE, PA computed for 2m and 4m thresholds (PA$_2$ and PA$_4$, respectively). %
    Finally the results of our model are shown in the bottom of the table. The error bars represent 95\% confidence intervals.}
    \label{tab:all_results}
\end{table*}

To evaluate Talk2Car-Destination and our proposed model, we implemented or adapted the following baselines.
\iftoggle{arxiv}{Additional details on the naive baselines: Random Point, Random Road Point, Pick Ego Car, Random Object, Pick Referred Object, are provided in the Appendix~\ref{app:baselines}.}

\paragraph{SinglePoint} The layout tensor encoded using ResNet18 \cite{he2016deep} to a vector of $\s{R}^{1024}$, concatenated with the sentence-BERT command embedding of $\s{R}^{768}$ is fed through a multi-layer perceptron (MLP) that regresses the input to two-dimensional destinations. 
    
\paragraph{UnimodalNormal} Predicts the destination distribution given the layout tensor and the command after encoding them as in SinglePoint. We model the distribution as a bi-variate Gaussian, by predicting the distribution mean and covariance matrix. We minimize the negative log-likelihood (NLL) of the target destinations under the predicted distribution.
    
\paragraph{MDN} Predicts destination distribution similarly to UnimodalNormal, except that we model distribution as a mixture of bi-variate Gaussian distributions. For each component, we predict the mean, covariance matrix and the mixture weight. We minimize the NLL %
of the target destinations under the predicted Gaussian mixture.
    
\paragraph{NonParametric} The distribution of the destinations is modeled as a histogram over possible locations in a grid, where each grid cell corresponds to a physical location in the top-down view. We minimize the cross-entropy between the predicted distribution over grid cells. The locations in the grid where the cells correspond to ground truth destination locations have the value 1.0 and 0.0 otherwise. We treat the rows and columns of the grid independently to keep the output dimensionality low.

\paragraph{Adapted RegFlow %
 \cite{zikeba2020regflow}}  FlowNet \cite{dosovitskiy2015flownet} is used to encode the top-down view in $\s{R}^{1024}$. Next, we pass the sentence-BERT command embedding through a linear layer of $\m{W}^{768\times768}$ and then concatenate it to the layout encoding. The remainder of the model is unchanged and a hypernetwork is used to train a continuous normalising flow (CNF) model to output a distribution. This model is trained by minimizing the NLL of the CNF.
    
\paragraph{Endpoint VAE \cite{mangalam2020not}} We adapt the endpoint prediction components of PECNet, which is a trajectory prediction model that takes a goal-conditioned approach, where the model first predicts the trajectory endpoint and, subsequently, the trajectory. We utilize the Endpoint VAE, which is the model component that performs the endpoint prediction, and adapt it for our task of destination prediction. In PECNet, the Endpoint VAE encodes the past trajectory and the ground truth endpoint into a latent destination distribution during training. The decoder is then fed a sample drawn from the latent distribution, and the past trajectory representation outputs predicted endpoints. During inference, the latent samples are drawn from a zero-mean isotropic Gaussian. We adapt this approach to our setting by replacing past trajectory encoding with the representation of the spatial layout from the ResNet-18 and the command embedding obtained from the Sentence-BERT.
    
\paragraph{Adapted GoalGAN %
 \cite{dendorfer2020goal}} GoalGAN has a RoutingModule to predict the path, a GoalModule to predict the destination and a MotionEncoder to predict the last locations of the objects. We remove the MotionEncoder as we do not require it. Then, our top-down layout tensor is processed in an encoder-decoder architecture with as output a probability map over the image. After the encoder stage, we introduce the command to the encoded feature map by concatenating it along the channel dimension, after which a convolutional layer is used to project it to a lower dimension. The RoutingModule is kept as-is but we set the length of the path to one instead of $N$ as to predict the destination. We keep the GAN as-is.

\section{Results}

\subsection{3D Object Detector}
FCOS3D achieves a mean Average Precision (mAP) of $30.73$ on the Talk2Car test set scenes.
Additionally, we evaluate the quality of predictions on the referred object detection task.
We project the top-32 confidence scoring predictions to the frontal view and evaluate the rate at which at least one sample achieves an IoU $\geq 0.5$ with the ground truth referred object in the Talk2Car test set. 
We find that such a bounding box exists in 92\% of the cases which is the same as with the bounding boxes of \cite{deruyttere2020commands}, and therefore, no substantial increase in IoU is expected solely from improved bounding box predictions.
The average distance between the 3D bounding box of the ground truth referred object and their 3D predicted bounding is 1.7m on the Talk2Car test set.

\subsection{Referred Object Detectors}
Table~\ref{tab:referred_object} showcases that our model outperforms two out of three state-of-the-art models on the referred object detection task in Talk2Car.
Comparing to \cite{rufus2020cosine}, on which we based our detector, we observe a relative improvement of 2\% in IoU$_{0.5}$ while having less than half the parameters. 

\subsection{Destination Prediction}

In Table \ref{tab:all_results}, we find that our model for destination prediction significantly outperforms all baselines in terms of $PA_2$, $PA_4$, and MDE.
We also find that our model manages to generate 47.56\% of its samples in a radius of 4m (roughly the length of a small passenger vehicle) from the nearest ground truth destination. We also see that the median distance of our base model is 5.05m which is 3.4\% lower than our closest competitor.
As evident from the discrepancy between the ADE and MDE scores, our model, as well as other baseline models such as the Endpoint VAE and GoalGAN, tend to produce large outliers more often than simpler baselines like UnimodalNormal and MDN. However, in an inference setting, our model can be easily adapted by using the top-K components in terms of their mixture weight magnitude, and re-normalizing the mixture weights. This significantly reduces the number of outliers, and when the number of mixture components is thus set to 32 during inference, our model outperforms all baselines in terms of the ADE as well. This is a very desireable feature of our model, which gives it a significant advantage over other competing models. %

\begin{figure*}[h]
    \centering
    \subfloat[\gls{destination_predictor} - No referred object information - Frontal]{{\includegraphics[width=0.45\linewidth]{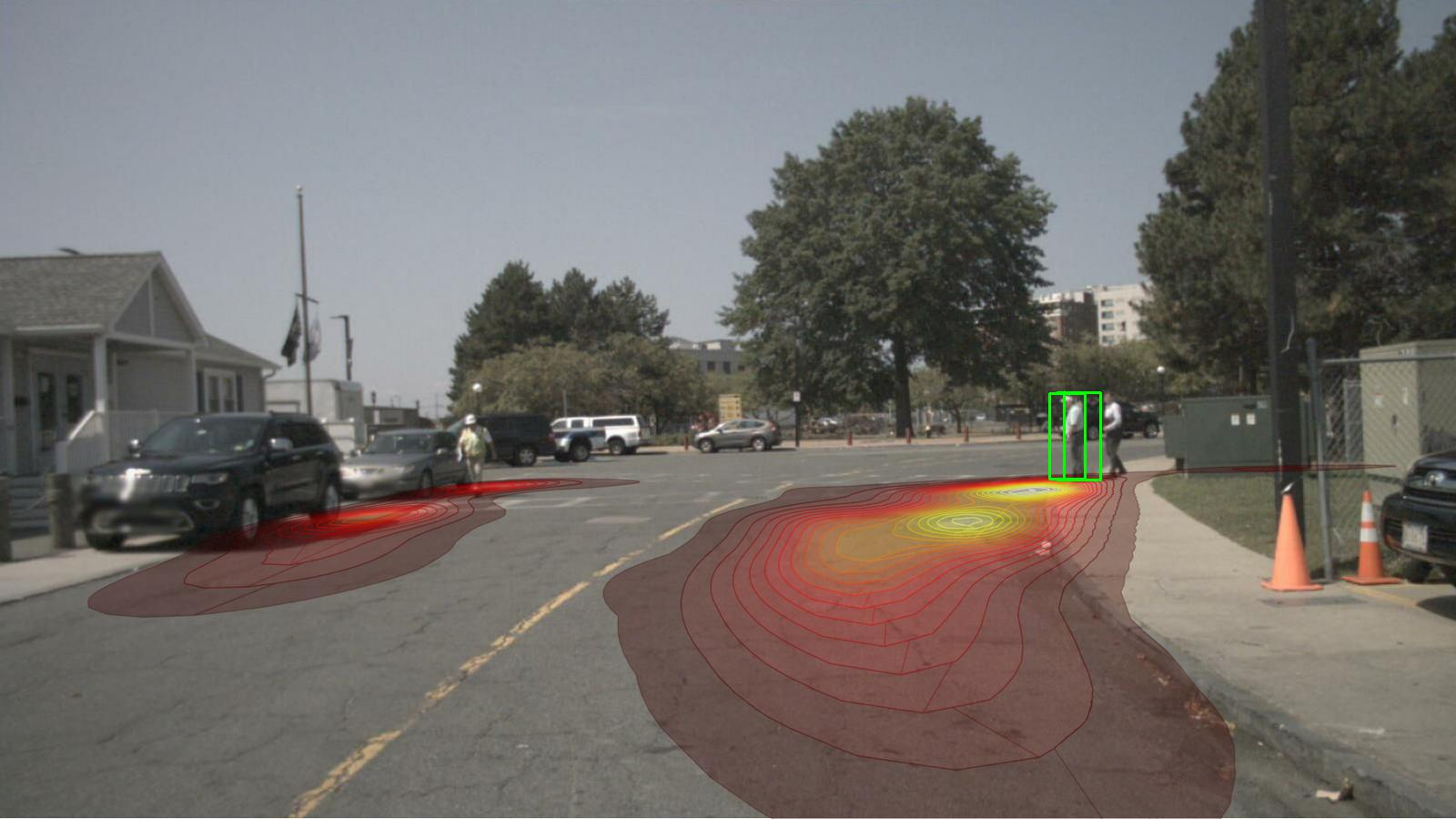}}}
    \subfloat[\gls{destination_predictor} - No referred object information - Top-down]{{\includegraphics[width=0.45\linewidth]{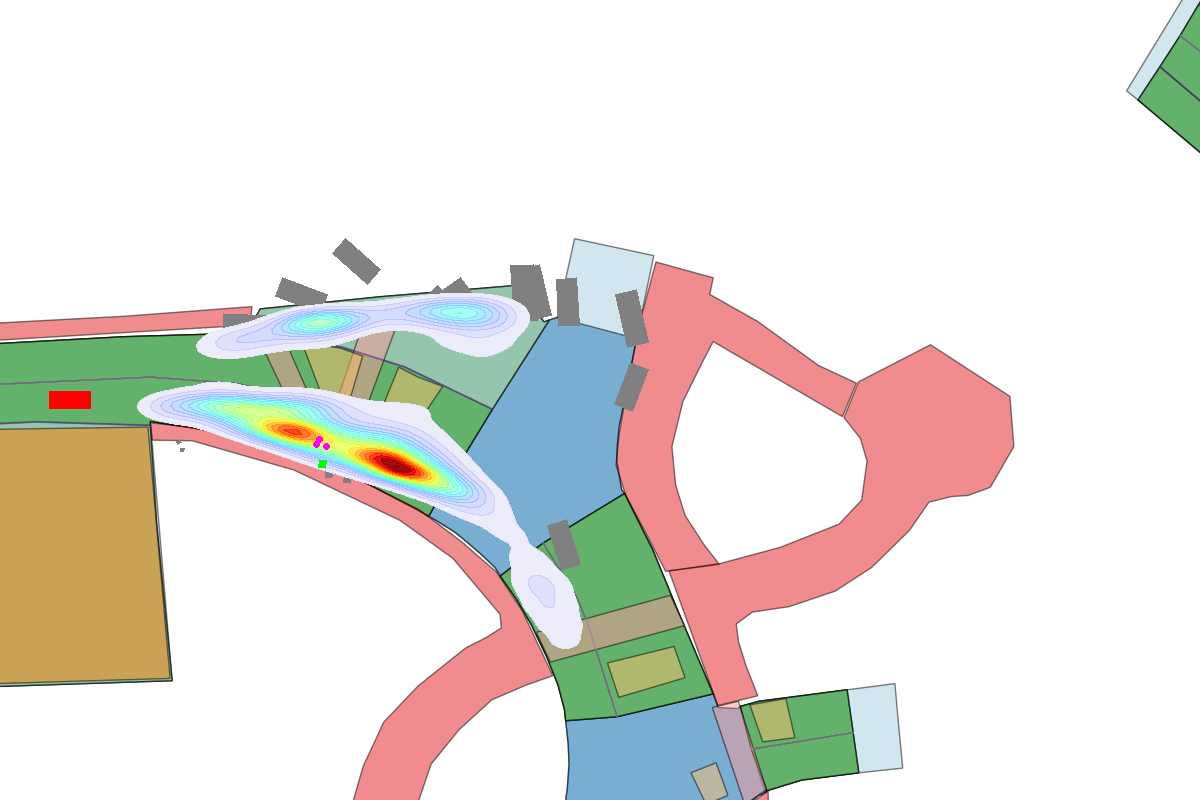}}}
    \qquad
      \subfloat[\gls{destination_predictor} - With referred object - Frontal]{{\includegraphics[width=0.45\linewidth]{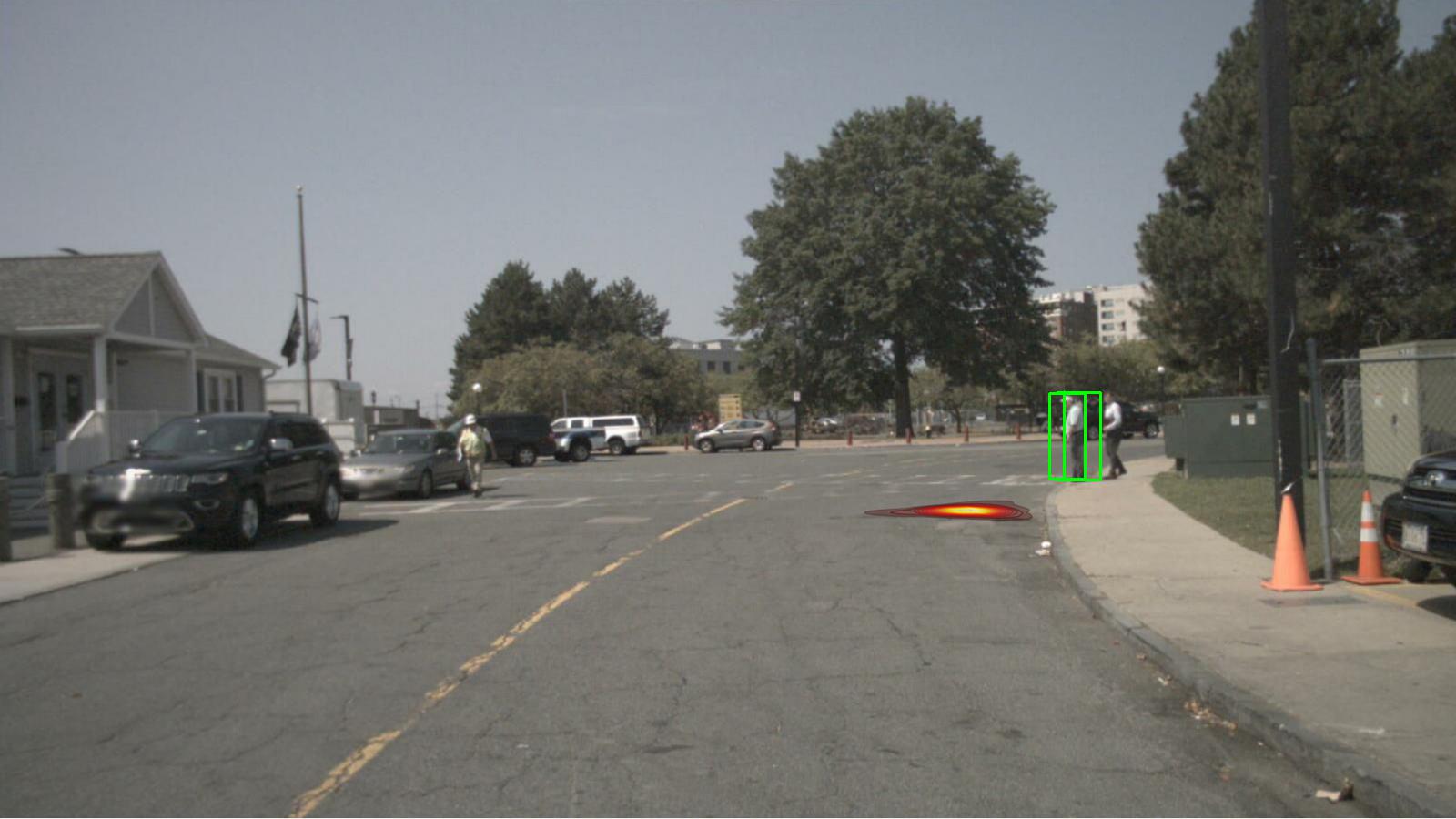} }}
    \subfloat[\gls{destination_predictor} - With referred object - Top-down]{{\includegraphics[width=0.45\linewidth]{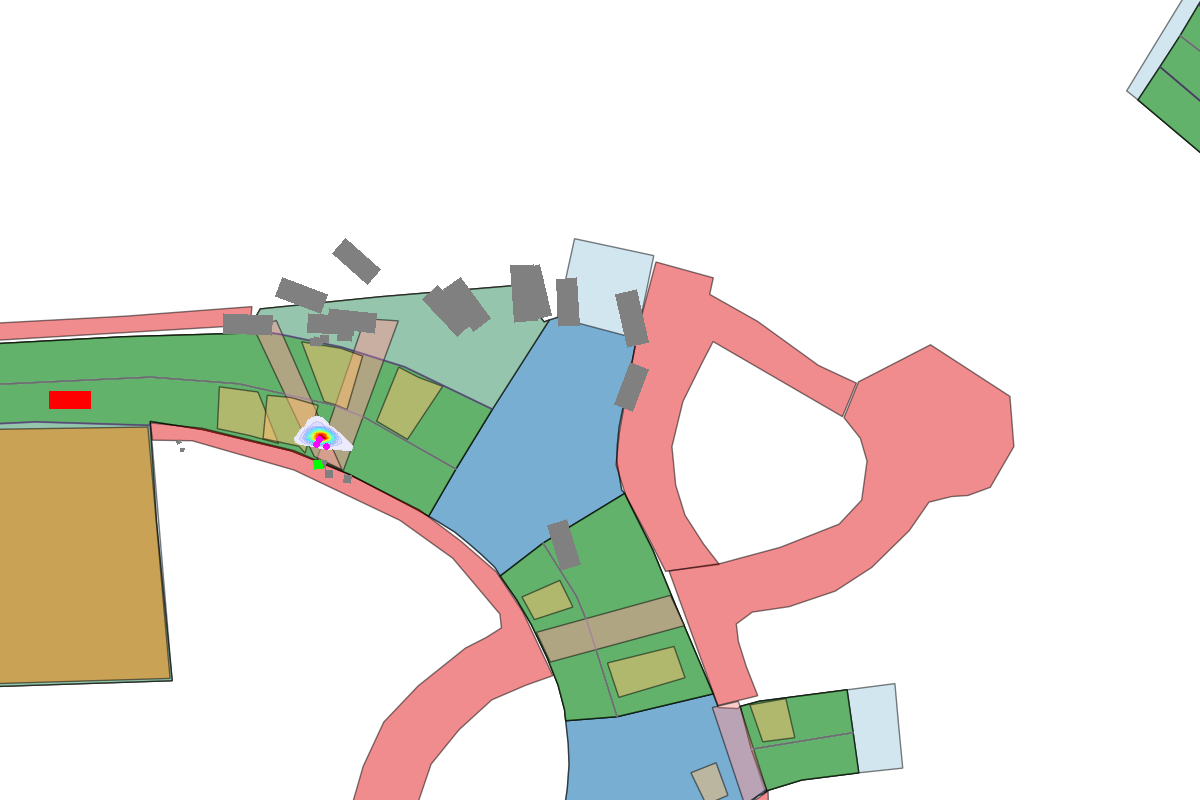}}}
    \caption{The issued command was: ``The man standing closest to the road is coming with us. Park next to him''. This image shows the heatmaps of the model that does not use the information of the referred object prediction (top row) versus the model that uses this information (bottom row). The red car on the top-down view is the ego car. The purple dots are the ground truth destination. The gray boxes represent other detected objects. The green rectangle in the top-down and in the frontal view indicates the referred object. In the top row, the model outputs heatmaps at different locations with increasing probability near objects of the same type as the referred object. In contrast, the model in the bottom row can pinpoint where it needs to output its heatmap as it has information about the referred object.}%
    \label{fig:ref_vs_no_ref}%
\end{figure*}

\subsubsection{Influence of Referred Object for Destination Prediction}
\label{subsubsect:exp_ref_vs_no_ref}
\begin{table}
    \centering
    \begin{adjustbox}{width=\linewidth}
    \begin{tabular}{c||c|c|c|c}
    \diaghead{
        \theadfont Diag Column}%
          {Method}{Metric} & ADE [m] & MDE [m] & PA$_2$[\%] & PA$_4$ [\%]  \\
      \hline 
      \hline
    \gls{destination_predictor} (Base) - NoRef & $13.95 \pm 0.40$ & $11.15$ & $11.25 \pm 0.52$ & $25.19 \pm 0.94$ \\
    \gls{destination_predictor} (Base) - Ref & $\mathbf{8.21 \pm 0.36}$ & $\mathbf{5.05}$ & $\mathbf{28.45 \pm 1.28}$ & $\mathbf{47.56 \pm 1.52}$
    \end{tabular}
    \end{adjustbox}
    \caption{Comparison %
    of our \gls{destination_predictor} %
    with (Ref) and without (NoRef) the %
    referred object prediction as additional input.}
\label{tab:ref_obj_inf_paper}
\end{table}

It can be seen in Table~\ref{tab:ref_obj_inf_paper} that the referred object information is crucial for the model's ability to predict the destination accurately. 
In Figure \ref{fig:ref_vs_no_ref} we see the difference between the presence and absence of referred object information during training. As the command refers to a person, the model that does not have the referred object information, and for which the referred object was encoded as just another object of its particular class in the input layout tensor, tends to predict high probabilities near all persons on the top-down view. The model that does have the referred object information available, on the other hand, manages to pinpoint the correct destination in front of the man on the right, standing closest to the road. In general, we see that the model learns to adhere to very general traffic rules, keep to the road, and perform rudimentary commands without the referred object information. However, with the referred object information, it manages to pinpoint the correct destination.

\section{Conclusion}
In this paper, we propose a new challenging spatial language understanding extension to Talk2Car called Talk2Car-Destination, where a model needs to predict the ego car's destination in the physical 3D world after receiving a passenger command, i.e., ``Park next to the man on the bench''. 
To handle these types of commands, a referred object detector is also required. Our proposed referred object detector outperforms some of the state-of-the-art models on Talk2Car while using considerably fewer parameters.
We also show that our Talk2Car-Destination task is not trivial by evaluating multiple baselines and (modified) existing state-of-the-art models. 
Hence, we believe this new dataset can be used as a benchmark for exciting future research. 
Additionally, we propose the \acrfull{destination_predictor}, which predicts distribution parameters at each feature map location at different resolution scales, and aggregates them into a distribution mixture. \gls{destination_predictor} achieves a relative increase in performance of 32\% for $PA_2$ and 15\% on $PA_4$ and 16\% in MDE.
In this work, we focused on the top-down view alone. However, in future work, one could investigate incorporating the visual information from the camera view. Furthermore, one could also examine whether jointly predicting the referred object and destination in a multi-task end-to-end fashion can yield benefits to the individual tasks.

\section{Acknowledgements}
This project was supported by Flanders AI, and the MACCHINA project from KU Leuven (grant number C14/18/065).
We also wish to thank NVIDIA for providing us with two RTX Titan XPs. 

\clearpage

\bibliography{references}

\begin{thebibliography}{56}
\providecommand{\natexlab}[1]{#1}

\bibitem[{Akbari et~al.(2019)Akbari, Karaman, Bhargava, Chen, Vondrick, and
  Chang}]{akbari2019multi}
Akbari, H.; Karaman, S.; Bhargava, S.; Chen, B.; Vondrick, C.; and Chang, S.-F.
  2019.
\newblock Multi-level multimodal common semantic space for image-phrase
  grounding.
\newblock In \emph{Proceedings of the IEEE/CVF Conference on Computer Vision
  and Pattern Recognition}, 12476--12486.

\bibitem[{Alahi et~al.(2016)Alahi, Goel, Ramanathan, Robicquet, Fei-Fei, and
  Savarese}]{alahi2016social}
Alahi, A.; Goel, K.; Ramanathan, V.; Robicquet, A.; Fei-Fei, L.; and Savarese,
  S. 2016.
\newblock Social lstm: Human trajectory prediction in crowded spaces.
\newblock In \emph{Proceedings of the IEEE Conference on Computer Vision and
  Pattern Recognition}, 961--971.

\bibitem[{Anderson et~al.(2018)Anderson, He, Buehler, Teney, Johnson, Gould,
  and Zhang}]{anderson2018bottom}
Anderson, P.; He, X.; Buehler, C.; Teney, D.; Johnson, M.; Gould, S.; and
  Zhang, L. 2018.
\newblock Bottom-up and top-down attention for image captioning and visual
  question answering.
\newblock In \emph{Proceedings of the IEEE Conference on Computer Vision and
  Pattern Recognition}, 6077--6086.

\bibitem[{Bowman et~al.(2015)Bowman, Angeli, Potts, and
  Manning}]{bowman2015large}
Bowman, S.~R.; Angeli, G.; Potts, C.; and Manning, C.~D. 2015.
\newblock A large annotated corpus for learning natural language inference.
\newblock \emph{arXiv preprint arXiv:1508.05326}.

\bibitem[{Caesar et~al.(2020)Caesar, Bankiti, Lang, Vora, Liong, Xu, Krishnan,
  Pan, Baldan, and Beijbom}]{caesar2020nuscenes}
Caesar, H.; Bankiti, V.; Lang, A.~H.; Vora, S.; Liong, V.~E.; Xu, Q.; Krishnan,
  A.; Pan, Y.; Baldan, G.; and Beijbom, O. 2020.
\newblock nuscenes: A multimodal dataset for autonomous driving.
\newblock In \emph{Proceedings of the IEEE/CVF Conference on Computer Vision
  and Pattern Recognition}, 11621--11631.

\bibitem[{Chen et~al.(2019)Chen, Suhr, Misra, Snavely, and
  Artzi}]{chen2019touchdown}
Chen, H.; Suhr, A.; Misra, D.; Snavely, N.; and Artzi, Y. 2019.
\newblock Touchdown: Natural language navigation and spatial reasoning in
  visual street environments.
\newblock In \emph{Proceedings of the IEEE/CVF Conference on Computer Vision
  and Pattern Recognition}, 12538--12547.

\bibitem[{Chen et~al.(2018)Chen, Rubanova, Bettencourt, and
  Duvenaud}]{chen2018neural}
Chen, R.~T.; Rubanova, Y.; Bettencourt, J.; and Duvenaud, D. 2018.
\newblock Neural ordinary differential equations.
\newblock \emph{arXiv preprint arXiv:1806.07366}.

\bibitem[{Chen et~al.(2015)Chen, Kundu, Zhu, Berneshawi, Ma, Fidler, and
  Urtasun}]{chen20153d}
Chen, X.; Kundu, K.; Zhu, Y.; Berneshawi, A.~G.; Ma, H.; Fidler, S.; and
  Urtasun, R. 2015.
\newblock 3d object proposals for accurate object class detection.
\newblock In \emph{Advances in Neural Information Processing Systems},
  424--432. Citeseer.

\bibitem[{Collell, Deruyttere, and Moens(2021)}]{collell2021probing}
Collell, G.; Deruyttere, T.; and Moens, M.-F. 2021.
\newblock Probing Spatial Clues: Canonical Spatial Templates for Object
  Relationship Understanding.
\newblock \emph{Ieee Access}, 9: 134298--134318.

\bibitem[{Dai et~al.(2020)Dai, Luo, Ding, and Shao}]{stacked_vlbert}
Dai, H.; Luo, S.; Ding, Y.; and Shao, L. 2020.
\newblock Commands for Autonomous Vehicles by Progressively Stacking
  Visual-Linguistic Representations.
\newblock In \emph{Proceedings of the 16th European Conference on Computer
  Vision, 2020. Commands for Autonomous Vehicles (C4AV) ECCV Workshop}.

\bibitem[{Dendorfer, Osep, and Leal-Taix{\'e}(2020)}]{dendorfer2020goal}
Dendorfer, P.; Osep, A.; and Leal-Taix{\'e}, L. 2020.
\newblock Goal-GAN: Multimodal Trajectory Prediction Based on Goal Position
  Estimation.
\newblock In \emph{Proceedings of the Asian Conference on Computer Vision}.

\bibitem[{Deng et~al.(2018)Deng, Wu, Wu, Hu, Lyu, and Tan}]{deng2018visual}
Deng, C.; Wu, Q.; Wu, Q.; Hu, F.; Lyu, F.; and Tan, M. 2018.
\newblock Visual grounding via accumulated attention.
\newblock In \emph{Proceedings of the IEEE Conference on Computer Vision and
  Pattern Recognition}, 7746--7755.

\bibitem[{Deruyttere, Milewski, and Moens(2021)}]{deruyttere2021giving}
Deruyttere, T.; Milewski, V.; and Moens, M.-F. 2021.
\newblock Giving commands to a self-driving car: How to deal with uncertain
  situations?
\newblock \emph{Engineering Applications of Artificial Intelligence}, 103:
  104257.

\bibitem[{Deruyttere et~al.(2020)Deruyttere, Vandenhende, Grujicic, Liu,
  Van~Gool, Blaschko, Tuytelaars, and Moens}]{deruyttere2020commands}
Deruyttere, T.; Vandenhende, S.; Grujicic, D.; Liu, Y.; Van~Gool, L.; Blaschko,
  M.; Tuytelaars, T.; and Moens, M.-F. 2020.
\newblock Commands 4 autonomous vehicles (c4av) workshop summary.
\newblock In \emph{European Conference on Computer Vision}, 3--26. Springer.

\bibitem[{Deruyttere et~al.(2019)Deruyttere, Vandenhende, Grujicic, Van~Gool,
  and Moens}]{deruyttere2019talk2car}
Deruyttere, T.; Vandenhende, S.; Grujicic, D.; Van~Gool, L.; and Moens, M.~F.
  2019.
\newblock Talk2Car: Taking Control of Your Self-Driving Car.
\newblock In \emph{Proceedings of the 2019 Conference on Empirical Methods in
  Natural Language Processing and the 9th International Joint Conference on
  Natural Language Processing (EMNLP-IJCNLP)}, 2088--2098.

\bibitem[{Dosovitskiy et~al.(2015)Dosovitskiy, Fischer, Ilg, Hausser, Hazirbas,
  Golkov, Van Der~Smagt, Cremers, and Brox}]{dosovitskiy2015flownet}
Dosovitskiy, A.; Fischer, P.; Ilg, E.; Hausser, P.; Hazirbas, C.; Golkov, V.;
  Van Der~Smagt, P.; Cremers, D.; and Brox, T. 2015.
\newblock Flownet: Learning optical flow with convolutional networks.
\newblock In \emph{Proceedings of the IEEE International Conference on Computer
  Vision}, 2758--2766.

\bibitem[{Du et~al.(2021)Du, Fu, Liu, and Wang}]{du2021visual}
Du, Y.; Fu, Z.; Liu, Q.; and Wang, Y. 2021.
\newblock Visual Grounding with Transformers.
\newblock \emph{arXiv preprint arXiv:2105.04281}.

\bibitem[{Engelcke et~al.(2017)Engelcke, Rao, Wang, Tong, and
  Posner}]{engelcke2017vote3deep}
Engelcke, M.; Rao, D.; Wang, D.~Z.; Tong, C.~H.; and Posner, I. 2017.
\newblock Vote3deep: Fast object detection in 3d point clouds using efficient
  convolutional neural networks.
\newblock In \emph{2017 IEEE International Conference on Robotics and
  Automation (ICRA)}, 1355--1361. IEEE.

\bibitem[{Grujicic et~al.(2020)Grujicic, Radevski, Tuytelaars, and
  Blaschko}]{grujicic2020learning}
Grujicic, D.; Radevski, G.; Tuytelaars, T.; and Blaschko, M. 2020.
\newblock Learning to ground medical text in a 3D human atlas.
\newblock In \emph{Proceedings of the 24th Conference on Computational Natural
  Language Learning}, 302--312.

\bibitem[{Guzman-Rivera, Batra, and Kohli(2012)}]{guzman2012multiple}
Guzman-Rivera, A.; Batra, D.; and Kohli, P. 2012.
\newblock Multiple Choice Learning: Learning to Produce Multiple Structured
  Outputs.
\newblock In \emph{NIPS}, volume~1, 3. Citeseer.

\bibitem[{He et~al.(2016)He, Zhang, Ren, and Sun}]{he2016deep}
He, K.; Zhang, X.; Ren, S.; and Sun, J. 2016.
\newblock Deep residual learning for image recognition.
\newblock In \emph{Proceedings of the IEEE Conference on Computer Vision and
  Pattern Recognition}, 770--778.

\bibitem[{Hu et~al.(2016)Hu, Xu, Rohrbach, Feng, Saenko, and
  Darrell}]{hu2016natural}
Hu, R.; Xu, H.; Rohrbach, M.; Feng, J.; Saenko, K.; and Darrell, T. 2016.
\newblock Natural language object retrieval.
\newblock In \emph{Proceedings of the IEEE Conference on Computer Vision and
  Pattern Recognition}, 4555--4564.

\bibitem[{Hudson and Manning(2018)}]{Hudson2018}
Hudson, D.~A.; and Manning, C.~D. 2018.
\newblock Compositional Attention Networks for Machine Reasoning.
\newblock \emph{CoRR}, abs/1803.03067.

\bibitem[{Kamath et~al.(2021)Kamath, Singh, LeCun, Misra, Synnaeve, and
  Carion}]{kamath2021mdetr}
Kamath, A.; Singh, M.; LeCun, Y.; Misra, I.; Synnaeve, G.; and Carion, N. 2021.
\newblock MDETR--Modulated Detection for End-to-End Multi-Modal Understanding.
\newblock \emph{arXiv preprint arXiv:2104.12763}.

\bibitem[{Krishna(2019)}]{krishna2019easyturk}
Krishna, R. 2019.
\newblock EasyTurk: A Wrapper for Custom AMT Tasks.
\newblock \url{https://github.com/ranjaykrishna/easyturk}.

\bibitem[{Lang et~al.(2019)Lang, Vora, Caesar, Zhou, Yang, and
  Beijbom}]{lang2019pointpillars}
Lang, A.~H.; Vora, S.; Caesar, H.; Zhou, L.; Yang, J.; and Beijbom, O. 2019.
\newblock Pointpillars: Fast encoders for object detection from point clouds.
\newblock In \emph{Proceedings of the IEEE/CVF Conference on Computer Vision
  and Pattern Recognition}, 12697--12705.

\bibitem[{Lee et~al.(2017)Lee, Choi, Vernaza, Choy, Torr, and
  Chandraker}]{lee2017desire}
Lee, N.; Choi, W.; Vernaza, P.; Choy, C.~B.; Torr, P.~H.; and Chandraker, M.
  2017.
\newblock Desire: Distant future prediction in dynamic scenes with interacting
  agents.
\newblock In \emph{Proceedings of the IEEE Conference on Computer Vision and
  Pattern Recognition}, 336--345.

\bibitem[{Liang et~al.(2020)Liang, Jiang, Murphy, Yu, and
  Hauptmann}]{liang2020garden}
Liang, J.; Jiang, L.; Murphy, K.; Yu, T.; and Hauptmann, A. 2020.
\newblock The garden of forking paths: Towards multi-future trajectory
  prediction.
\newblock In \emph{Proceedings of the IEEE/CVF Conference on Computer Vision
  and Pattern Recognition}, 10508--10518.

\bibitem[{Liu et~al.(2018)Liu, Sharma, Camps, and Sznaier}]{liu2018dyan}
Liu, W.; Sharma, A.; Camps, O.; and Sznaier, M. 2018.
\newblock Dyan: A dynamical atoms-based network for video prediction.
\newblock In \emph{Proceedings of the European Conference on Computer Vision
  (ECCV)}, 170--185.

\bibitem[{Lourentzou, Morales, and Zhai(2017)}]{lourentzou2017text}
Lourentzou, I.; Morales, A.; and Zhai, C. 2017.
\newblock Text-based geolocation prediction of social media users with neural
  networks.
\newblock In \emph{2017 IEEE International Conference on Big Data (Big Data)},
  696--705. IEEE.

\bibitem[{Luc et~al.(2017)Luc, Neverova, Couprie, Verbeek, and
  LeCun}]{luc2017predicting}
Luc, P.; Neverova, N.; Couprie, C.; Verbeek, J.; and LeCun, Y. 2017.
\newblock Predicting deeper into the future of semantic segmentation.
\newblock In \emph{Proceedings of the IEEE International Conference on Computer
  Vision}, 648--657.

\bibitem[{Luo et~al.(2020)Luo, Dai, Shao, and Ding}]{luo2020c4av}
Luo, S.; Dai, H.; Shao, L.; and Ding, Y. 2020.
\newblock C4AV: Learning Cross-Modal Representations from Transformers.
\newblock In \emph{European Conference on Computer Vision}, 33--38. Springer.

\bibitem[{Makansi et~al.(2019)Makansi, Ilg, Cicek, and
  Brox}]{makansi2019overcoming}
Makansi, O.; Ilg, E.; Cicek, O.; and Brox, T. 2019.
\newblock Overcoming limitations of mixture density networks: A sampling and
  fitting framework for multimodal future prediction.
\newblock In \emph{Proceedings of the IEEE/CVF Conference on Computer Vision
  and Pattern Recognition}, 7144--7153.

\bibitem[{Mangalam et~al.(2020)Mangalam, Girase, Agarwal, Lee, Adeli, Malik,
  and Gaidon}]{mangalam2020not}
Mangalam, K.; Girase, H.; Agarwal, S.; Lee, K.-H.; Adeli, E.; Malik, J.; and
  Gaidon, A. 2020.
\newblock It is not the journey but the destination: Endpoint conditioned
  trajectory prediction.
\newblock In \emph{European Conference on Computer Vision}, 759--776. Springer.

\bibitem[{Messaoud et~al.(2020)Messaoud, Deo, Trivedi, and
  Nashashibi}]{messaoud2020trajectory}
Messaoud, K.; Deo, N.; Trivedi, M.~M.; and Nashashibi, F. 2020.
\newblock Trajectory prediction for autonomous driving based on multi-head
  attention with joint agent-map representation.
\newblock \emph{arXiv preprint arXiv:2005.02545}.

\bibitem[{Mohamed et~al.(2020)Mohamed, Qian, Elhoseiny, and
  Claudel}]{mohamed2020social}
Mohamed, A.; Qian, K.; Elhoseiny, M.; and Claudel, C. 2020.
\newblock Social-stgcnn: A social spatio-temporal graph convolutional neural
  network for human trajectory prediction.
\newblock In \emph{Proceedings of the IEEE/CVF Conference on Computer Vision
  and Pattern Recognition}, 14424--14432.

\bibitem[{Narayanan et~al.(2021)Narayanan, Moslemi, Pittaluga, Liu, and
  Chandraker}]{narayanan2021divide}
Narayanan, S.; Moslemi, R.; Pittaluga, F.; Liu, B.; and Chandraker, M. 2021.
\newblock Divide-and-Conquer for Lane-Aware Diverse Trajectory Prediction.
\newblock In \emph{Proceedings of the IEEE/CVF Conference on Computer Vision
  and Pattern Recognition}, 15799--15808.

\bibitem[{Othman(2021)}]{othman2021public}
Othman, K. 2021.
\newblock Public acceptance and perception of autonomous vehicles: a
  comprehensive review.
\newblock \emph{AI and Ethics}, 1--33.

\bibitem[{Reimers and Gurevych(2019)}]{reimers2019sentence}
Reimers, N.; and Gurevych, I. 2019.
\newblock Sentence-bert: Sentence embeddings using siamese bert-networks.
\newblock \emph{arXiv preprint arXiv:1908.10084}.

\bibitem[{Roddick, Kendall, and Cipolla(2018)}]{roddick2018orthographic}
Roddick, T.; Kendall, A.; and Cipolla, R. 2018.
\newblock Orthographic feature transform for monocular 3d object detection.
\newblock \emph{arXiv preprint arXiv:1811.08188}.

\bibitem[{Rodriguez, Fernando, and Li(2018)}]{rodriguez2018action}
Rodriguez, C.; Fernando, B.; and Li, H. 2018.
\newblock Action anticipation by predicting future dynamic images.
\newblock In \emph{Proceedings of the European Conference on Computer Vision
  (ECCV) Workshops}, 0--0.

\bibitem[{Rufus et~al.(2021)Rufus, Jain, Nair, Gandhi, and
  Krishna}]{rufus2021grounding}
Rufus, N.; Jain, K.; Nair, U. K.~R.; Gandhi, V.; and Krishna, K.~M. 2021.
\newblock Grounding Linguistic Commands to Navigable Regions.
\newblock In \emph{2021 IEEE/RSJ International Conference on Intelligent Robots
  and Systems (IROS)}.

\bibitem[{Rufus et~al.(2020)Rufus, Nair, Krishna, and Gandhi}]{rufus2020cosine}
Rufus, N.; Nair, U. K.~R.; Krishna, K.~M.; and Gandhi, V. 2020.
\newblock Cosine meets softmax: a tough-to-beat baseline for visual grounding.
\newblock In \emph{European Conference on Computer Vision}, 39--50. Springer.

\bibitem[{Schoettle and Sivak(2014)}]{schoettle2014survey}
Schoettle, B.; and Sivak, M. 2014.
\newblock A survey of public opinion about autonomous and self-driving vehicles
  in the US, the UK, and Australia.
\newblock Technical report, University of Michigan, Ann Arbor, Transportation
  Research Institute.

\bibitem[{Sriram et~al.(2019)Sriram, Maniar, Kalyanasundaram, Gandhi, Bhowmick,
  and Krishna}]{sriram2019talk}
Sriram, N.; Maniar, T.; Kalyanasundaram, J.; Gandhi, V.; Bhowmick, B.; and
  Krishna, K.~M. 2019.
\newblock Talk to the Vehicle: Language Conditioned Autonomous Navigation of
  Self Driving Cars.
\newblock In \emph{2019 IEEE/RSJ International Conference on Intelligent Robots
  and Systems (IROS)}, 5284--5290.

\bibitem[{Vasudevan, Dai, and Van~Gool(2021)}]{vasudevan2021talk2nav}
Vasudevan, A.~B.; Dai, D.; and Van~Gool, L. 2021.
\newblock Talk2nav: Long-range vision-and-language navigation with dual
  attention and spatial memory.
\newblock \emph{International Journal of Computer Vision}, 129(1): 246--266.

\bibitem[{Vora et~al.(2020)Vora, Lang, Helou, and
  Beijbom}]{vora2020pointpainting}
Vora, S.; Lang, A.~H.; Helou, B.; and Beijbom, O. 2020.
\newblock Pointpainting: Sequential fusion for 3d object detection.
\newblock In \emph{Proceedings of the IEEE/CVF Conference on Computer Vision
  and Pattern Recognition}, 4604--4612.

\bibitem[{Wang et~al.(2021)Wang, Zhu, Pang, and Lin}]{wang2021fcos3d}
Wang, T.; Zhu, X.; Pang, J.; and Lin, D. 2021.
\newblock FCOS3D: Fully Convolutional One-Stage Monocular 3D Object Detection.
\newblock \emph{arXiv preprint arXiv:2104.10956}.

\bibitem[{Williams, Nangia, and Bowman(2017)}]{williams2017broad}
Williams, A.; Nangia, N.; and Bowman, S.~R. 2017.
\newblock A broad-coverage challenge corpus for sentence understanding through
  inference.
\newblock \emph{arXiv preprint arXiv:1704.05426}.

\bibitem[{Yagi et~al.(2018)Yagi, Mangalam, Yonetani, and Sato}]{yagi2018future}
Yagi, T.; Mangalam, K.; Yonetani, R.; and Sato, Y. 2018.
\newblock Future person localization in first-person videos.
\newblock In \emph{Proceedings of the IEEE Conference on Computer Vision and
  Pattern Recognition}, 7593--7602.

\bibitem[{Yin, Zhou, and Kr{\"a}henb{\"u}hl(2021)}]{yin2021center}
Yin, T.; Zhou, X.; and Kr{\"a}henb{\"u}hl, P. 2021.
\newblock Center-based 3D Object Detection and Tracking.
\newblock \emph{Conference on Computer Vision and Pattern Recognition}.

\bibitem[{Yu et~al.(2018)Yu, Lin, Shen, Yang, Lu, Bansal, and
  Berg}]{yu2018mattnet}
Yu, L.; Lin, Z.; Shen, X.; Yang, J.; Lu, X.; Bansal, M.; and Berg, T.~L. 2018.
\newblock Mattnet: Modular attention network for referring expression
  comprehension.
\newblock In \emph{Proceedings of the IEEE Conference on Computer Vision and
  Pattern Recognition}, 1307--1315.

\bibitem[{Zheng et~al.(2021)Zheng, Tang, Jiang, and Fu}]{zheng2021se}
Zheng, W.; Tang, W.; Jiang, L.; and Fu, C.-W. 2021.
\newblock SE-SSD: Self-Ensembling Single-Stage Object Detector From Point
  Cloud.
\newblock In \emph{Proceedings of the IEEE/CVF Conference on Computer Vision
  and Pattern Recognition}, 14494--14503.

\bibitem[{Zhou, Wang, and Kr{\"a}henb{\"u}hl(2019)}]{zhou2019objects}
Zhou, X.; Wang, D.; and Kr{\"a}henb{\"u}hl, P. 2019.
\newblock Objects as points.
\newblock \emph{arXiv preprint arXiv:1904.07850}.

\bibitem[{Zhou and Tuzel(2018)}]{zhou2018voxelnet}
Zhou, Y.; and Tuzel, O. 2018.
\newblock Voxelnet: End-to-end learning for point cloud based 3d object
  detection.
\newblock In \emph{Proceedings of the IEEE Conference on Computer Vision and
  Pattern Recognition}, 4490--4499.

\bibitem[{Zieba et~al.(2020)Zieba, Przewiezlikowski, Smieja, Tabor, Trzcinski,
  and Spurek}]{zikeba2020regflow}
Zieba, M.; Przewiezlikowski, M.; Smieja, M.; Tabor, J.; Trzcinski, T.; and
  Spurek, P. 2020.
\newblock RegFlow: Probabilistic Flow-based Regression for Future Prediction.
\newblock \emph{arXiv preprint arXiv:2011.14620}.

\end{thebibliography}

\clearpage
\iftoggle{arxiv}{
\appendix
\counterwithin{figure}{section}
\counterwithin{table}{section}

\section*{Appendix}
Here we provide additional information about the dataset, a comparison between 3D object detectors, additional results of the evaluated models and also some qualitative results of the predictions.

\section{Dataset}
\label{app:dataset}
During annotation of Talk2Car-Destination, we provided annotators with a command from Talk2Car and its accompanying frontal view, top-down view, and the 3D bounding ground truth boxes of objects from nuScenes (see Figure \ref{fig:annotation}).
Additionally, we provided the annotators with a fixed list of intent categories they could choose from.
Table \ref{tab:intent_distribution} shows this list of intents together with the distribution of the intents over the three splits of Talk2Car-Destination.
We observe that the distribution of these intents is similar across the splits. 
We also observe that ``Park'' and ``Stop'' are the most commonly indicated intent categories, while the ``U-Turn Right`` and ``U-Turn Left'' are the least common.
During annotation, we also provided an ``other'' category for the intent. However, from Table \ref{tab:intent_distribution} we see that annotators chose this option in fewer than 0.6\% annotations.

\begin{table}[h]
    \centering
    \begin{tabular}{l||c|c|c}
\diaghead{
\theadfont Diag Column}%
  {Intent}{Split}&
\thead{Train}&\thead{Validation}&\thead{Test}\\
\hline \hline 
Turn Left&8.18\%&9.40\%&7.95\% \\
Turn Right&6.93\%&5.35\%&6.15\%\\
Change Lane Left&2.94\%&2.85\%&3.12\%\\
Change Lane Right&2.23\%&4.14\%&2.46\%\\
U-Turn Left&1.59\%&1.04\%&1.48\%\\
U-Turn Right&0.80\%&0.78\%&1.27\%\\
Park&19.52\%&18.64\%&18.82\%\\
Stop&19.12\%&18.72\%&19.23\%\\
Pick Up&2.89\%&2.85\%&3.49\%\\
Continue&4.00\%&4.40\%&4.59\%\\
Overtake&2.40\%&3.19\%&2.42\%\\
Drop Off&3.29\%&3.36\%&3.36\%\\
Follow&10.96\%&10.35\%&10.41\%\\
Slow Down&6.89\%&6.13\%&7.18\%\\
Wait&3.67\%&3.54\%&3.20\%\\
Approach&2.29\%&2.24\%&2.67\%\\
Move Away&1.94\%&2.42\%&1.72\%\\
Other&0.37\%&0.60\%&0.49\%\\
\hline
Total&100\%&100\%&100\%\\
\end{tabular}
    \caption{In the first column, we see the different intent categories in the Talk2Car-Destination dataset. We see how many commands in a particular split belong to a specific intent class in the three other columns. Overall we see the three splits follow the same intent distribution.}
    \label{tab:intent_distribution}
\end{table}

\section{3D Object Detector}
\label{app:3d_obj}
Both FCOS3D\footnote{https://github.com/open-mmlab/mmdetection3d} \cite{wang2021fcos3d} and CenterPoint\footnote{https://github.com/tianweiy/CenterPoint} \cite{yin2021center} were trained with the original hyperparameters provided by their respective authors. 
Both models had to be re-trained from scratch as the pre-trained checkpoints were trained on the nuScenes training set, which includes scenes from the Talk2Car test set.
Thus, to avoid any data leakage, we trained both models only on the scenes from the Talk2Car training set.
Some qualitative results from both models can be seen in Figure \ref{fig:fcos3d_vs_centerpoint_1}, \ref{fig:fcos3d_vs_centerpoint_2}, \ref{fig:fcos3d_vs_centerpoint_3}, and \ref{fig:fcos3d_vs_centerpoint_4}.

We perform the following two experiments to evaluate how well each model is suited for detecting the referred object from the Talk2Car dataset at various distances from the ego car.
For the first experiment, we measure the distance between the center of the ground truth bounding box of the referred object and the center of the nearest 3D predicted bounding box out of the top-64 scoring bounding boxes.
The second experiment measures the same distance between the ground truth bounding box of the referred object and the closest bounding box of the same class as the referred object from the top-64 scoring bounding boxes. 
We evaluate the average of such distances over six groups of commands, divided into six bins based on the distance between the referred object and the ego car.

The results from the first experiment are shown in Table \ref{tab:3d_detector_distance_bins_closest_any_box}.
We observe that CenterPoint has a better average distance for close objects (0-40m) than FCOS3D. However, from 40m and onward, we notice that FCOS3D becomes better than CenterPoint. This corresponds with what we see in Figure \ref{fig:fcos3d_vs_centerpoint_1} where CenterPoint stops making predictions after a certain distance. The results from the second experiment are shown in Table \ref{tab:3d_detector_distance_bins_closest_same_class_box}. Here we find that FCOS3D becomes better across all bins compared to CenterPoint, once the matching prediction of object class is also taken as a requirement. This indicates that CenterPoint has issues with correctly classifying objects, which is in line with the findings of \cite{vora2020pointpainting} where they show that two semantically different objects at a certain distance can appear very similar according to their point clouds, making it difficult for the LIDAR based method to distinguish between them.

Finally, FCOS3D can also be used to directly extract visual features of the image objects in parallel with predicting the 3D bounding boxes. 
With CenterPoint, the visual information from the image is not available in the point cloud, hence, requiring another model to extract these features. Doing so would, in our view, add unnecessary latency for a time-critical task. Therefore, we find FCOS3D to be the better choice for our particular setting.

\begin{table}[h]
    \centering
    \begin{tabular}{l||c|c}
     \diaghead{
        \theadfont Long Sentence For P}%
          {Distance Bins}{Model}&
        \thead{FCOS3D}&\thead{CenterPoint} \\
        \hline \hline 
        0 - 20m (44.85\%) & 1.371m & 1.095m \\
        20 - 40m (38.54\%) & 1.347m & 1.124m\\
        40-60m  (11.68\%) & 2.015m & 3.181m\\
        60-80m (3.85\%) & 2.384m & 22.888m\\
        80-100m (0.94\%) & 4.013m & 43.243m\\
        100-120m (0.12\%) & 3.830m & 51.426m\\
        \hline
        0-120m (100.00\%) & 1.504m & 2.649m\\
    \end{tabular}
    \caption{The minimum distance between the ground truth referred object and the closest top-64 scoring bounding box. Between parentheses, we indicate how many commands fall into a specific bin. The bins represent the distance between the center of the 3D referred object box and the ego car.}
    \label{tab:3d_detector_distance_bins_closest_any_box}
\end{table}

\begin{table}[h]
    \centering
    \begin{tabular}{l||c|c}
     \diaghead{
        \theadfont Long Sentence For P}%
          {Distance Bins}{Model}&
        \thead{FCOS3D}&\thead{CenterPoint} \\
        \hline \hline 
        0 - 20m (44.85\%) & 1.564m & 6.272m \\
        20 - 40m (38.54\%) & 1.507m & 6.159m\\
        40-60m  (11.68\%) & 2.397m & 12.688m\\
        60-80m (3.85\%) & 2.792m & 38.477m\\
        80-100m (0.94\%) & 4.121m & 51.367m\\
        100-120m (0.12\%) & 3.830m & 56.710m\\
        \hline
        0-120m (100.00\%) & 1.713m & 8.707m\\
    \end{tabular}
    \caption{The minimum distance between the ground truth referred object and the closest top-64 scoring bounding box of the same predicted class as the referred object. Between parentheses, we indicate how many commands fall into a specific bin. The bins represent the distance between the center of the 3D referred object box and the ego car.}
    \label{tab:3d_detector_distance_bins_closest_same_class_box}
\end{table}

\begin{figure*}%
    \centering
    \subfloat[\centering FCOS3D]{{\includegraphics[width=0.75\linewidth]{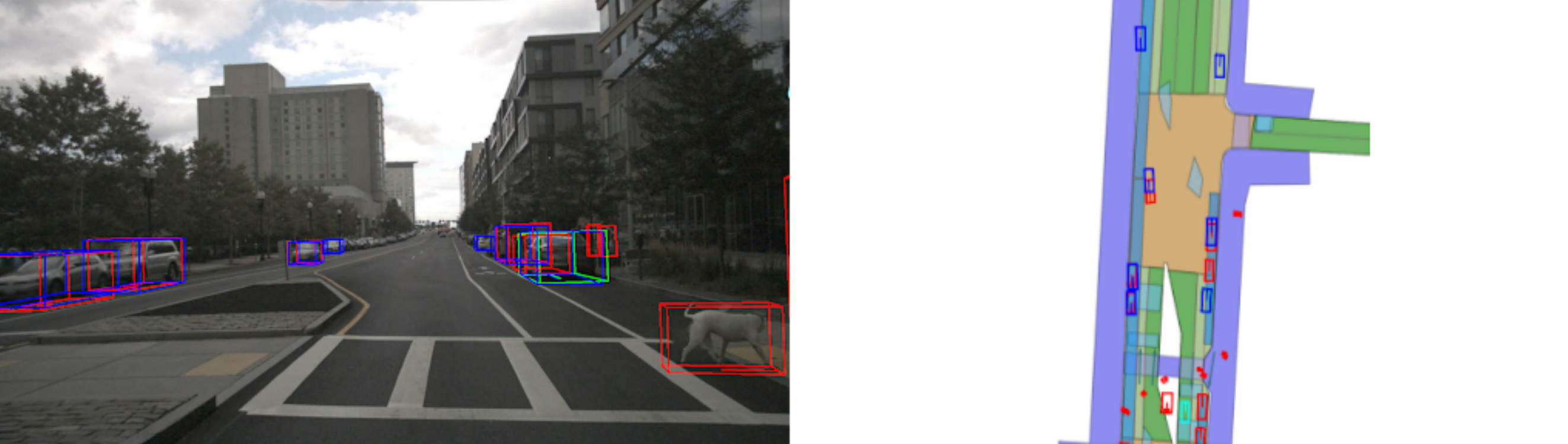} }}%
    \qquad
    \subfloat[\centering CenterPoint]{{\includegraphics[width=0.75\linewidth]{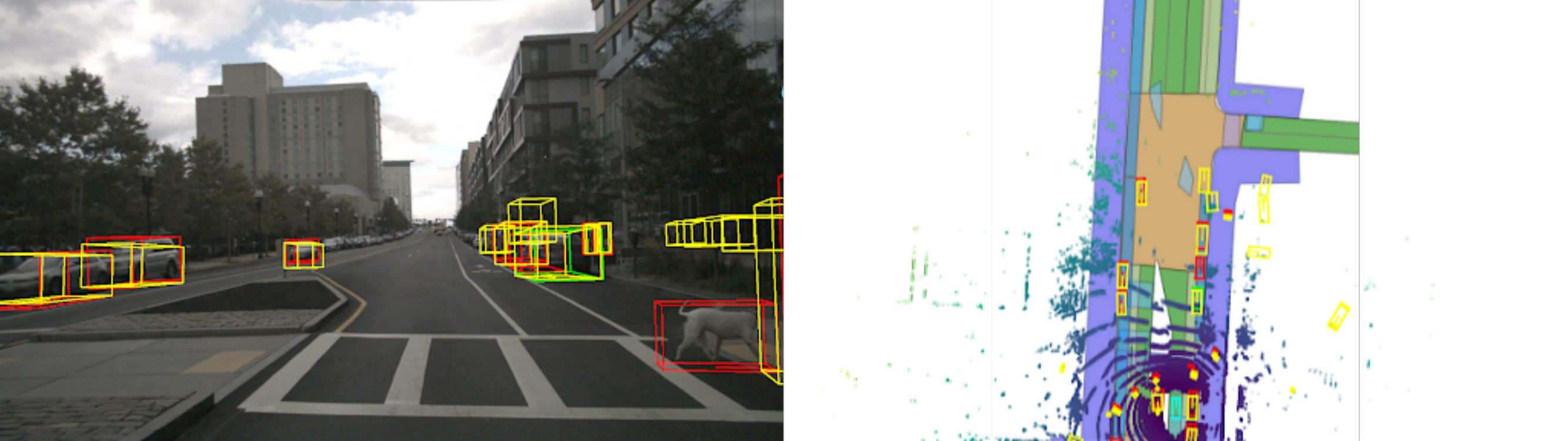} }}%
    \caption{Comparison of 3D object detection for FCOS3D and CenterPoint. The red bounding boxes on both images represent the ground truth detections, the green bounding box represents the referred object, while the FCOS3D and CenterPoint detections are shown in blue and yellow, respectively. It can be seen from the image that the FCOS3D detects objects at greater distances than CenterPoint.}%
    \label{fig:fcos3d_vs_centerpoint_1}%
\end{figure*}

\begin{figure*}%
    \centering
    \subfloat[\centering FCOS3D]{{\includegraphics[width=0.75\linewidth]{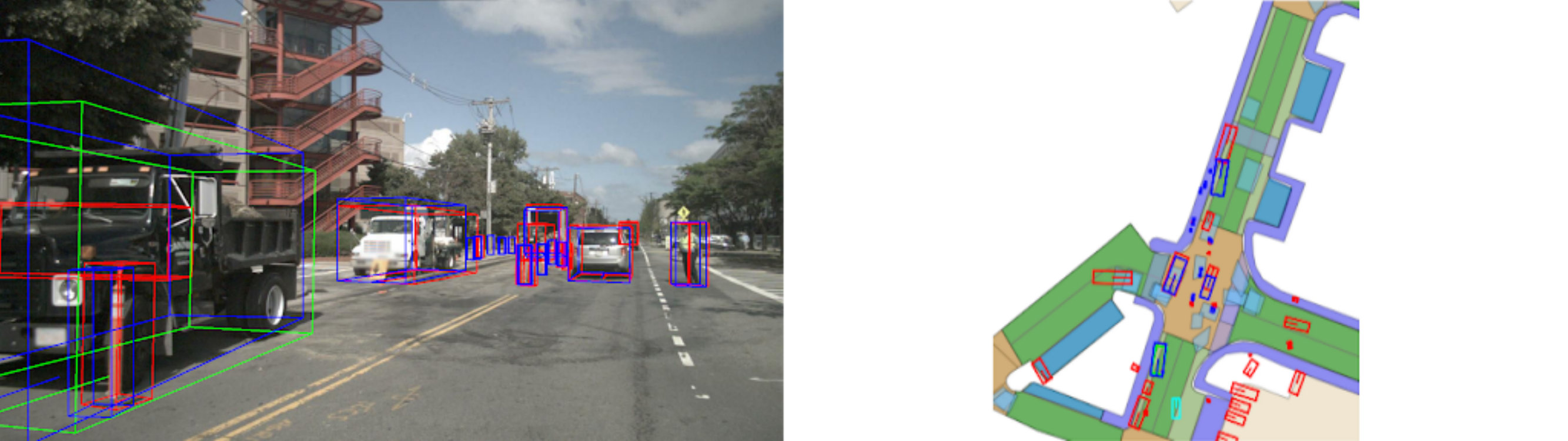} }}%
    \qquad
    \subfloat[\centering CenterPoint]{{\includegraphics[width=0.75\linewidth]{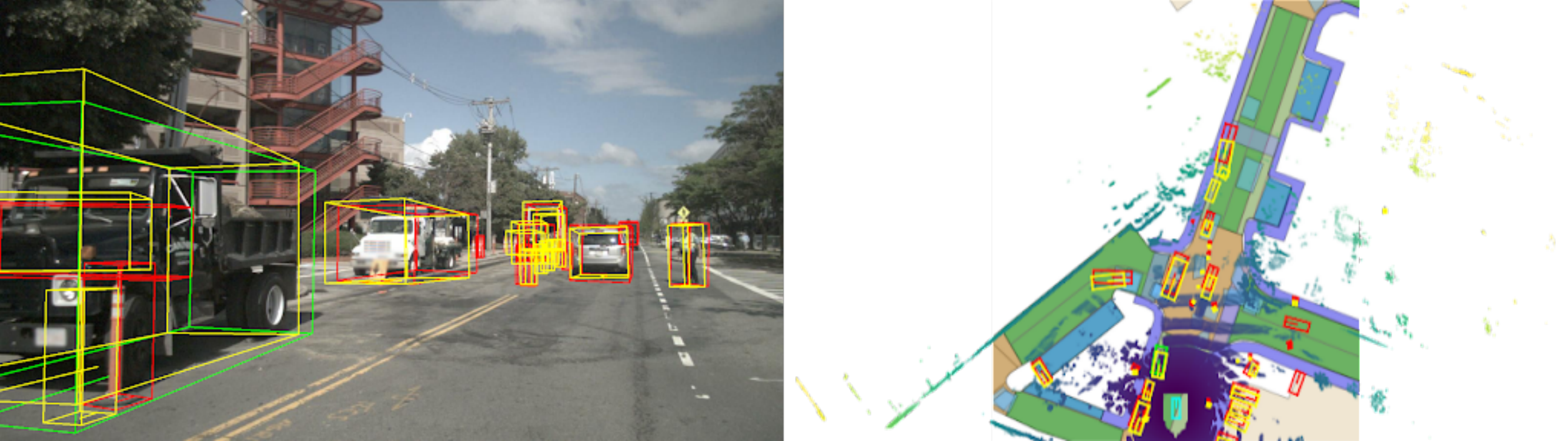} }}%
    \caption{Comparison of 3D object detection for FCOS3D and CenterPoint. The red bounding boxes on both images represent the ground truth detections, the green bounding box represents the referred object, while the FCOS3D and CenterPoint detections are shown in blue and yellow, respectively. It can be seen from the detected traffic cone on the bottom left that the projected bounding box from CenterPoint produces a bounding box that is shifted away from the actual object. This is likely due to the spatial and temporal parallax between the camera and the LIDAR, causing the projections of the LIDAR points to the camera view to appear shifted for nearby objects.}%
    \label{fig:fcos3d_vs_centerpoint_2}%
\end{figure*}

\begin{figure*}%
    \centering
    \subfloat[\centering FCOS3D]{{\includegraphics[width=0.75\linewidth]{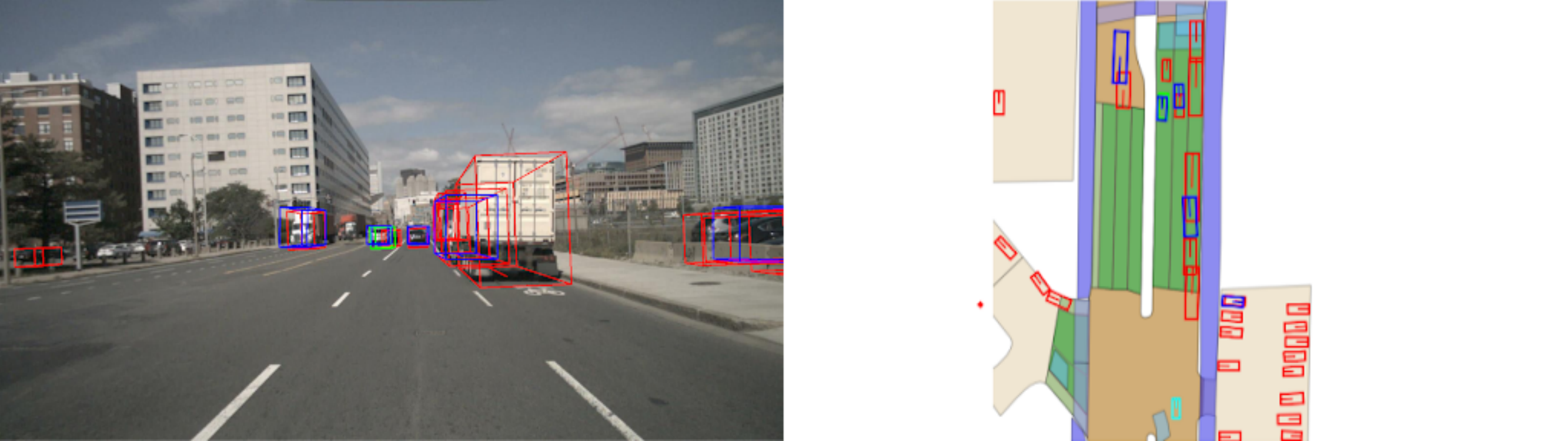} }}%
    \qquad
    \subfloat[\centering CenterPoint]{{\includegraphics[width=0.75\linewidth]{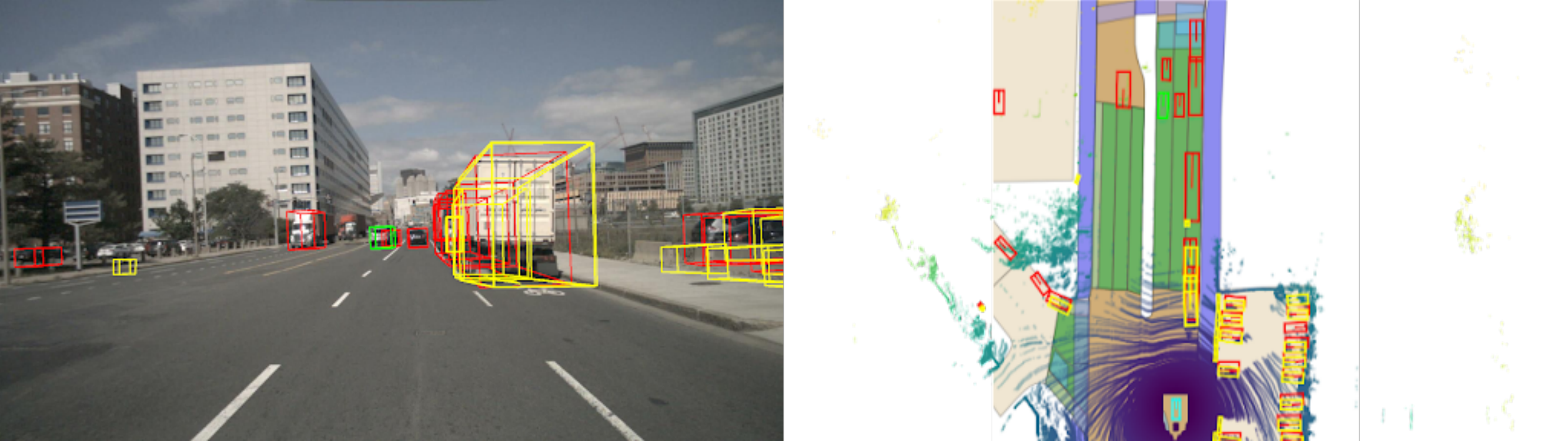} }}%
    \caption{Comparison of 3D object detection for FCOS3D and CenterPoint. The red bounding boxes on both images represent the ground truth detections, the green bounding box represents the referred object, while the FCOS3D and CenterPoint detections are shown in blue and yellow, respectively. It can be seen from the image that in this instance, the referred object, indicated by the green bounding box, is further away down the road. The referred object is detected by the vision-based FCOS3D, while being out of range of the LIDAR sensor used by CenterPoint.}%
    \label{fig:fcos3d_vs_centerpoint_3}%
\end{figure*}

\begin{figure*}%
    \centering
    \subfloat[\centering FCOS3D]{{\includegraphics[width=0.75\linewidth]{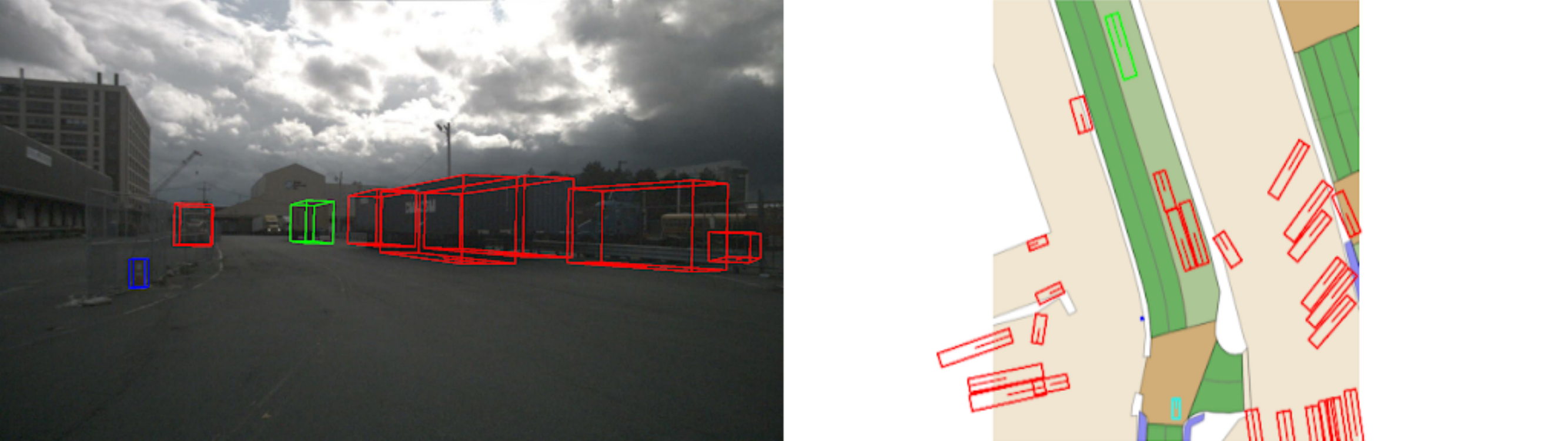} }}%
    \qquad
    \subfloat[\centering CenterPoint]{{\includegraphics[width=0.75\linewidth]{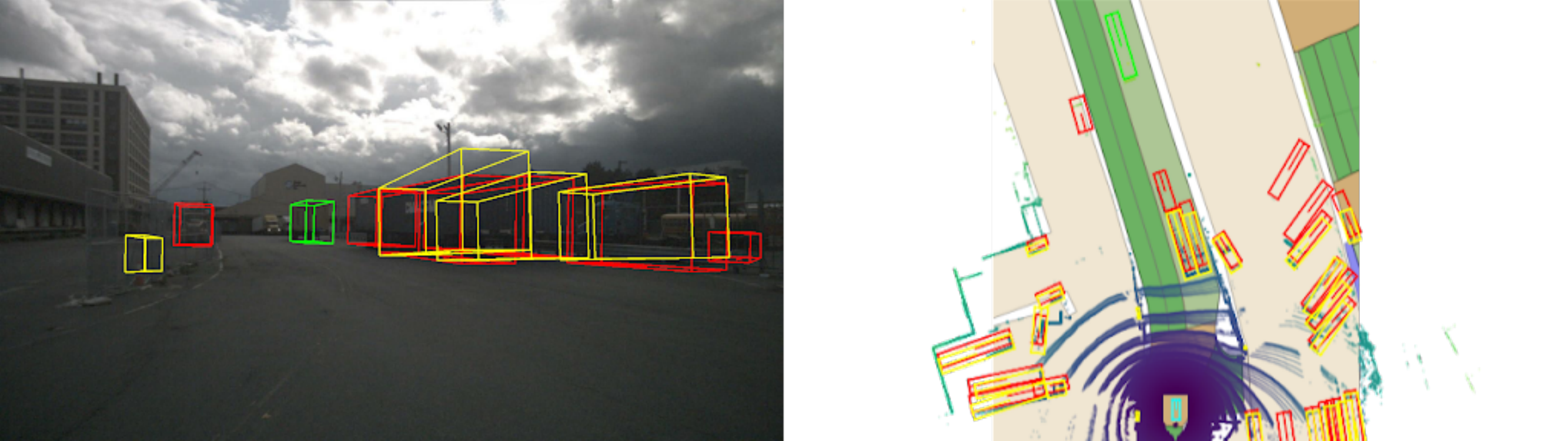} }}%
    \caption{Comparison of 3D object detection for FCOS3D and CenterPoint The red bounding boxes on both images represent the ground truth detections, the green bounding box represents the referred object, while the FCOS3D and CenterPoint detections are shown in blue and yellow, respectively. It can be seen from the image that in the case when the objects are occluded or when their color blends into the background, the LIDAR-based CenterPoint manages to detect such objects. In contrast, in the case of FCOS3D, these objects tend to remain undetected.}%
    \label{fig:fcos3d_vs_centerpoint_4}%
\end{figure*}

\section{Models}
\label{app:baselines}

In this section we discuss the naive baselines and the hyperparameters used for our evaluated methods.

\subsection{Naive Baselines}
The explanations of our naive baselines were omitted from the main paper due to space constraints and we believe the names of these methods are indicative of their approach. 
However, for the interested reader we list their descriptions below.

\paragraph{Random Point} Selects a random point in the $120 \times 80$ m layout as the predicted destination.

\paragraph{Random Road Point} Selects a random point on a drivable surface in the $120 \times 80$ m layout as the predicted destination. The drivable surface can be detected based on the standardized color scheme used for the representation of different surfaces in the top-down layout.

\paragraph{Pick Ego Car} Selects the location of the ego car as the predicted destination. 

\paragraph{Random Object} Selects the location of one of the detected 3D objects as the predicted destination.%
    
\paragraph{Pick Referred Object} Selects the location of the predicted referred object as the predicted destination. 

\subsection{Hyperparameters, Design Choices and Training Details}
This subsection discusses the used hyperparameters, design choices and training details for the models in this paper. All models are implemented in PyTorch\footnote{https://pytorch.org/}.

\paragraph{Layout Encoding}
The input layout is encoded as a three dimensional tensor, with the specified height and width to which the original layout is resized, and 15 channels.

Three channels are reserved for the original RGB image of the road layout as seen from the top-down view, one channel for the ego car bounding box, one channel for the referred object bounding box, and ten channels reserved for the groups of detected object bounding boxes, split up according to their FCOS3D class label predictions.
The ten channels and breakdown corresponds to ten different class labels used by FCOS3D: [car, truck, trailer, bus, construction\_vehicle, bicycle, motorcycle, pedestrian, traffic\_cone, barrier].

We assume the top-down view of the road layout to be readily available. We argue that this is realistic in a practical setting, as an increasing number of systems can provide a detailed overhead road layout in real-time, given the GPS coordinates of the car's current location \footnote{A non-exhaustive list of some of these HD map providers are (and in no particular order): \url{https://www.momenta.cn/en/}, \url{https://www.deepmap.ai/}, \url{https://civilmaps.com/}, and \url{https://www.tomtom.com/products/hd-map/}}.

\paragraph{Referred Object Detector}
The referred object detector (Sentence-BERT+FCOS3D) is trained using the Adam optimizer with a learning rate of $5 \times 10^{-4}$ and weight decay of $10^{-4}$. 
Additionally, we multiply the learning rate by $10^{-1}$ when the average rate of exceeding the IoU of $0.5$ on the validation set does not increase for $3$ epochs. 
We also sweep over the following learning rates with Adam: $3 \times 10^{-4}$, $10^{-4}$, and $10^{-3}$. 
Furthermore, we attempt multiplying the learning rate by $10^{-1}$ after 4 and 8 epochs, but find that the parameters mentioned earlier gave us the best results.
The model is trained for 20 epochs with a batch size of 32. 
The checkpoint with the highest average rate of exceeding the IoU $> 0.5$ on the validation set is used on the test set.
The model is trained with the predicted bounding boxes from FCOS3D, where we use the bounding box whose projection to the camera view scored the highest IoU with the ground truth bounding box as the target. 

\paragraph{SinglePoint}
The base training setup consists of training the model for up to 50 epochs, where the training is stopped if there is no improvement in the validation loss over a period of 10 epochs, and keeping track of the model that achieves the lowest loss on the validation set.
The loss used is the mean distance between the prediction and the nearest ground truth destination. 
We use the Adam optimizer with the learning rate of $3 \times 10^{-5}$ and $\beta_1$ and $\beta_2$ coefficients used in the computation of running averages of gradients set to 0.9 and 0.999, respectively.
We use a batch size of 16 during training and apply gradient clipping with the magnitude threshold of 5. 
Additionally, a ResNet18 model is used to encode the layout tensor, whose height and width are set to 200 and 300, respectively.

\paragraph{NonParametric}
We use the same base setup as in SinglePoint, with the addition of two hyperparameters, $\sigma$ and $t_{\sigma}$. 
These represent the standard deviation and the bounds of truncated Gaussians used for separately creating soft target vectors for the $x$ and $y$ axis. 
To create these soft targets, we center a truncated Gaussian around the vector indices $j$ corresponding to ground-truth locations, as opposed to using multi-hot encoding with binary entries:
\begin{equation}
    t_{ij} = 
        \begin{cases}
            \frac{1}{\sqrt{2\pi}\sigma} \exp^{-\frac{(i - j)^2}{2\sigma^2}} & \text{for } |i - j| \leq  t_{\sigma}\\
            0 & \text{for } |i - j| > t_{\sigma}
        \end{cases} ,
\end{equation}
where $i \in \s{I}$ is the set of all possible indices along a certain axis (i.e. $x$ or $y$ axis).
The entries in the target vector $\v{t}$ are computed as $t_i = \sum_{j \in \s{J}} t_{ij}/ {\sum_{k \in \s{I}, j \in \s{J}} t_{kj}}$ where $\s{J}$ is the set of all vector indices that correspond to a coordinate of a ground truth destination.
We use the value of 11 for $t_{\sigma}$, and the value of 3 for $\sigma$. We use 200 for the height and 300 for the width of the layout grid.
As loss, we use the KL-Divergence between the predicted distribution over discrete positions along each axis, and the target vectors $t$ of each axis. The positions along $x$ and $y$ are treated independently and their loss contributions are summed up.

\paragraph{UnimodalNormal}
We use the same base setup as in the baselines above, except for the Adam learning rate, which is set to $10^{-4}$. We found that predicting only the standard deviations along $x$ and $y$ and assuming no correlation outperformed predicting the full covariance matrix of a bivariate Gaussian.

\paragraph{MDN}
We use the same base setup as in UnimodalNormal, except for the learning rate reduced to $3 \times 10^{-5}$, and the addition of the number of mixture components as an additional parameter. We found three to be the best number of mixture components among the candidates ranging from one to five. Unlike in the UnimodalNormal, we found it preferable to predict the full covariance matrix of each Gaussian mixture component, or explicitly, the lower-triangular Cholesky component of the covariance matrix.

\paragraph{Adapted GoalGAN} We adapted the official GoalGAN implementation\footnote{\url{https://github.com/dendorferpatrick/GoalGAN}} and trained it with the original hyperparameters provided by \citet{dendorfer2020goal}. 
Early stopping was applied on the validation loss if the loss did not decrease for more than ten epochs. The model was trained with the layout dimensions $120 \times 80$.

\paragraph{Endpoint VAE}
We use a ResNet-18 to encode the layout tensor the same way as in the SinglePoint baseline. We then use the default settings for the Endpoint VAE as in the official implementation of PECNet\footnote{\url{https://github.com/HarshayuGirase/PECNet}}, except for increasing the input dimensionality of the latent encoder and latent decoder to 912, to accommodate for the increased size of the command and layout embedding compared to the PECNet's past trajectory encoding, as well as increasing the size of the first hidden layer of the latent encoder to 256. We use the Adam learning rate of $3 \times 10^{-5}$, the batch size of 32, and the weight decay term of $10^{-3}$. We set the weight for the KL-Divergence loss term to $0.1$, and the standard deviation for the zero-mean isotropic Gaussian used for drawing latent samples during inference is set to $1.3$. 

\paragraph{Adapted RegFlow}

We adapt the official RegFlow implementation\footnote{https://github.com/maciejzieba/regressionFlow} and use the original hyperparameters provided by the authors, with the exception of introducing batch norm, which we did as we found the training to be very unstable otherwise. We train the model with image sizes of $600 \times 400$, $300 \times 200$, and $150 \times 100$, and find that $150 \times 100$ achieves the highest result. We consider the learning rates  $2 \times 10^{-4}$, $2 \times 10^{-5}$ and $2 \times 10^{-6}$ and find that the original $2 \times 10^{-5}$ achieves the best performance. A batch size of 10 was used during training.

\paragraph{\gls{destination_predictor}}
We use 192 and 288 as the height and width dimensions of the layout tensor, respectively, in order to avoid round offs in the calculation of the dimensions of downsampled feature maps, and allow for the accurate calculation of the corresponding locations in the original layout grid. This small reduction of the height and width of the layout was found to have no consequence on the performance of other models. After processing the input layout with an FPN on top of a ResNet-18 backbone modified to accept a 15 channel input (non-pretrained), we obtain four 256 channel feature maps, downsampled at the rates of 4, 8, 16 and 32. 
We use five blocks shared across different scales, each consisting of a convolutional layer with 256 channels, kernel size of 3 and the padding of 1, followed by group normalization with 32 channel groups and a ReLU. We find that using the encoded command to attend to the output of the second out of the five blocks yielded the best performance. The mean, standard deviation and mixture weight prediction heads are also convolutional layers, with the kernel size of 3, padding of 1, with the output channel count of 2, 2 and 1, respectively. In order to ensure that the predicted standard deviations are positive, we use $1 + ELU(\sigma_{pred}) + 10^{-5}$ as the final standard deviation for the Gaussian mixture components, where $\sigma_{pred}$ is the output of the prediction head used to compute the standard deviation. We train the model for 50 epochs, while using the batch size of 32, the Adam learning rate of $3 \times 10^{-5}$ and $\beta_1$ and $\beta_2$ coefficients used in the computation of running averages of gradients set to 0.9 and 0.999, respectively, while applying gradient clipping with the magnitude threshold of 5.

\section{Additional Experimental Analysis}
The following results are not the main findings of this paper but provide additional insight for future research.

\subsection{Breakdown of Results Based on Intent}
In this subsection we break down the results of the evaluated baselines over the intents of the commands on the Talk2Car-Destination test set. 
In Table \ref{tab:intent_ade} we see the results in terms of ADE.
We notice that both our model PDPC - Top 32 and Endpoint VAE perform the best over all intent categories.
In Table \ref{tab:intent_MDE} we observe that PDPC - Top 32 is nearly the best on all intent categories, achieving the MDE as low as $2.89$ for the intent ``Pick Up''.
In Table \ref{tab:intent_pa_2} and \ref{tab:intent_pa_4} we see the results for $PA_2$ and $PA_4$ respectively. 
In both tables we observe that PDPC - Top 32 is the most model across all intents and also notice that in some cases (i.e. ``U-Turn Left'' and ``U-Turn Right'') our $PA_2$ is nearly double that of the second best model. 
In the $PA_4$ table, we even see that for some intents (i.e. ``Change Lane Left, ``Stop'', ``Pick Up''), more than 50\% of our predictions fall in 4 meters (which is one car length) of the ground truth destination.

\subsection{Qualitative Analysis}
In this experiment, we compare the qualitative results from \gls{destination_predictor} against the results from the second-best performing model, the Endpoint VAE.
As this latter model outputs multiple destination predictions by sampling the zero-mean isotropic Gaussian latent distribution and feeding the samples to the decoder, we do not represent its output in the frontal and top-down view with a heatmap but rather directly display the predicted points. 
As shown in Figure \ref{fig:endpoint_vae_vs_ours_1}, we observe that \gls{destination_predictor} outputs a heatmap that corresponds with the ground truth destination points. For Endpoint VAE, we also notice that its predictions are rather close to the ground truth destination. However, its predictions are outside the road layout. 
We see a similar pattern in Figure \ref{fig:endpoint_vae_vs_ours_2}.
Please note that the inputs to both models are identical, and thus each model needs to learn the implications of the road layout. 
From these images, we observe that \gls{destination_predictor} does a better job at keeping its predictions on the road.
In Figure \ref{fig:endpoint_vae_vs_ours_3}, we see that both models make predictions that are in the vicinity of the ground truth referred object. However, the predictions from \gls{destination_predictor} are closer to the ground truth destination than those of Endpoint VAE.

\subsection{Influence of Top-K Components}
From our results, we see that we can improve the results from the base \gls{destination_predictor} model by only selecting the top-k components of the mixture. Figure~\ref{fig:top-k-vs-base-1} showcases an example of the difference between using the top-8 components and using all mixture components. One can see that the base model indicates more possible destinations than the top-8 model. Hence, reducing the number of components can reduce the multimodality of the output distribution. This is not always the case, as can be seen in Figure \ref{fig:top-k-vs-base-2}. In this image, both models indicate that there is only one possible destination region.

\subsection{Failure Cases}
Figure \ref{fig:failure_cases} showcases two failure cases.
In Figure \ref{fig:failure_cases}(a) and \ref{fig:failure_cases}(b) we show a failure case from the referred object detector side. As we see, the detector selected the first car on the left while the command was referring to the second car on the left. However, even while the predicted referred object is wrong, we observe that our \gls{destination_predictor} still makes a very sensible prediction relative to this wrongly predicted referred object.

In Figure \ref{fig:failure_cases}(c) and \ref{fig:failure_cases}(d) we show a failure case stemming from a failure on the 3D object detector side, where the 3D object detector did not detect the truck at all. As a side effect, the referred object detector can not predict the correct referred object, but it still makes a prediction. Then, the \gls{destination_predictor} tries to make a sensible prediction based on the wrongly predicted referred object.

\begin{figure*}[h]
    \centering
    \subfloat[\gls{destination_predictor} Base -  Frontal]{{\includegraphics[width=0.45\linewidth]{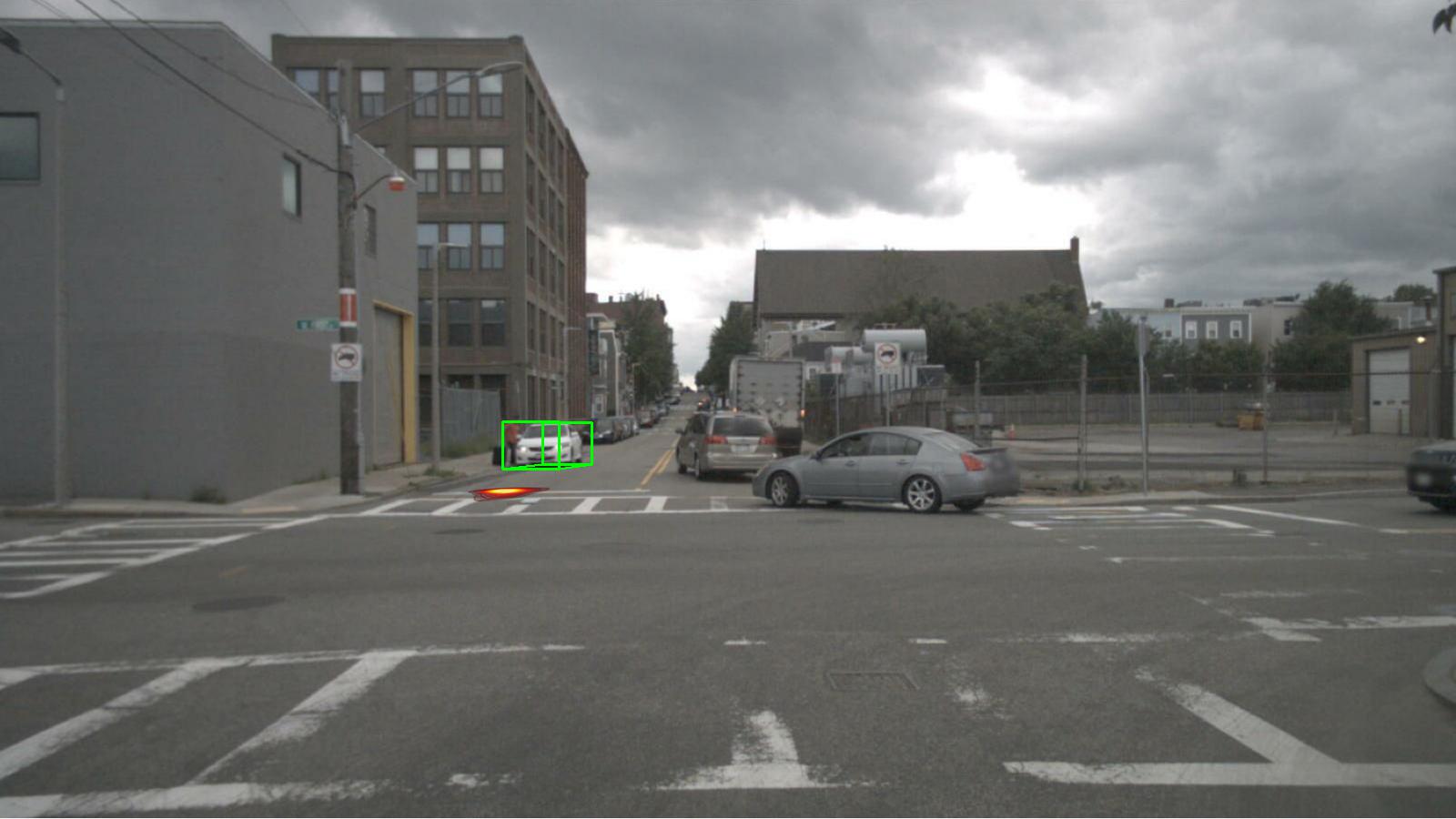}}}
    \subfloat[\gls{destination_predictor} Base - Top-down]{{\includegraphics[width=0.45\linewidth]{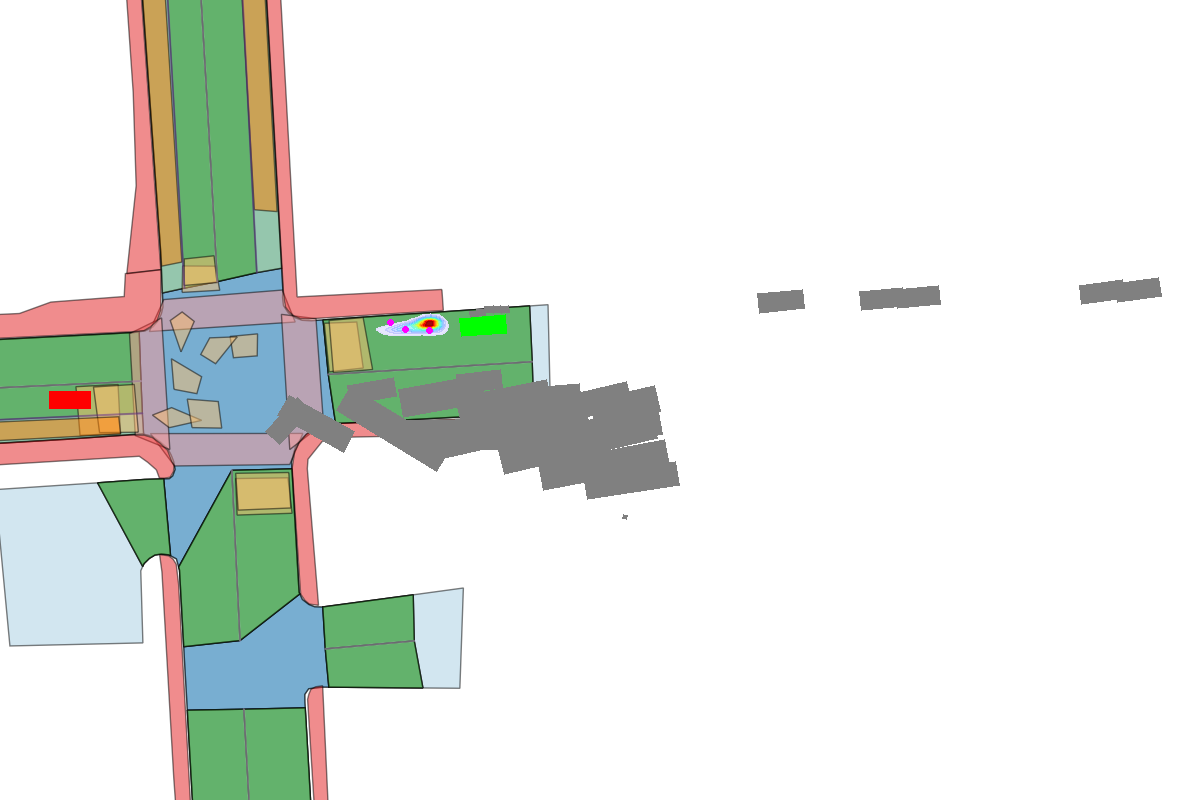}}}
    \qquad
      \subfloat[Endpoint VAE - Frontal]{{\includegraphics[width=0.45\linewidth]{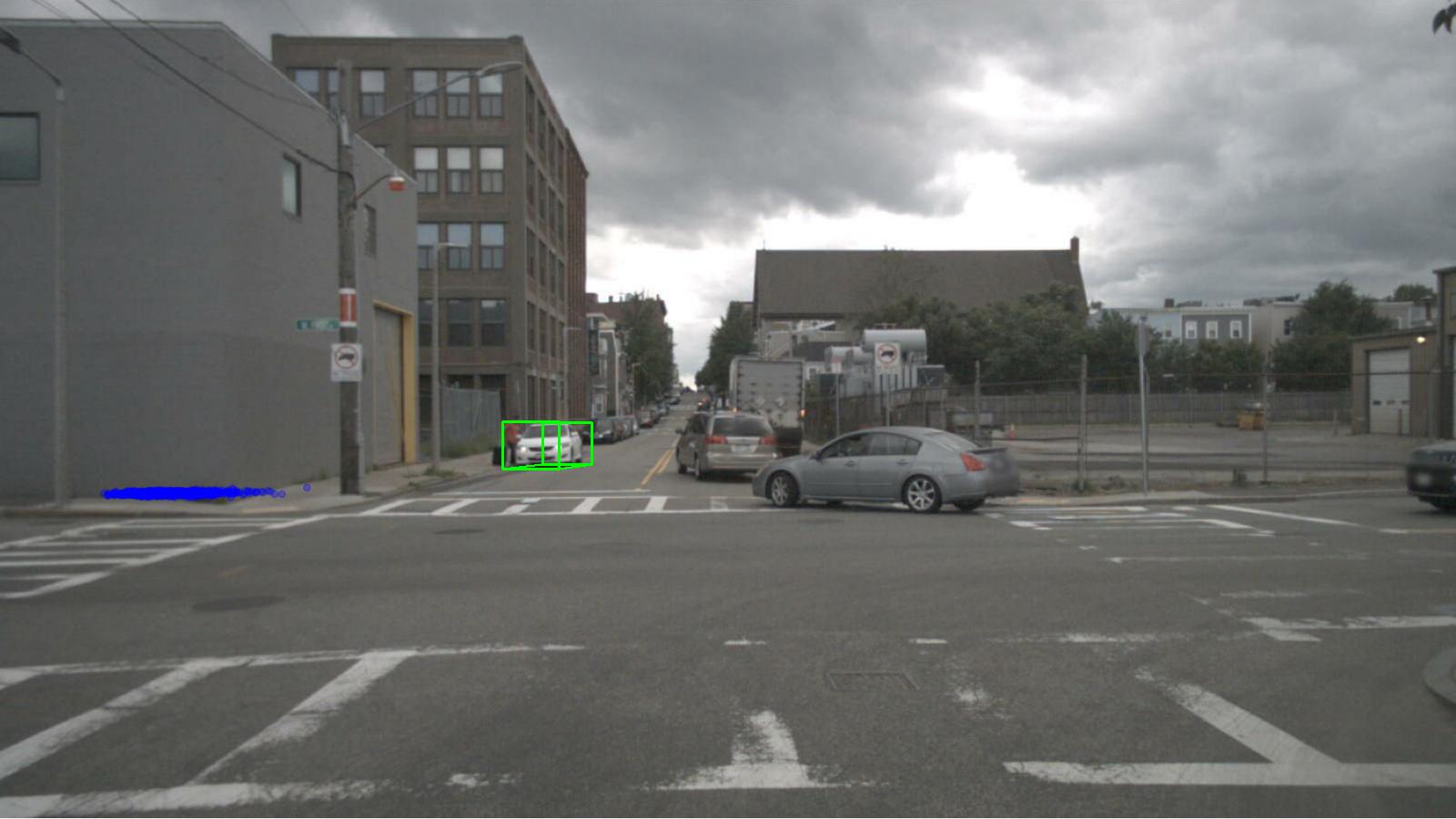}}}
    \subfloat[Endpoint VAE - Top-down]{{\includegraphics[width=0.45\linewidth]{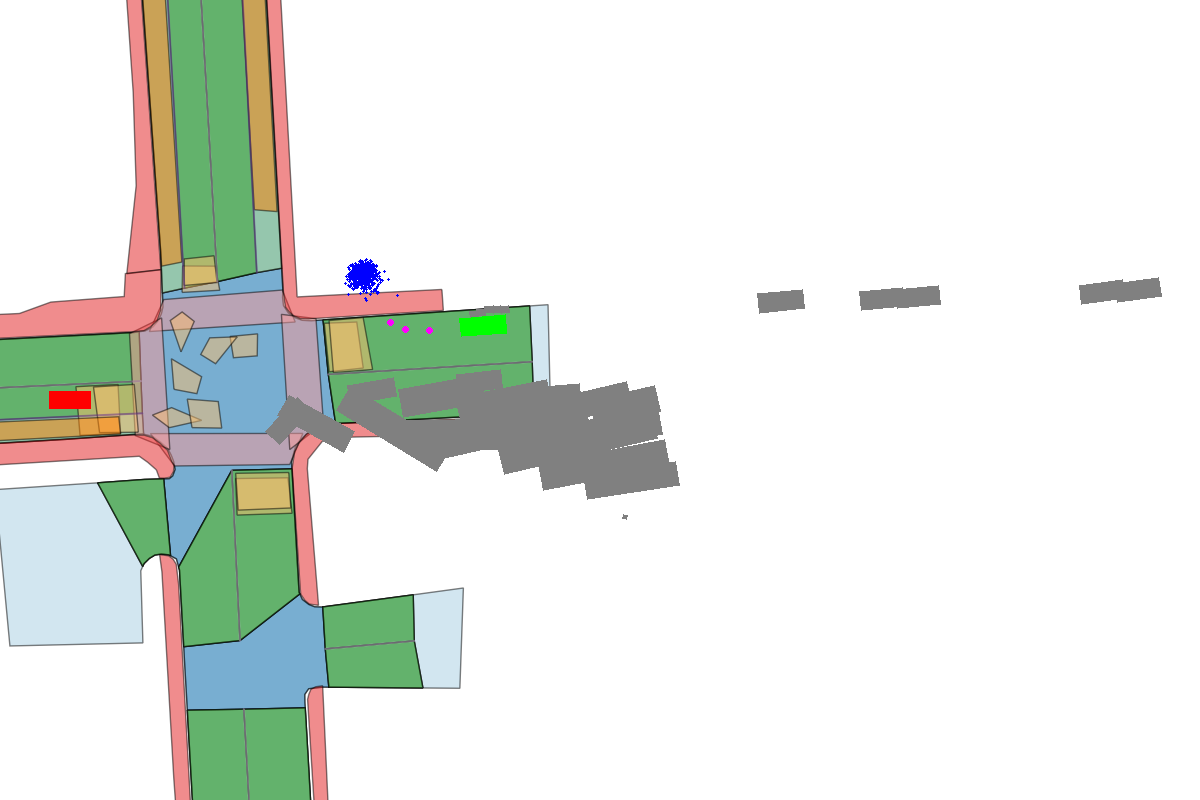}}}
    \caption{The issued command was: ``Park in front of that white car on the left side of the road.''. In this figure, we present the destination prediction of our \gls{destination_predictor} base model (top row) against the second-best model, Endpoint VAE (bottom row). 
    The red car on the top-down view is the ego car. The purple dots are the ground truth destination. The gray boxes represent other detected objects. The green rectangle in the top-down and frontal view is the referred object. In the figure, we see that the predicted heatmap from \gls{destination_predictor} corresponds with the ground truth destination and is located on a drivable surface. On the other hand, we see the predicted samples from Endpoint VAE are rather close to the actual ground truth destination, but outside of the road.
    }%
    \label{fig:endpoint_vae_vs_ours_1}%
\end{figure*}

\begin{figure*}[h]
    \centering
    \subfloat[\gls{destination_predictor} Base - Frontal]{{\includegraphics[width=0.45\linewidth]{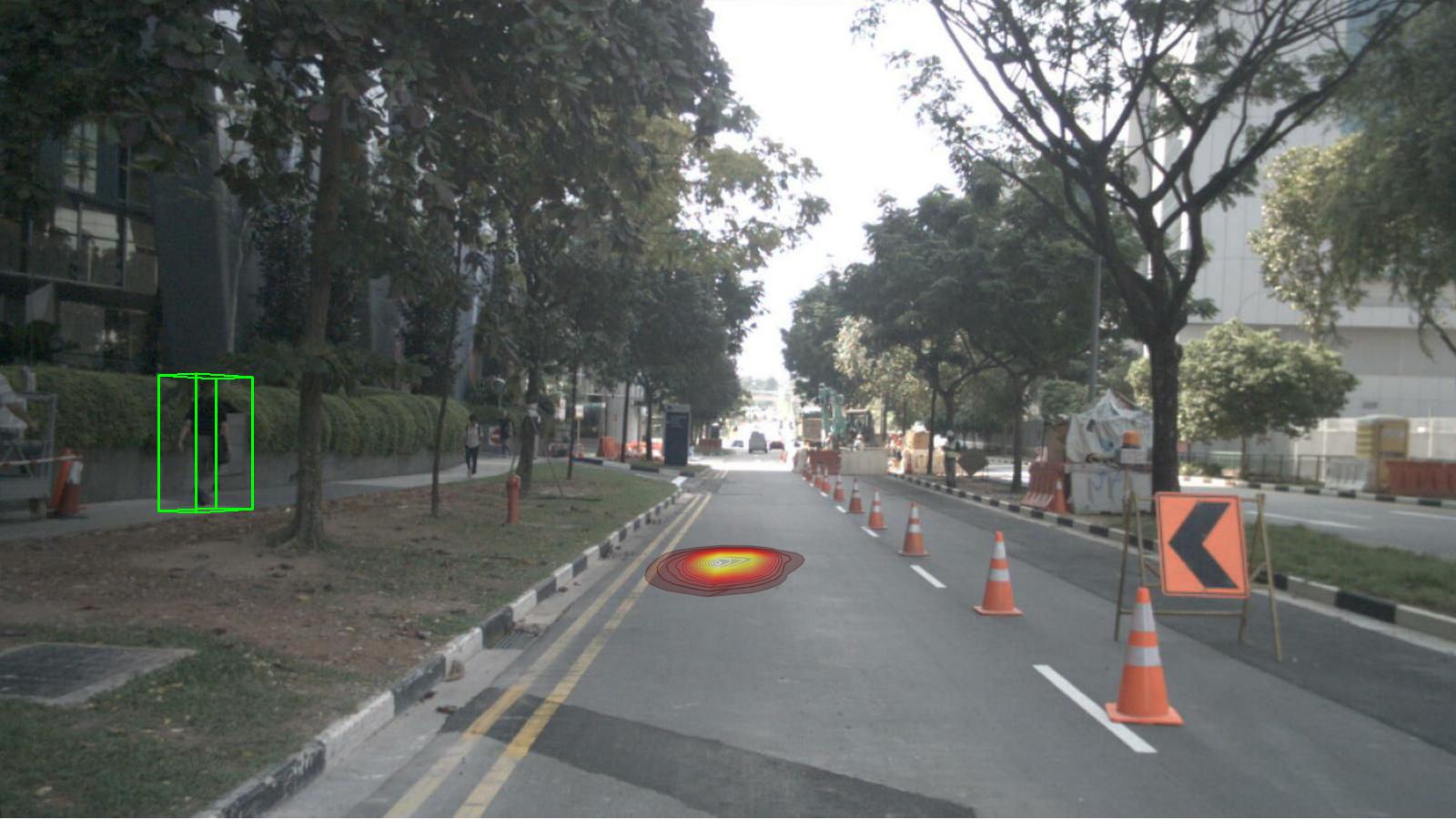}}}
    \subfloat[\gls{destination_predictor} Base - Top-down]{{\includegraphics[width=0.45\linewidth]{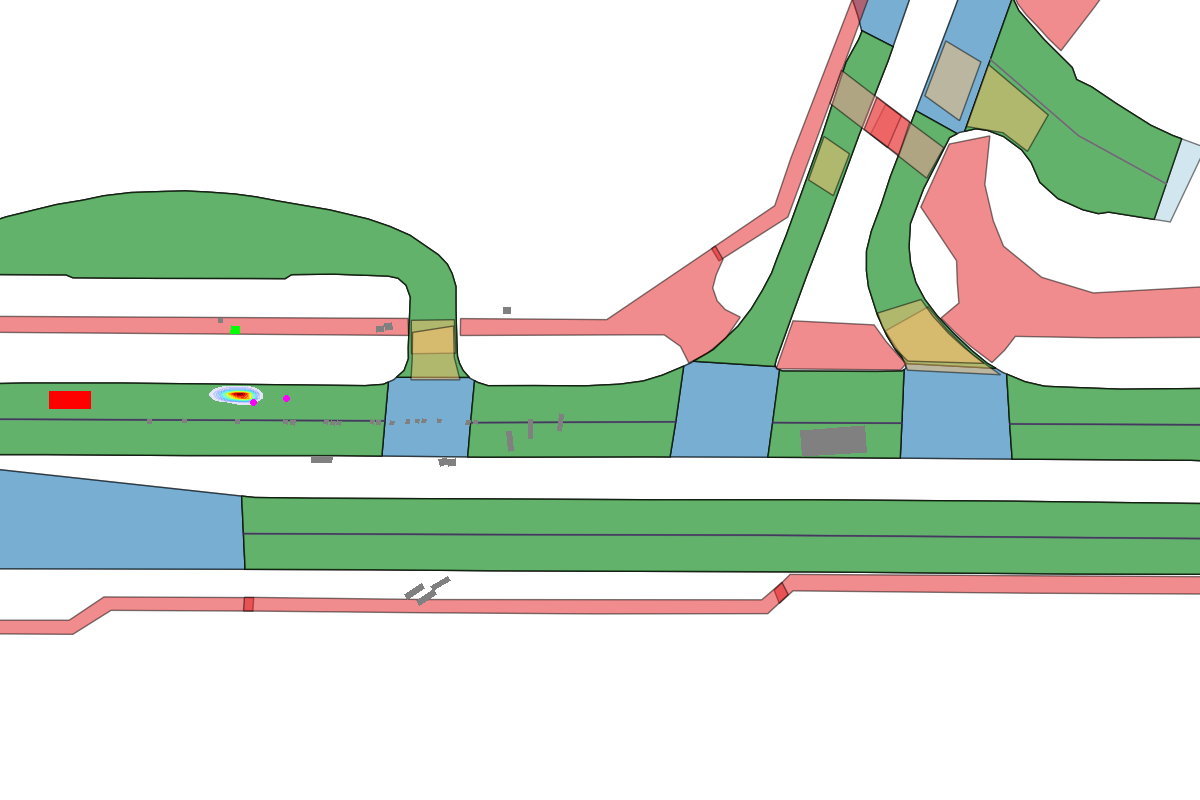}}}
    \qquad
      \subfloat[Endpoint VAE - Frontal]{{\includegraphics[width=0.45\linewidth]{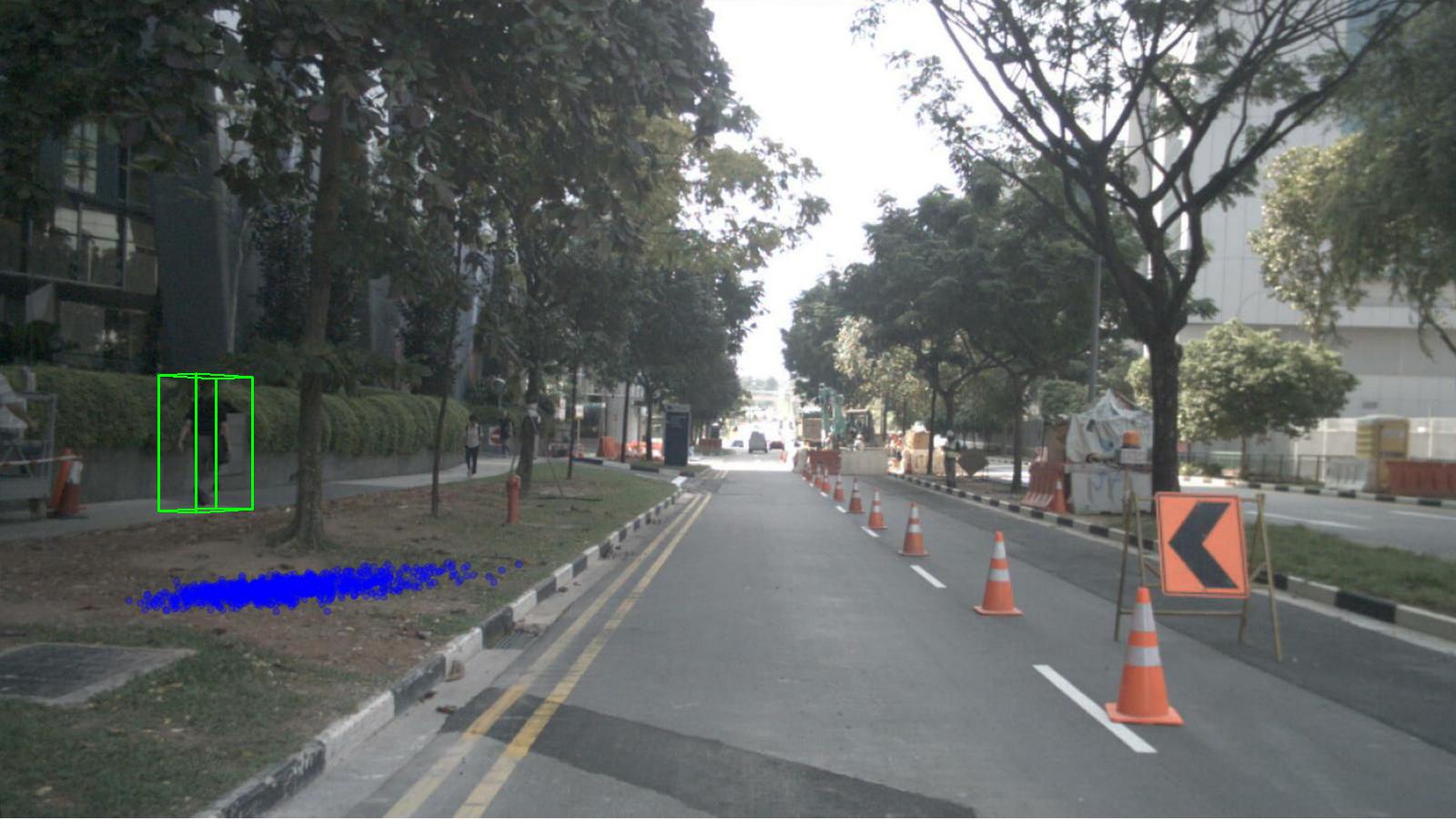}}}
    \subfloat[Endpoint VAE - Top-down]{{\includegraphics[width=0.45\linewidth]{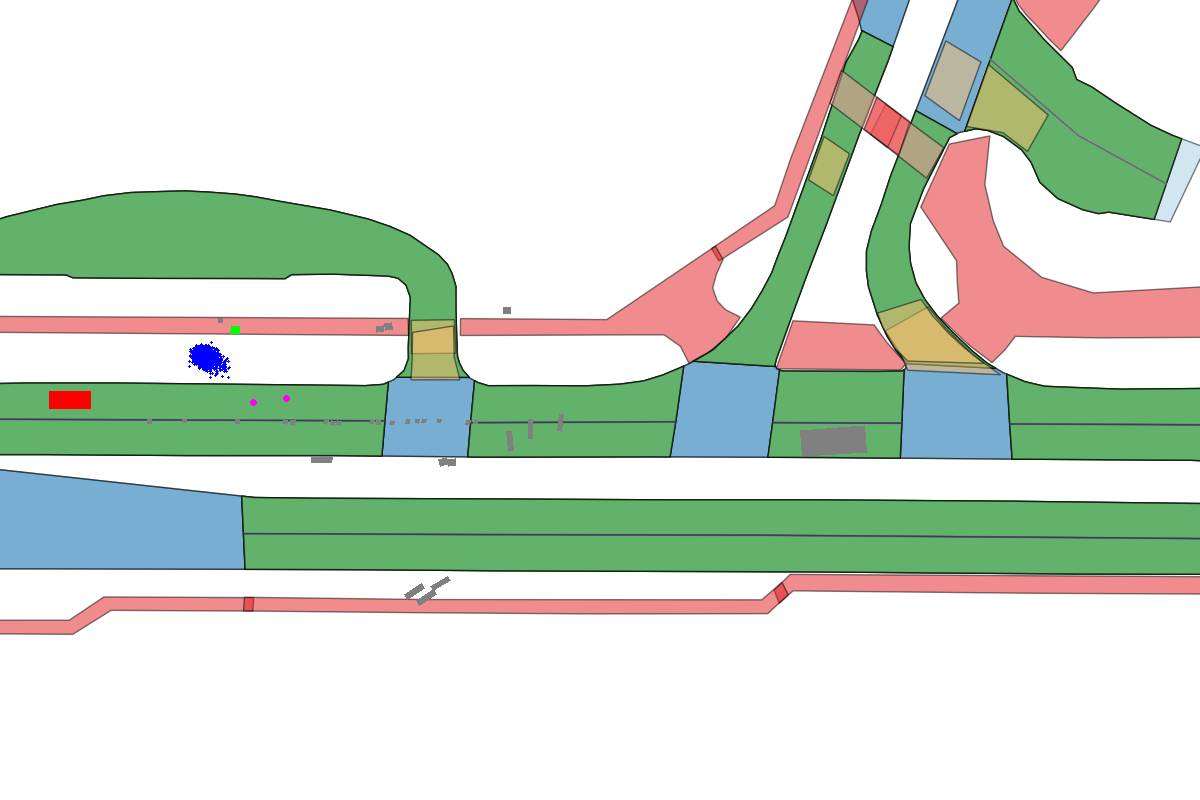}}}
    \caption{The issued command was: ``Honk at my friend matt there as we pass.''. 
    In this figure, we present the destination prediction of our \gls{destination_predictor} base model (top row) against the second-best model, Endpoint VAE (bottom row). 
    The red car on the top-down view is the ego car. The purple dots are the ground truth destination. The gray boxes represent other detected objects. The green rectangle in the top-down and frontal view is the referred object. In the figure, we see that the predicted heatmap from \gls{destination_predictor} corresponds with the ground truth destination and is inside the road layout. On the other hand, we again see the predicted samples from Endpoint VAE are rather close to the actual ground truth destination but outside of the road layout.
    }
    \label{fig:endpoint_vae_vs_ours_2}%
\end{figure*}

\begin{figure*}[h]
    \centering
    \subfloat[\gls{destination_predictor} Base - Frontal]{{\includegraphics[width=0.45\linewidth]{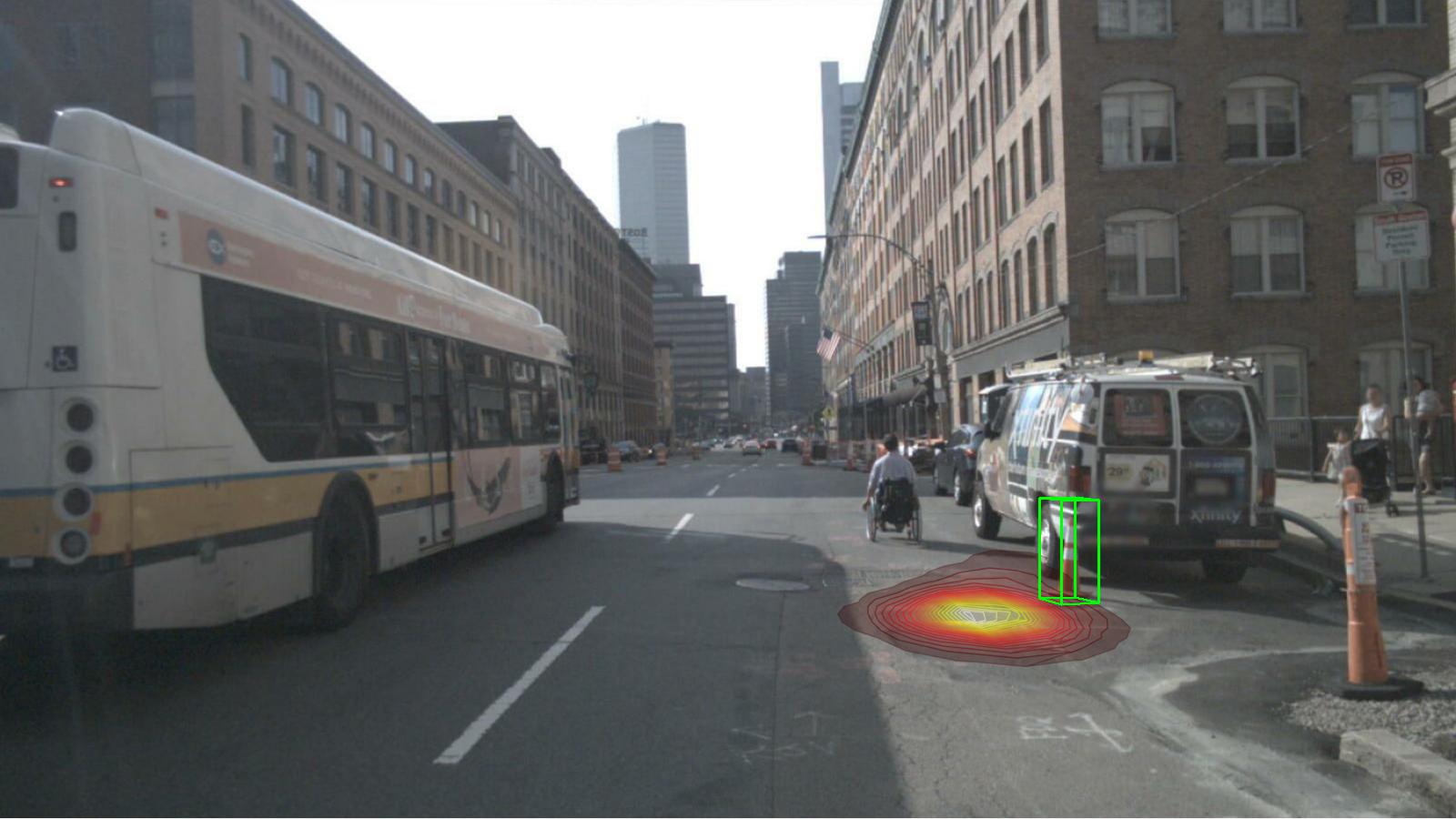}}}
    \subfloat[\gls{destination_predictor} Base - Top-down]{{\includegraphics[width=0.45\linewidth]{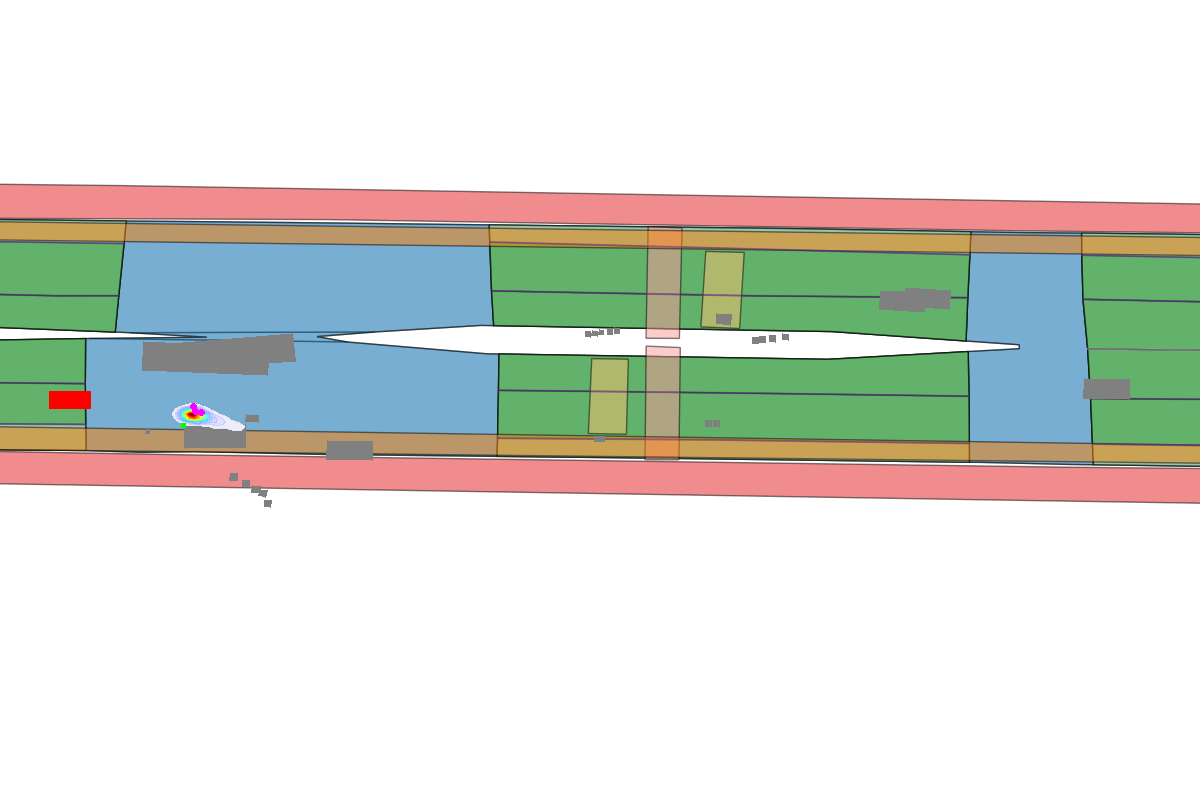}}}
    \qquad
      \subfloat[Endpoint VAE - Frontal]{{\includegraphics[width=0.45\linewidth]{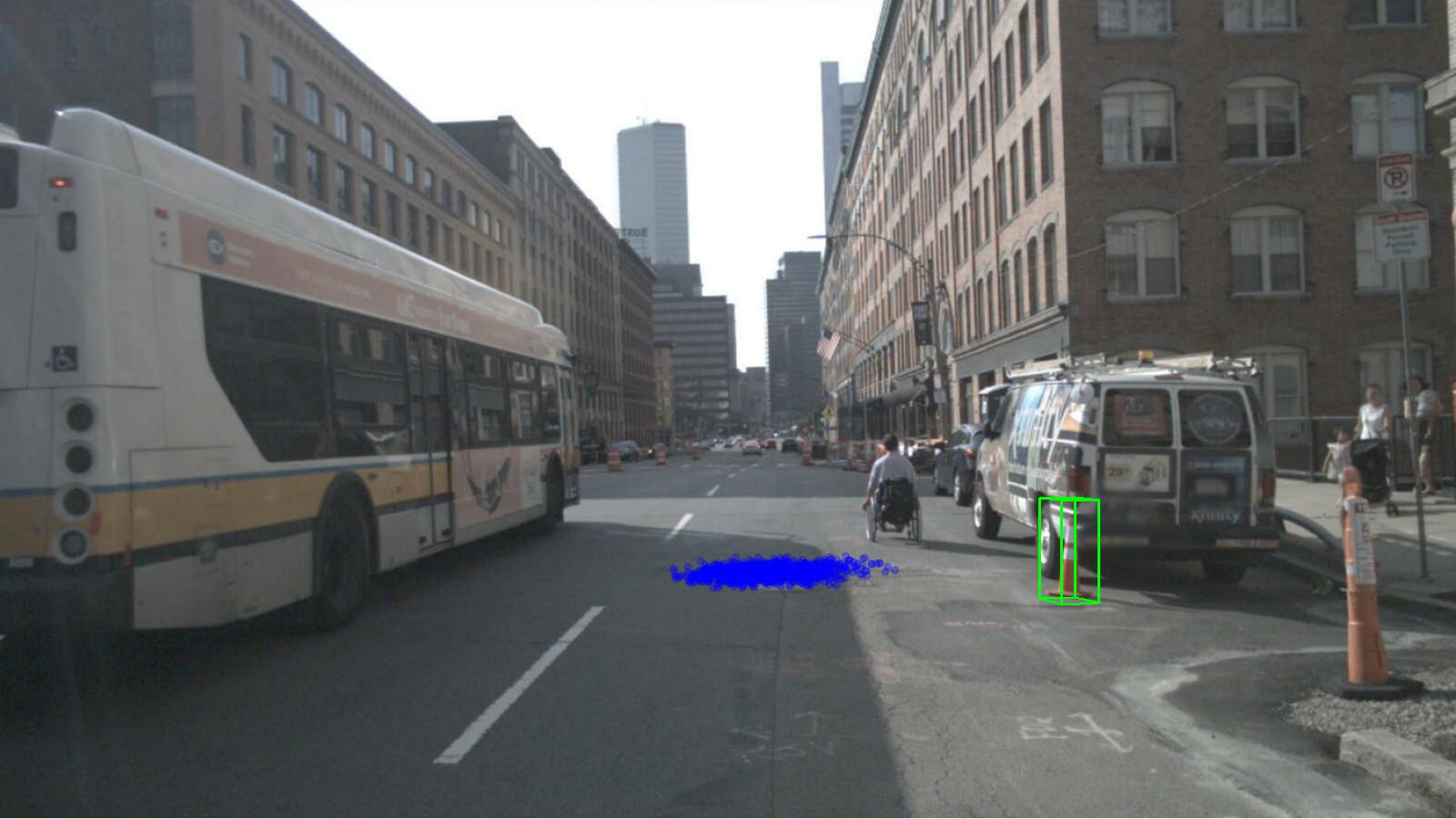}}}
    \subfloat[Endpoint VAE - Top-down]{{\includegraphics[width=0.45\linewidth]{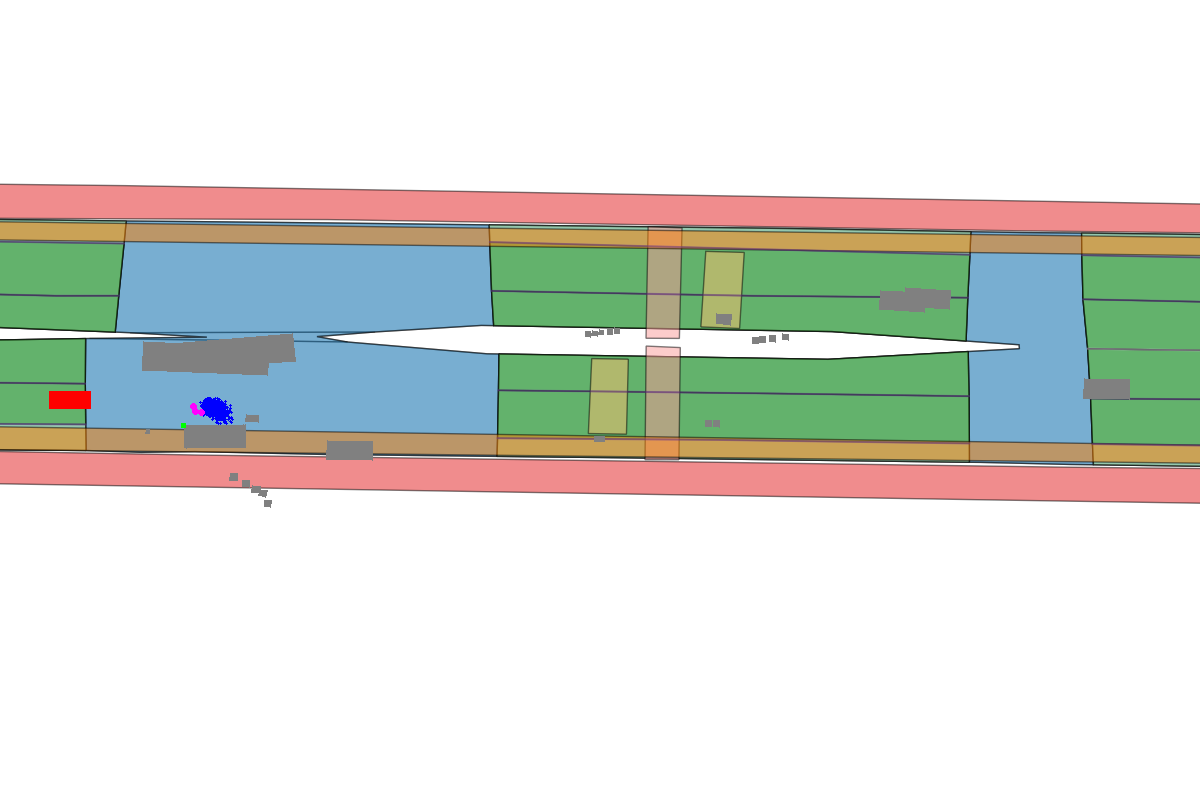}}}
    \caption{The issued command was: ``Pull up next to that second traffic cone.''. In this figure, we present the destination prediction of our \gls{destination_predictor} base model (top row) against the second-best model, Endpoint VAE (bottom row). 
    The red car on the top-down view is the ego car. The purple dots are the ground truth destination. The gray boxes represent other detected objects. The green rectangle in the top-down and frontal view is the referred object. In the figure, we notice that the predictions from both models are rather accurate. However, the prediction from \gls{destination_predictor} is even closer to the ground truth destination than the prediction from Endpoint VAE.
    }%
    \label{fig:endpoint_vae_vs_ours_3}%
\end{figure*}

\begin{figure*}[h]
    \centering
    \subfloat[\gls{destination_predictor} Top-8 - Frontal]{{\includegraphics[width=0.45\linewidth]{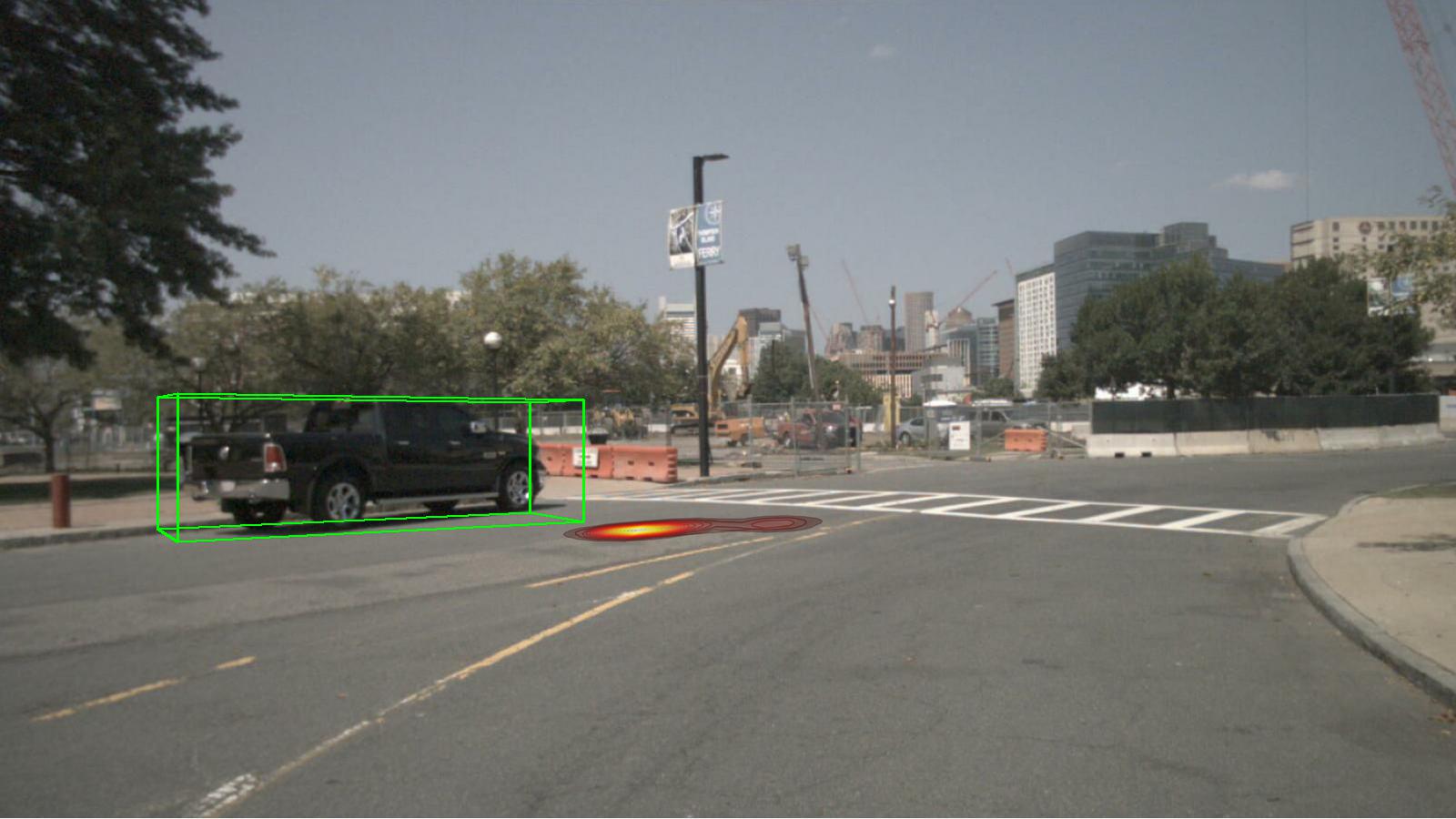}}}
    \subfloat[\gls{destination_predictor} Top-8 - Top-down]{{\includegraphics[width=0.45\linewidth]{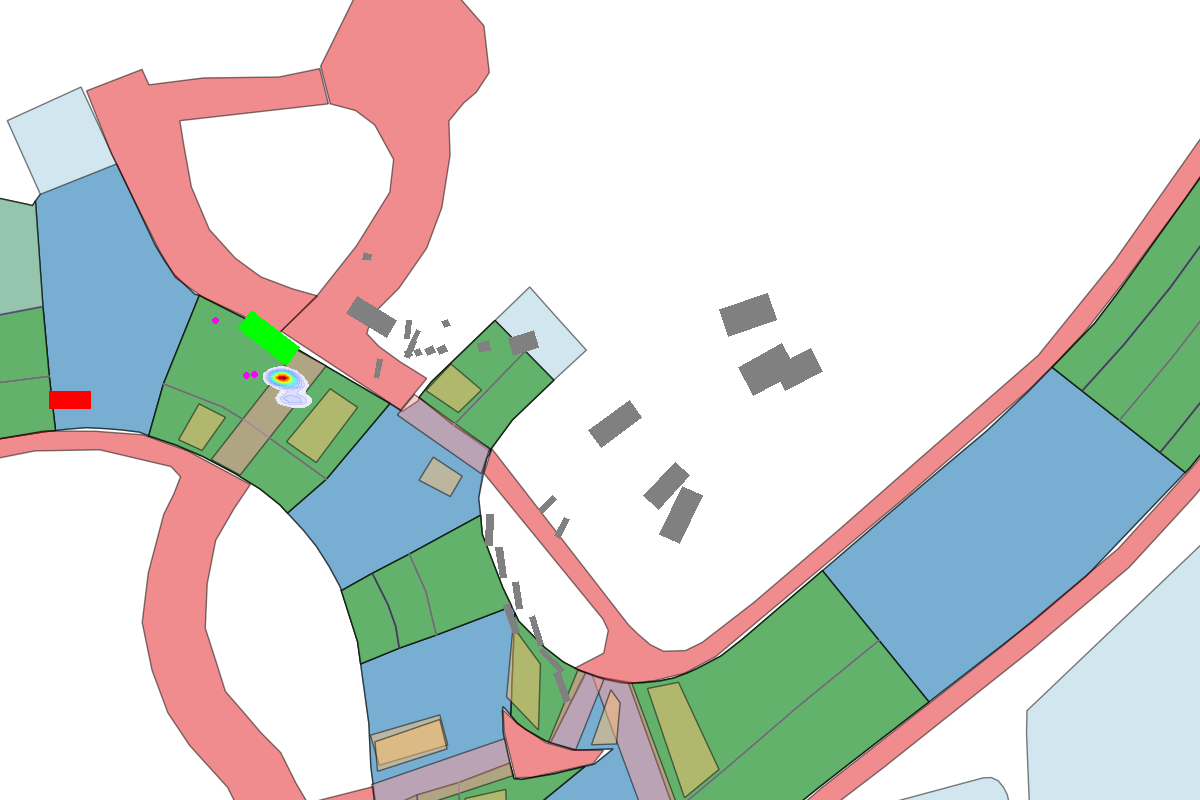}}}
    \qquad
      \subfloat[\gls{destination_predictor} Base - Frontal]{{\includegraphics[width=0.45\linewidth]{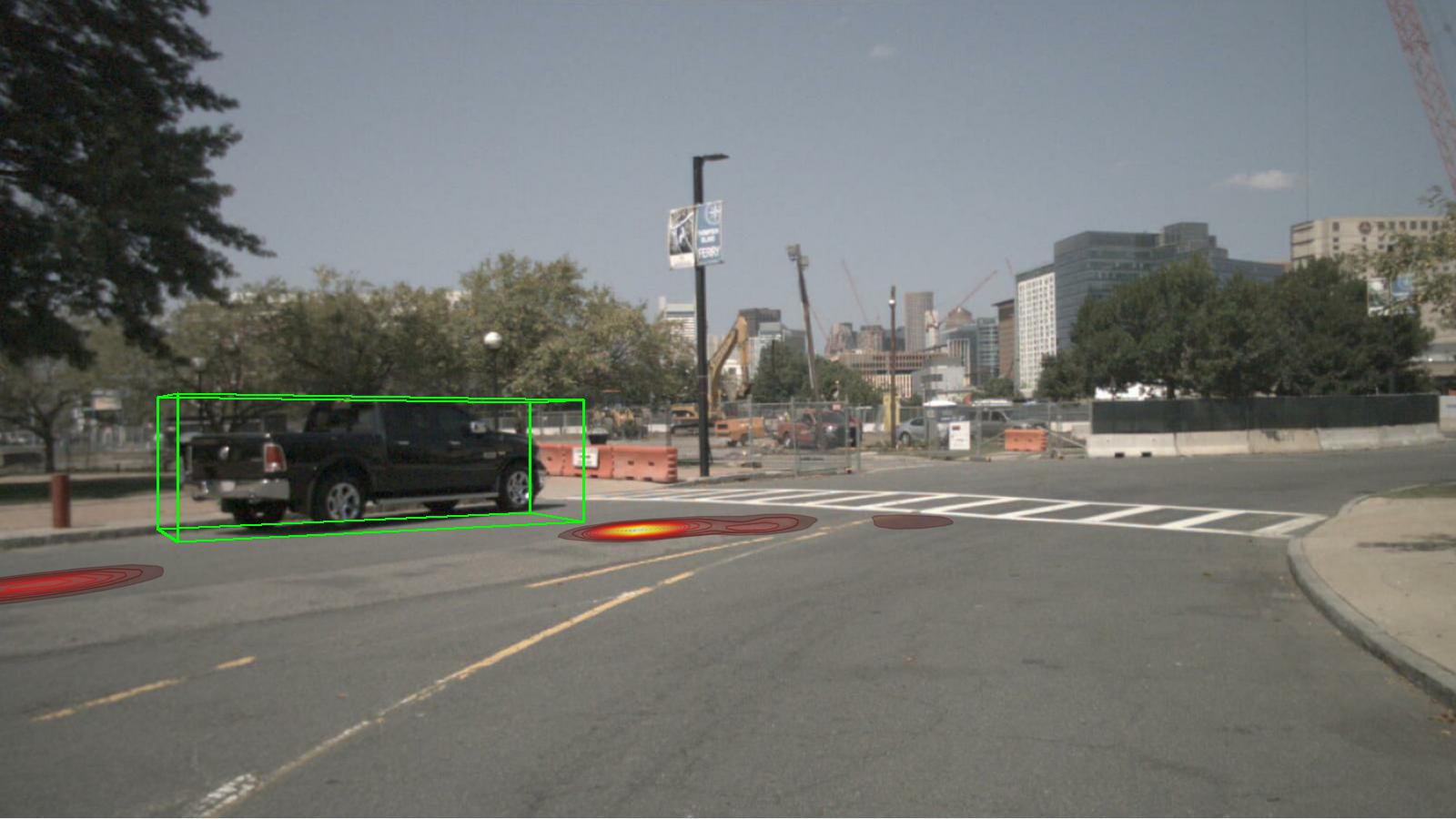} }}
    \subfloat[\gls{destination_predictor} Base - Top-down]{{\includegraphics[width=0.45\linewidth]{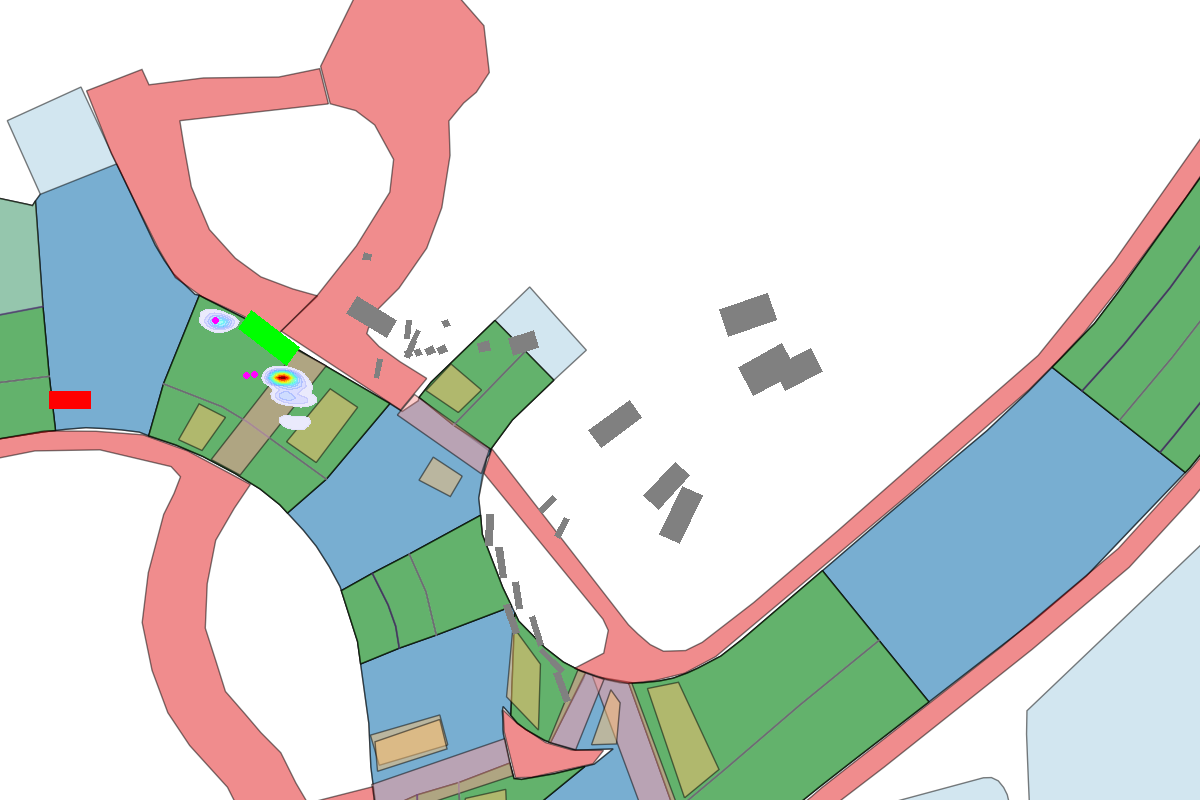}}}
    \caption{The issued command was: ``Pull up next to that truck.''. This image showcases the difference between \gls{destination_predictor} using only the top-8 ranked components of the mixture (top row) against the same model using all components of the mixture. One can see that the heatmap of the base model (bottom row) contains more modes than in the case when only the top-8 ranked components are used. 
    The red car on the top-down view is the ego car.
    The purple dots are the ground truth destination. The gray boxes represent other detected objects.
    The green rectangle in the top-down view and the frontal view indicates the referred object.}%
    \label{fig:top-k-vs-base-1}%
\end{figure*}

\begin{figure*}[h]
    \centering
    \subfloat[\gls{destination_predictor} Top-8 - Frontal]{{\includegraphics[width=0.45\linewidth]{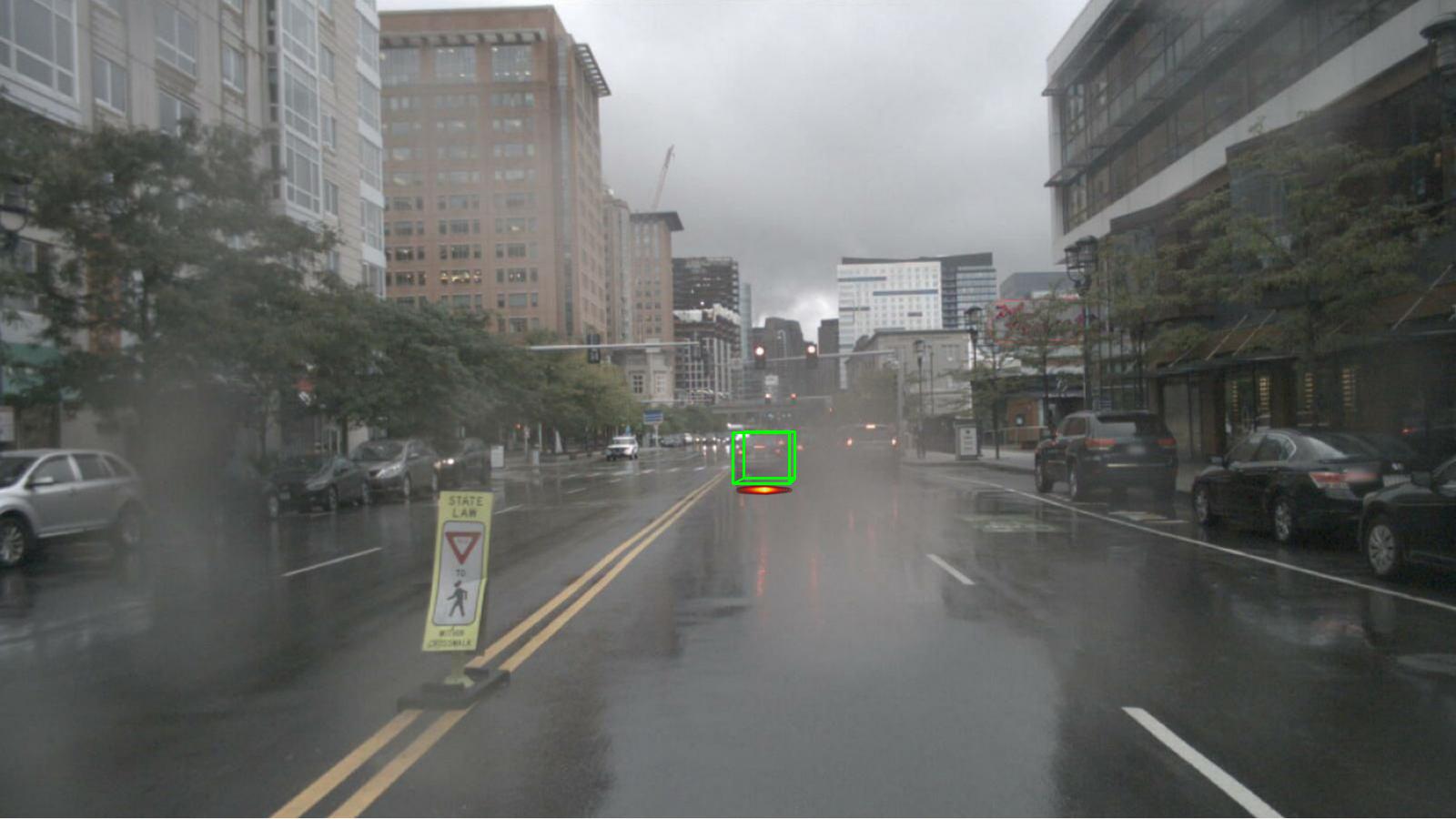}}}
    \subfloat[\gls{destination_predictor} Top-8 - Top-down]{{\includegraphics[width=0.45\linewidth]{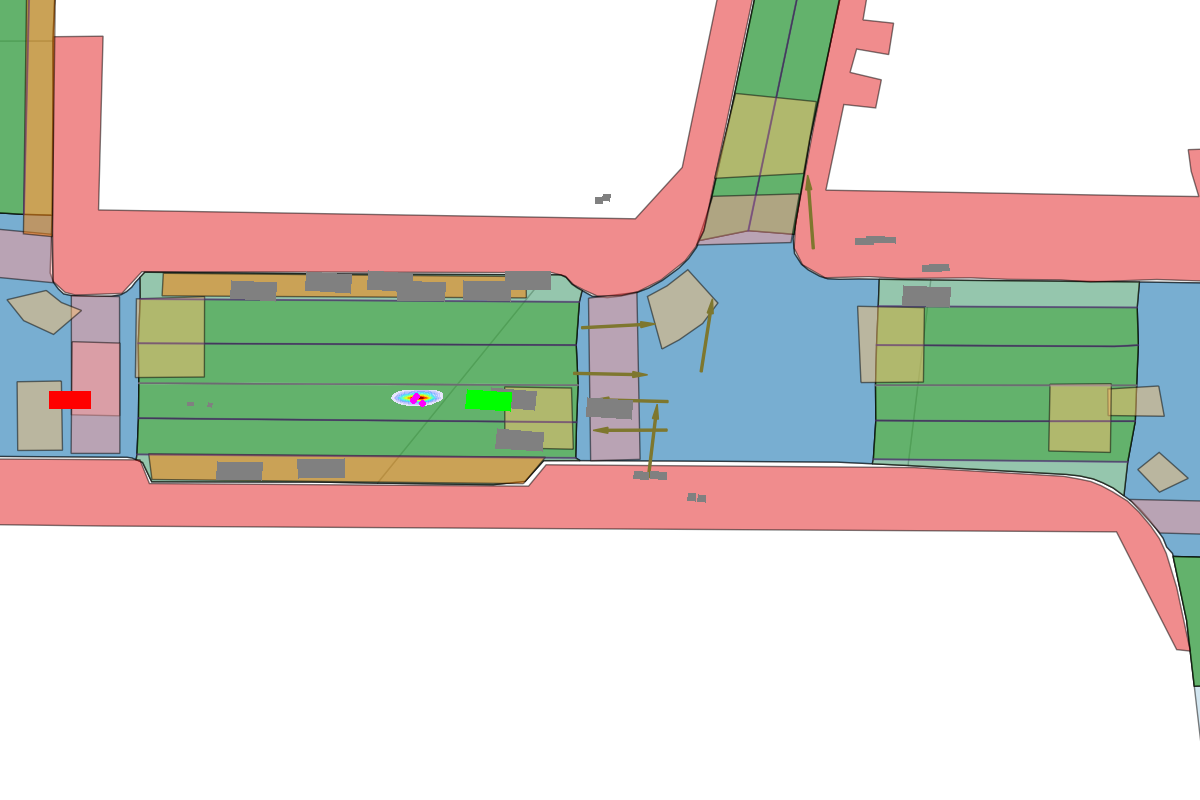}}}
    \qquad
      \subfloat[\gls{destination_predictor} Base - Frontal]{{\includegraphics[width=0.45\linewidth]{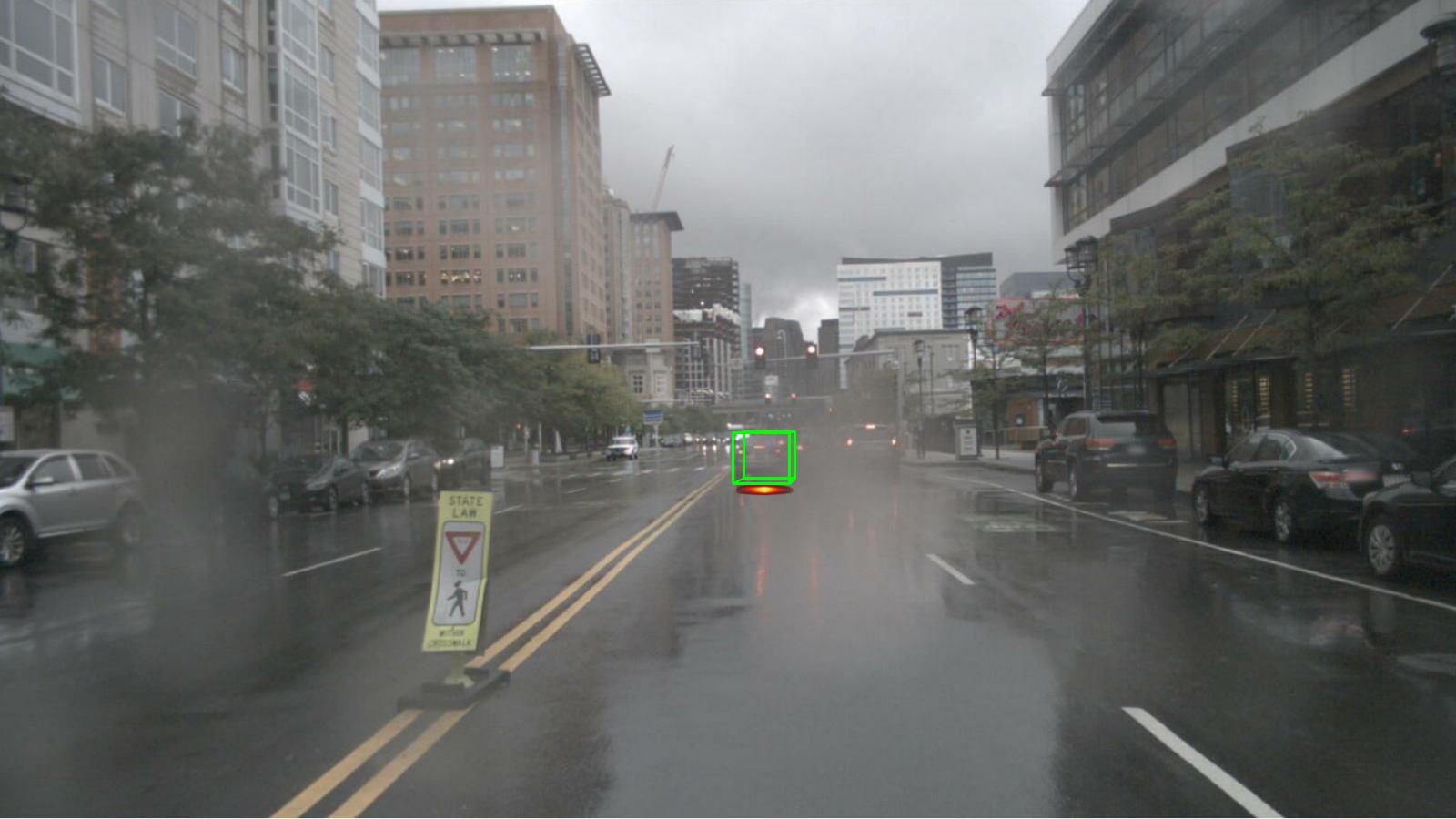} }}
    \subfloat[\gls{destination_predictor} Base - Top-down]{{\includegraphics[width=0.45\linewidth]{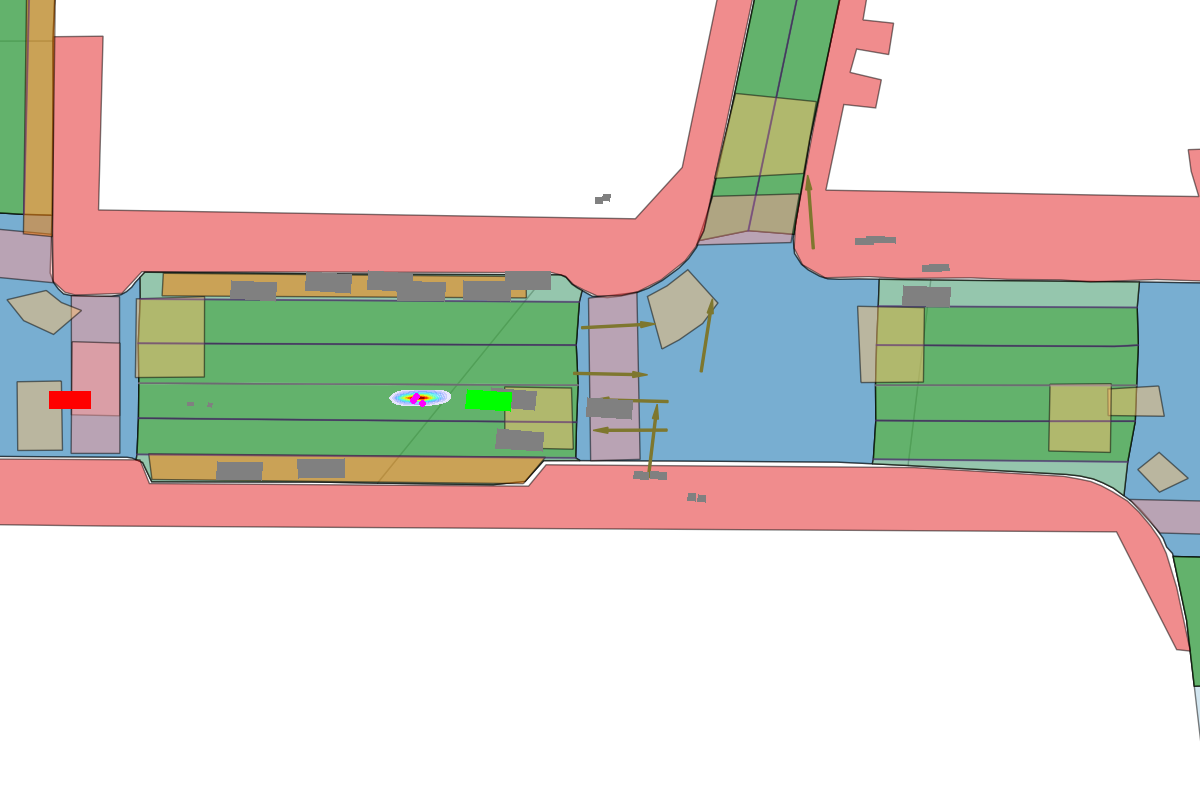}}}
    \caption{The issued command was: ``Slow down, a car is in front of you.''. This image showcases the difference between \gls{destination_predictor} using only the top-8 ranked components of the mixture (top row) against the same model using all components of the mixture. In this case, reducing the number of components has no substantial influence on the produced heatmap as the model believes there is only one possible destination.
    The red car on the top-down view is the ego car.
    The purple dots are the ground truth destination. The gray boxes represent other detected objects.
    The green rectangle in the top-down view and the frontal view indicates the referred object. }%
    \label{fig:top-k-vs-base-2}%
\end{figure*}

\begin{figure*}[h]
    \centering
    \subfloat[\gls{destination_predictor} Base - Frontal]{{\includegraphics[width=0.45\linewidth]{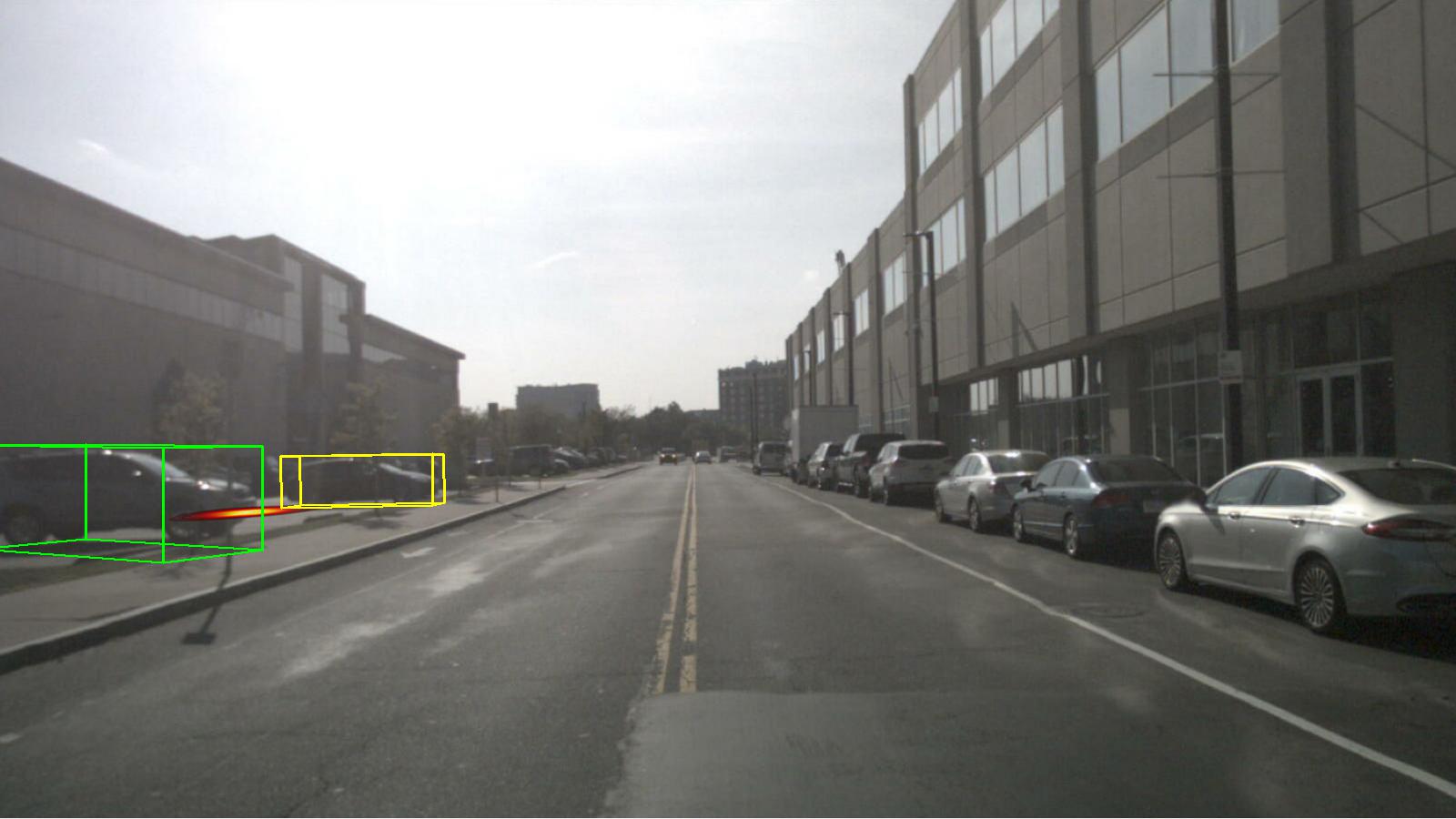}}}
    \subfloat[\gls{destination_predictor} Base - Top-down]{{\includegraphics[width=0.45\linewidth]{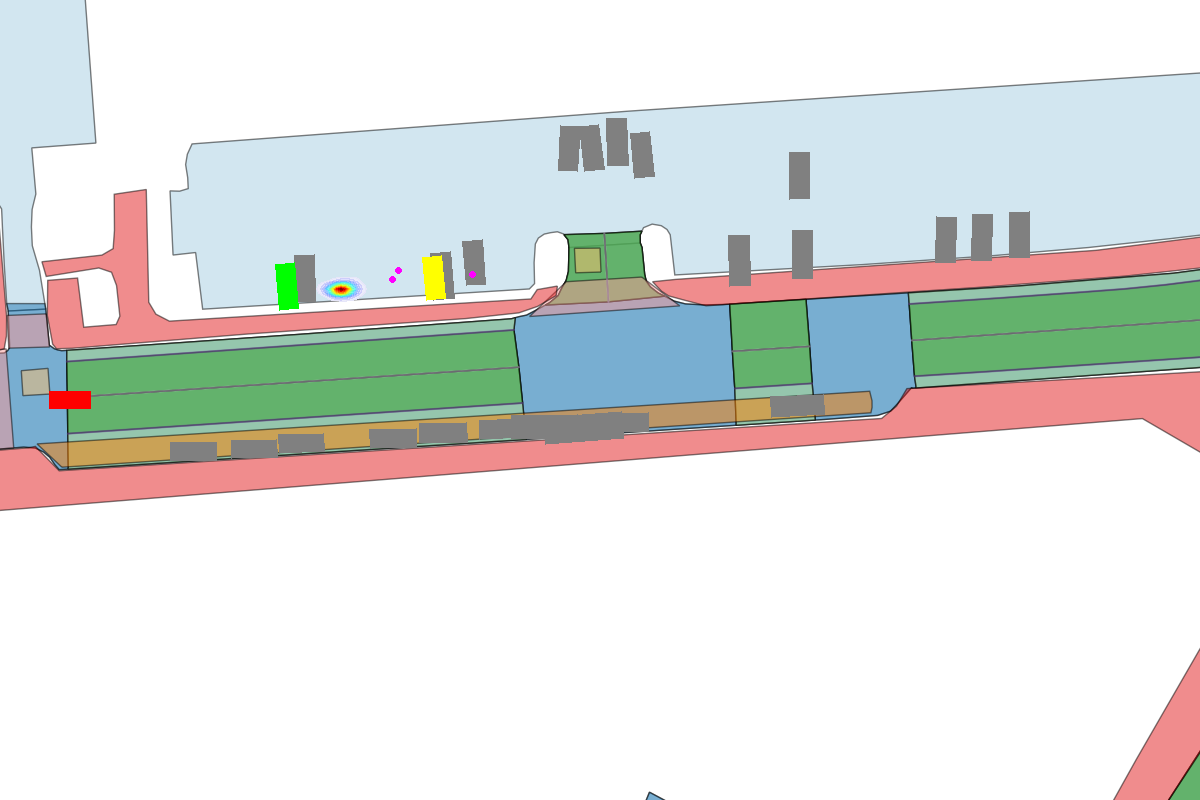}}}
    \caption*{The issued command was: ``Get a parking spot near the second car on the left side.''. This image showcases a failure case from the referred object detector side as it selected the wrong object as the referred object.}
    \qquad
    \subfloat[\gls{destination_predictor} Base - Frontal]{{\includegraphics[width=0.45\linewidth]{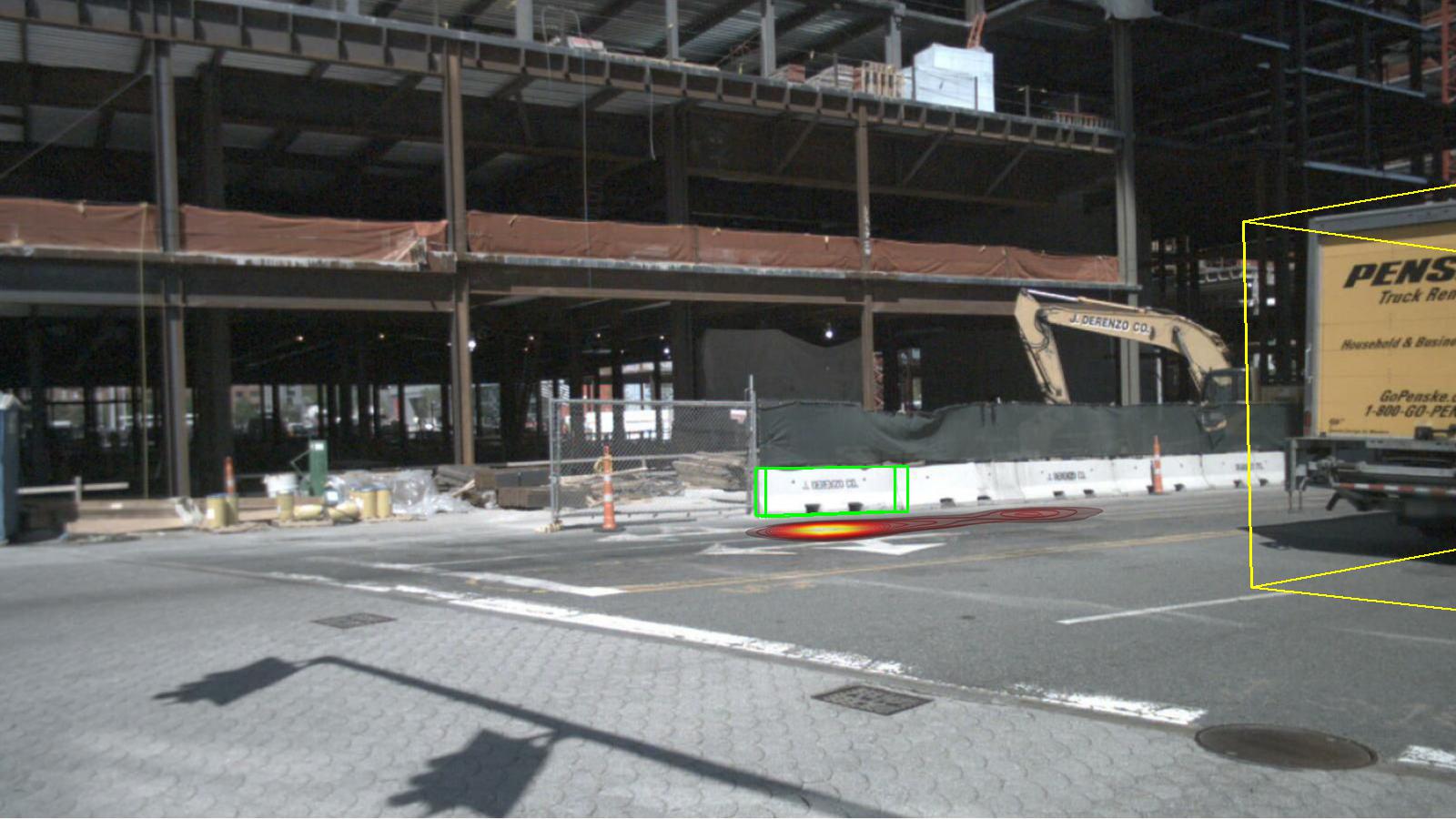}}}
    \subfloat[\gls{destination_predictor} Base - Top-down]{{\includegraphics[width=0.45\linewidth]{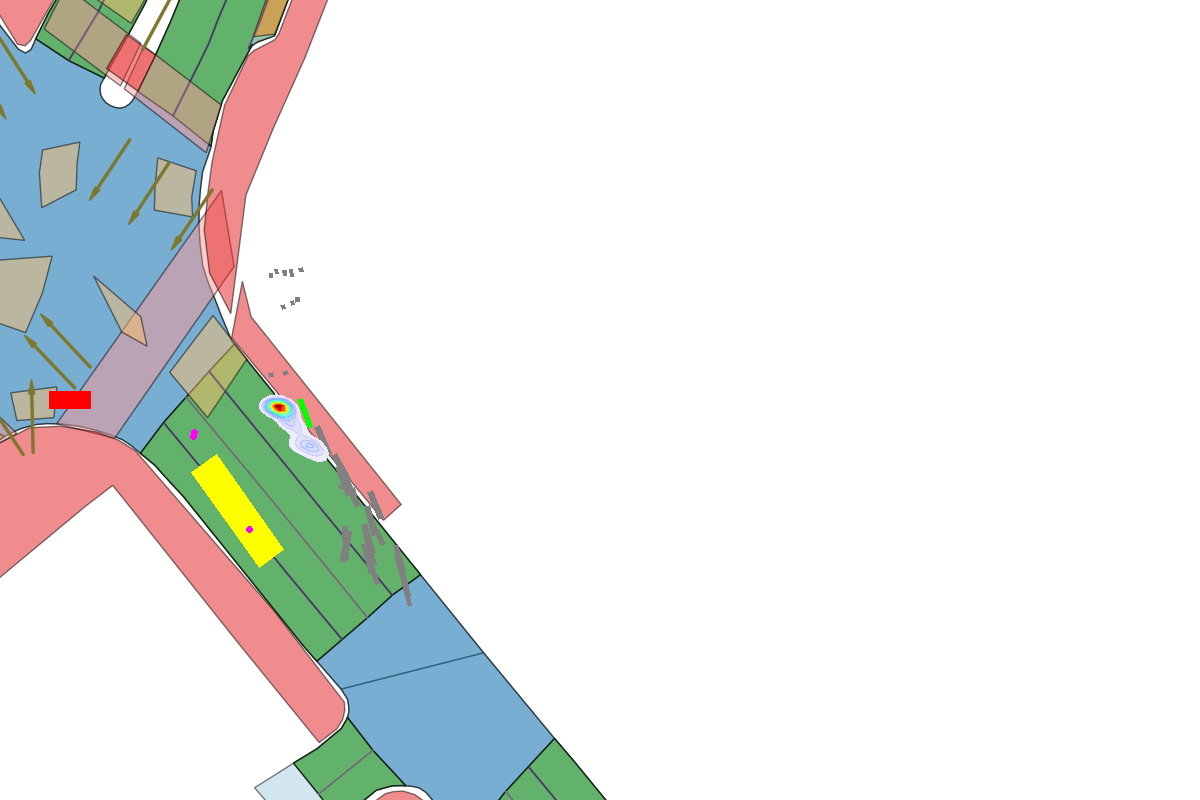}}}
    \caption*{The issued command was: ``Get in the lane behind this truck''. This image showcases a failure case from the 3D object detector side as the truck was not even detected.
    As a side result, the referred object detector could not predict the correct referred object.
 }
    
    \caption{Examples of failure cases. 
    The predicted object is indicated with a green bounding box in the frontal view and top-down view. 
    The ground truth 3D bounding box of the ground truth referred object is indicated in yellow in both the frontal and top-down view.
    The red car on the top-down view is the ego car.
    The purple dots are the ground truth destination. The gray boxes represent other detected objects.
    }%
    \label{fig:failure_cases}%
\end{figure*}

     \begin{table*}
        \centering
        \begin{adjustbox}{width=1\linewidth}
        \begin{tabular}{c||c|c|c|c|c|c|c|c|c|c}
        \diaghead{
        \theadfont Diag Column}%
          {Intent}{Model}&
        \thead{SinglePoint}&\thead{NonParametric}&\thead{UnimodalNormal}&\thead{MDN}&\thead{Adapted GoalGAN}&\thead{Endpoint VAE}&\thead{Adapted RegFlow}&\thead{\gls{destination_predictor}}&\thead{\gls{destination_predictor} - Top 64}&\thead{\gls{destination_predictor} - Top 32}\\ \hline \hline
Turn Left&9.94&12.18&10.20&10.48&9.40&9.98&19.71&9.47&8.83&\textbf{8.55}\\
Turn Right&\textbf{9.81}&12.81&10.34&10.66&10.21&10.11&20.22&10.82&10.43&10.28\\
Change Lane Left&5.43&6.58&5.94&5.25&5.69&\textbf{4.73}&14.09&5.07&4.89&4.75\\
Change Lane Right&7.55&9.77&8.19&7.61&8.85&7.57&16.21&7.71&7.39&\textbf{7.22}\\
U-Turn Left&7.50&9.20&8.10&7.55&9.47&7.39&15.31&7.86&7.19&\textbf{7.05}\\
U-Turn Right&\textbf{6.99}&8.36&7.48&7.39&7.04&7.46&15.24&8.75&8.52&8.48\\
Park&9.08&10.07&9.03&8.94&9.03&8.98&18.00&9.04&8.89&\textbf{8.80}\\
Stop&6.93&7.85&6.66&6.60&7.83&\textbf{6.48}&17.15&6.90&6.72&6.61\\
Pick Up&7.47&8.91&7.09&\textbf{6.53}&9.63&6.67&17.57&6.95&6.82&6.74\\
Continue&8.53&10.96&8.63&8.53&10.82&\textbf{8.09}&18.62&8.96&8.65&8.47\\
Overtake&10.68&13.50&10.40&10.46&11.34&\textbf{9.56}&20.06&10.30&10.07&9.98\\
Drop Off&8.71&8.96&8.17&8.01&9.13&7.75&16.76&7.85&7.68&\textbf{7.51}\\
Follow&8.42&9.82&8.34&8.06&8.11&\textbf{7.75}&16.94&8.22&8.07&7.99\\
Slow Down&6.31&7.46&6.29&6.35&7.04&\textbf{6.18}&16.75&6.88&6.68&6.56\\
Wait&\textbf{6.46}&8.52&7.13&6.96&6.47&6.70&18.84&8.07&7.81&7.60\\
Approach&9.33&11.96&8.99&9.02&10.14&8.21&18.85&8.46&8.20&\textbf{7.92}\\
Move Away&\textbf{7.51}&9.21&7.97&7.76&8.76&7.75&17.22&8.16&7.94&7.70\\
Other&8.49&10.14&\textbf{7.81}&9.34&9.91&8.39&17.68&9.04&8.40&7.94\\

\end{tabular}
\end{adjustbox}
\caption{The results of the evaluated models on the test set in terms of ADE [m] (the lower the better) over the different intent categories. Best results of each row are indicated in bold.}
\label{tab:intent_ade}
\end{table*}

        \begin{table*}
        \centering
        \begin{adjustbox}{width=1\linewidth}
        \begin{tabular}{c||c|c|c|c|c|c|c|c|c|c}
        \diaghead{
        \theadfont Diag Column}%
          {Intent}{Model}&
        \thead{SinglePoint}&\thead{NonParametric}&\thead{UnimodalNormal}&\thead{MDN}&\thead{Adapted GoalGAN}&\thead{Endpoint VAE}&\thead{Adapted RegFlow}&\thead{\gls{destination_predictor}}&\thead{\gls{destination_predictor} - Top 64}&\thead{\gls{destination_predictor} - Top 32}\\ \hline \hline
Turn Left&7.67&9.91&7.89&7.94&6.85&7.54&18.00&6.19&5.37&\textbf{5.11}\\
Turn Right&7.26&10.00&7.96&8.23&6.95&7.23&17.78&6.85&6.09&\textbf{5.61}\\
Change Lane Left&3.91&4.58&4.41&3.86&3.79&3.96&12.63&3.34&3.13&\textbf{3.01}\\
Change Lane Right&4.61&5.90&6.06&4.78&4.62&5.34&15.13&4.26&4.12&\textbf{3.82}\\
U-Turn Left&5.95&7.23&5.40&5.64&6.25&5.69&14.37&5.28&4.61&\textbf{4.36}\\
U-Turn Right&6.21&6.18&5.26&5.88&5.00&5.95&13.23&5.97&5.41&\textbf{4.87}\\
Park&6.24&6.73&6.07&5.56&5.45&5.72&15.68&5.51&5.39&\textbf{5.26}\\
Stop&4.45&5.34&4.05&4.22&4.77&4.13&15.28&3.91&3.67&\textbf{3.42}\\
Pick Up&5.69&6.17&4.95&4.45&5.69&4.51&15.63&3.35&2.93&\textbf{2.89}\\
Continue&5.18&7.03&6.53&5.45&5.35&5.32&16.30&5.90&5.31&\textbf{4.82}\\
Overtake&8.23&10.98&8.71&7.91&7.67&7.41&16.39&7.81&\textbf{7.24}&7.33\\
Drop Off&5.03&6.25&4.15&4.09&5.32&4.81&15.03&3.97&\textbf{3.67}&3.70\\
Follow&5.56&7.11&5.69&5.20&\textbf{4.80}&5.13&14.88&5.23&4.93&4.83\\
Slow Down&4.30&4.89&4.01&3.99&4.24&4.15&15.59&3.92&3.71&\textbf{3.52}\\
Wait&5.22&7.32&5.47&5.49&5.20&5.47&16.98&4.77&4.31&\textbf{3.90}\\
Approach&5.32&6.75&5.92&5.09&5.05&\textbf{4.99}&15.82&5.93&5.40&5.19\\
Move Away&6.04&7.44&6.59&\textbf{5.65}&8.23&5.79&16.22&6.20&5.97&5.83\\
Other&5.69&7.18&4.08&6.39&6.47&\textbf{3.98}&16.52&7.65&6.85&5.66\\
\end{tabular}
\end{adjustbox}
\caption{The results of the evaluated models on the test set in terms of MDE [m] (the lower the better) over the different intent categories. Best results of each row are indicated in bold.}
\label{tab:intent_MDE}
\end{table*}

        \begin{table*}
        \centering
        \begin{adjustbox}{width=1\linewidth}
        \begin{tabular}{c||c|c|c|c|c|c|c|c|c|c}
        \diaghead{
        \theadfont Diag Column}%
          {Intent}{Model}&
        \thead{SinglePoint}&\thead{NonParametric}&\thead{UnimodalNormal}&\thead{MDN}&\thead{Adapted GoalGAN}&\thead{Endpoint VAE}&\thead{Adapted RegFlow}&\thead{\gls{destination_predictor}}&\thead{\gls{destination_predictor} - Top 64}&\thead{\gls{destination_predictor} - Top 32}\\ \hline \hline
Turn Left&8.76&6.09&7.93&7.74&13.79&11.13&7.03&18.96&19.65&\textbf{20.45}\\
Turn Right&7.33&5.07&7.85&6.37&13.50&7.34&6.61&18.56&19.37&\textbf{20.67}\\
Change Lane Left&23.68&20.78&23.07&28.71&33.64&24.62&17.83&43.11&43.80&\textbf{44.81}\\
Change Lane Right&11.67&12.42&16.05&16.96&21.51&15.87&14.73&26.69&27.11&\textbf{28.35}\\
U-Turn Left&8.33&9.21&14.77&13.57&14.48&18.17&9.91&31.84&33.37&\textbf{36.02}\\
U-Turn Right&16.13&12.74&16.87&12.64&15.70&10.64&16.66&29.71&30.24&\textbf{31.05}\\
Park&10.02&10.74&15.26&13.48&20.56&12.80&11.36&27.89&28.41&\textbf{29.14}\\
Stop&17.27&16.83&22.38&22.76&27.61&21.94&14.81&32.56&32.98&\textbf{34.14}\\
Pick Up&12.94&15.56&20.61&21.61&21.57&21.16&15.90&32.85&33.07&\textbf{34.00}\\
Continue&19.64&11.15&16.91&17.15&21.08&14.47&13.92&22.64&23.11&\textbf{23.68}\\
Overtake&6.78&8.51&9.67&8.06&14.93&13.76&8.29&16.19&16.44&\textbf{16.83}\\
Drop Off&14.63&15.95&21.90&20.89&24.74&21.45&17.22&33.64&34.23&\textbf{34.88}\\
Follow&11.42&11.22&19.17&17.59&24.21&17.63&14.88&28.96&29.29&\textbf{29.81}\\
Slow Down&18.86&21.11&25.93&23.22&29.49&21.08&17.58&34.89&35.36&\textbf{36.31}\\
Wait&12.82&13.52&17.44&14.84&27.26&14.47&9.87&31.17&31.67&\textbf{32.90}\\
Approach&15.38&14.32&23.27&22.13&27.49&10.74&17.59&31.88&32.46&\textbf{33.47}\\
Move Away&14.29&8.67&11.34&11.50&15.43&12.62&9.51&16.75&17.70&\textbf{18.12}\\
Other&16.67&12.22&17.38&7.33&17.98&17.52&14.96&16.33&17.93&\textbf{19.23}
\end{tabular}
\end{adjustbox}
\caption{The results of the evaluated models on the test set in terms of $PA_2$ [\%] (the higher the better) over the different intent categories. Best results of each row are indicated in bold.}
\label{tab:intent_pa_2}
\end{table*}

        \begin{table*}
        \centering
        \begin{adjustbox}{width=1\linewidth}
        \begin{tabular}{c||c|c|c|c|c|c|c|c|c|c}
        \diaghead{
        \theadfont Diag Column}%
          {Intent}{Model}&
        \thead{SinglePoint}&\thead{NonParametric}&\thead{UnimodalNormal}&\thead{MDN}&\thead{Adapted GoalGAN}&\thead{Endpoint VAE}&\thead{Adapted RegFlow}&\thead{\gls{destination_predictor}}&\thead{\gls{destination_predictor} - Top 64}&\thead{\gls{destination_predictor} - Top 32}\\ \hline\hline
Turn Left&25.26&17.04&22.91&21.56&32.27&27.53&19.91&41.88&43.31&\textbf{44.92}\\
Turn Right&20.67&15.53&22.12&19.48&32.85&21.77&19.46&39.63&41.41&\textbf{43.12}\\
Change Lane Left&50.00&45.56&48.29&54.51&57.10&52.09&46.78&61.90&62.88&\textbf{63.92}\\
Change Lane Right&43.33&32.59&36.91&41.37&44.41&38.89&36.33&50.20&51.01&\textbf{52.38}\\
U-Turn Left&27.78&25.05&36.40&34.16&32.11&33.31&29.00&46.84&49.09&\textbf{51.24}\\
U-Turn Right&35.48&32.66&40.97&35.18&43.18&33.30&39.01&42.93&43.31&\textbf{43.49}\\
Park&29.41&28.78&35.30&34.01&41.96&33.32&30.46&43.38&43.98&\textbf{44.77}\\
Stop&44.35&38.78&47.96&47.75&48.00&48.32&35.49&53.64&54.45&\textbf{55.61}\\
Pick Up&43.53&36.44&42.44&46.71&42.93&45.55&35.80&55.08&55.41&\textbf{56.62}\\
Continue&38.39&28.64&37.08&40.39&42.08&37.80&33.05&44.33&45.24&\textbf{46.54}\\
Overtake&22.03&22.15&24.63&24.50&32.28&33.23&21.45&32.53&33.11&\textbf{33.25}\\
Drop Off&42.68&37.13&46.61&46.81&43.37&45.67&39.31&52.47&53.35&\textbf{54.20}\\
Follow&35.83&28.28&39.20&38.23&44.83&39.93&35.51&45.42&45.94&\textbf{46.47}\\
Slow Down&49.14&45.11&51.92&49.89&51.78&48.08&40.30&54.91&55.58&\textbf{56.61}\\
Wait&32.05&32.59&41.22&38.94&45.36&36.09&25.35&50.40&51.36&\textbf{52.94}\\
Approach&36.92&32.80&43.11&42.64&48.52&39.49&36.71&48.13&49.03&\textbf{49.79}\\
Move Away&23.81&23.26&29.61&31.30&31.30&31.06&27.47&37.25&38.17&\textbf{39.61}\\
Other&25.00&33.25&44.08&26.17&30.57&46.47&38.78&41.16&43.33&\textbf{47.84}
\end{tabular}
\end{adjustbox}
\caption{The results of the evaluated models on the test set in terms of $PA_4$ [\%] (the higher the better) over the different intent categories. Best results of each row are indicated in bold.}
\label{tab:intent_pa_4}
\end{table*}

}

\end{document}